\documentclass[]{style}

\microtypesetup{expansion=false}

\usepackage{amsmath,amssymb,mathtools}
\usepackage{algorithm}
\usepackage{algpseudocode}
\usepackage{booktabs}
\usepackage{enumitem}
\usepackage{float}
\usepackage{listings}
\usepackage{tabularx}
\usepackage{xspace}
\usepackage{hyperref}
\usepackage{url}
\usepackage{wrapfig}

\usepackage{amsmath,amsfonts,bm}

\def\eqref#1{Eq.~\ref{#1}}

\def\1{\bm{1}}

\DeclareMathAlphabet{\mathsfit}{\encodingdefault}{\sfdefault}{m}{sl}
\SetMathAlphabet{\mathsfit}{bold}{\encodingdefault}{\sfdefault}{bx}{n}

\usepackage{amsmath,amsfonts,bm}

\def\eqref#1{Eq.~\ref{#1}}

\def\1{\bm{1}}

\DeclareMathAlphabet{\mathsfit}{\encodingdefault}{\sfdefault}{m}{sl}
\SetMathAlphabet{\mathsfit}{bold}{\encodingdefault}{\sfdefault}{bx}{n}

\usepackage{multirow}
\usepackage{diagbox}
\usepackage{makecell}
\usepackage{tabularx}
\usepackage{graphicx}

\usepackage{array}
\usepackage{rotating}
\usepackage{colortbl}

\definecolor{aliceblue}{rgb}{0.94, 0.97, 1.0}
\definecolor{citecolor}{HTML}{0071BC}
\definecolor{linkcolor}{HTML}{ED1C24}
\definecolor{darkgreen}{HTML}{539165}

\makeatletter
\newcommand{\thickhline}{%
 \noalign {\ifnum 0=`}\fi \hrule height 1pt
 \futurelet \reserved@a \@xhline
}
\makeatother

\definecolor{MovieRollColor}{RGB}{240,246,255}

\renewcommand{\emph}[1]{\textit{#1}}
\DeclareTextFontCommand{\tablenum}{\bfseries\sffamily}
\DeclareTextFontCommand{\tablehead}{\bfseries\sffamily}
\title{MovieRoll: A Systematic Framework for Rolling Out Long-Form Videos}

\author[1,2,*,\ddag]{Yunyang Ge}
\author[2,*]{Weijie Kong}
\author[2,*]{Ryan Xu}
\author[2]{Yue Wu}
\author[2]{Qi Tian}
\author[2]{Jiangfeng Xiong}
\author[1,2]{Bin Lin}
\author[2,3]{Mingzhe Zheng}
\author[1]{Shenghai Yuan}
\author[1,2]{Zheyuan Liu}
\author[2,\S]{Jianwei Zhang}
\author[1,\dag]{Li Yuan}

\affiliation[1]{Peking University}
\affiliation[2]{Tencent Hunyuan}
\affiliation[3]{The Hong Kong University of Science and Technology}

\contribution[*]{Equal contribution}
\contribution[\S]{Project leader}
\contribution[\dag]{Corresponding author}

\abstract{Streaming video generation must sustain quality beyond the training horizon while handling self-generated history and growing cache costs. To address these challenges, we present \textbf{MovieRoll}, a \textbf{base-to-distilled} framework that connects streaming fine-tuning, long-rollout distillation, and long-horizon inference. We first establish a robust multi-step base through Window-wise Teacher Forcing (\textbf{WinTF}), which jointly denoises chunks at progressive noise levels with bidirectional attention within each window and causal access to preceding history. To retain historical context efficiently, hierarchical KV-cache compression (\textbf{HieraKV}) exploits layer-wise attention patterns, retaining full history in shallow layers and sink--recent history in deeper layers. This reduces cache memory by up to 78\% while improving generation quality. Building on this base, cumulative long tuning with distribution matching distillation (\textbf{CuLT-DMD}) holds parameters fixed throughout each long self-rollout and averages local gradients before updating. Each update therefore combines supervision across temporal positions while retaining only one local computation graph, enabling efficient long-rollout training of a few-step generator. To extend generation beyond the training horizon, KV recomputation bounds context for both models, while guidance annealing (\textbf{AnnealCFG}) reduces guidance toward the clean endpoint to stabilize MovieRoll-Base. Complementing these modeling advances, FBCA attention and PS-VAE decoding support efficient training and streaming output. On HeliosBench, MovieRoll-Base remains competitive at 30s after training on videos of up to 15s, and MovieRoll-Distilled achieves the highest total scores among the compared methods across single- and multi-event evaluations up to 60s, combining sustained naturalness with active motion.

}

\checkdata{\protect\makebox[1.2em][c]{\protect\raisebox{-0.15em}{\protect\includegraphics[width=1em,height=1em,keepaspectratio]{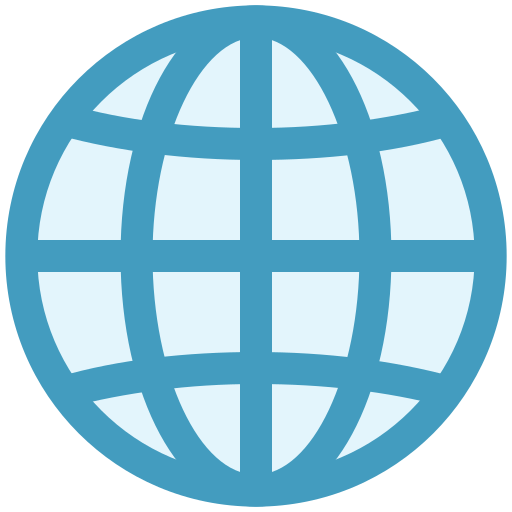}}}~\textsf{\textbf{Project Page:}} \href{https://Tencent-Hunyuan.github.io/MovieRoll/}{https://Tencent-Hunyuan.github.io/MovieRoll/}}
\checkdata{\protect\makebox[1.2em][c]{\protect\raisebox{-0.15em}{\protect\includegraphics[width=1em,height=1em,keepaspectratio]{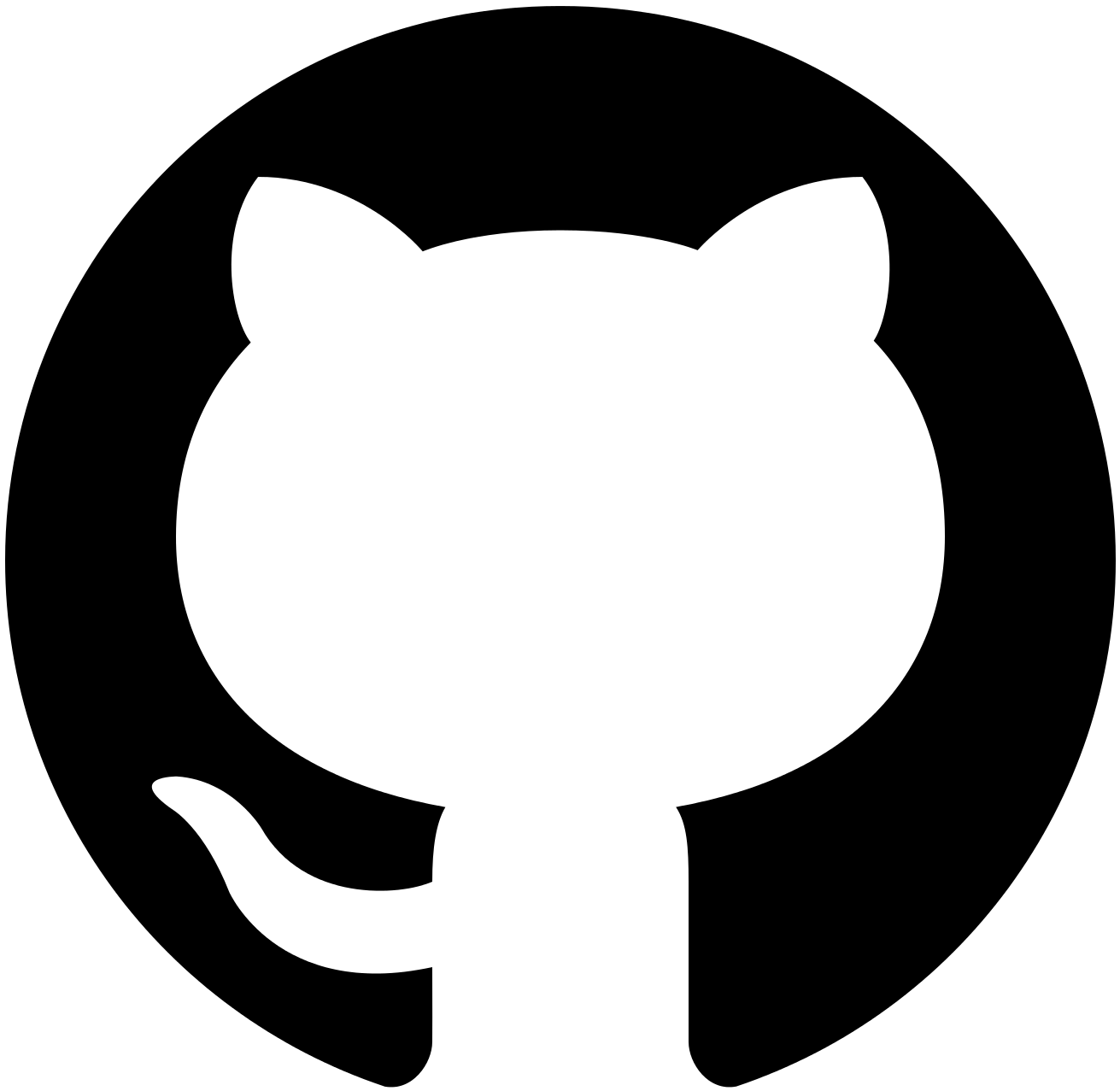}}}~\textsf{\textbf{Code:}} \href{https://github.com/Tencent-Hunyuan/MovieRoll}{https://github.com/Tencent-Hunyuan/MovieRoll}}
\checkdata{\protect\makebox[1.2em][c]{\protect\raisebox{-0.15em}{\protect\includegraphics[width=1em,height=1em,keepaspectratio]{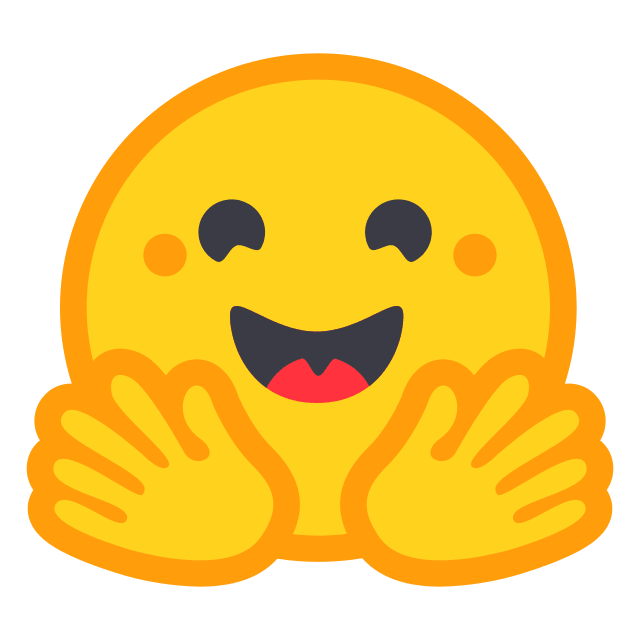}}}~\textsf{\textbf{Hugging Face:}} \href{https://huggingface.co/yunyangge/MovieRoll}{https://huggingface.co/yunyangge/MovieRoll}}
\checkdata{\protect\makebox[1.2em][c]{\protect\raisebox{-0.15em}{\protect\includegraphics[width=1em,height=1em,keepaspectratio]{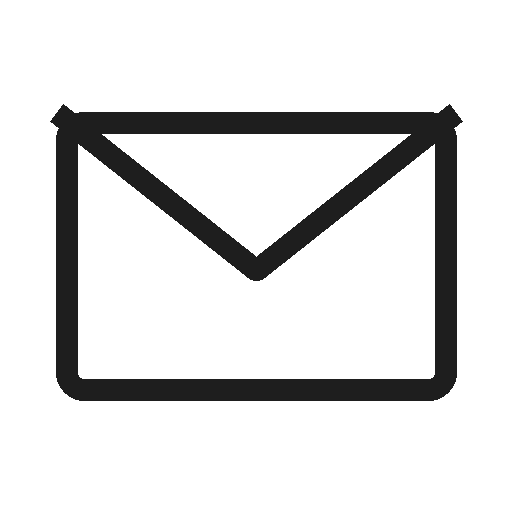}}}~\textsf{\textbf{Email Address:}} \href{mailto:yunyang@stu.pku.edu.cn}{yunyang@stu.pku.edu.cn}}

\newcommand{\internshipfootnote}{%
  \begingroup
  \renewcommand{\thefootnote}{\fnsymbol{footnote}}%
  \footnotetext[3]{Work done during internship at Tencent Hunyuan}%
  \endgroup
}

\begin{document}
\thispagestyle{firststyle}
\maketitle
\internshipfootnote
\vspace{-6mm}

\begin{figure*}[!t]
    \centering
    \includegraphics[width=\linewidth]{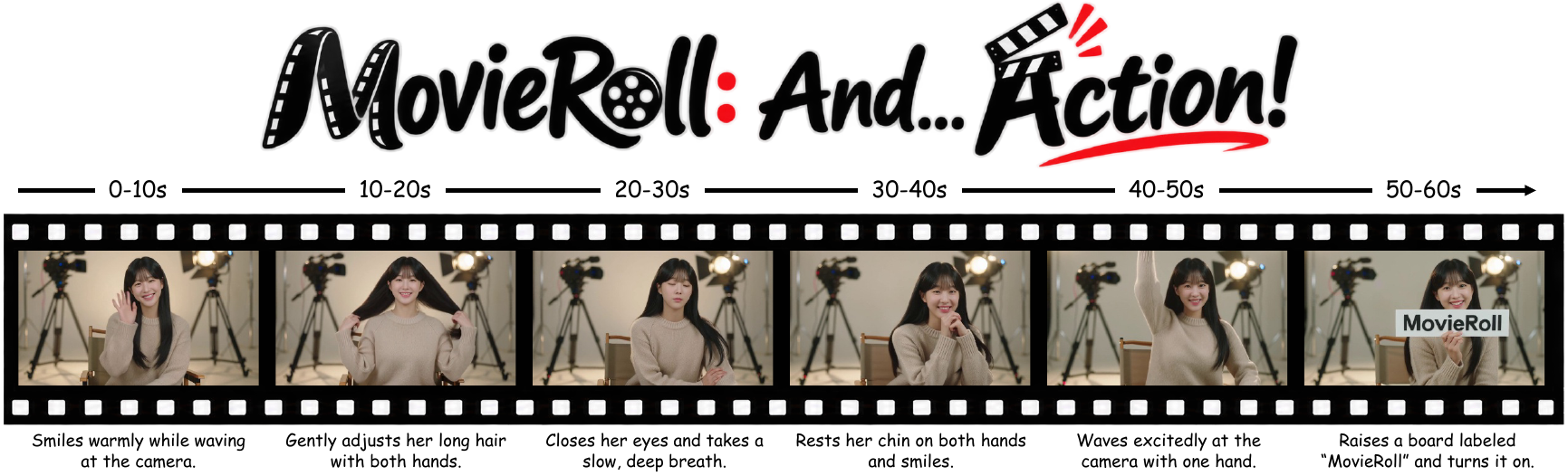}
    \caption{\textbf{MovieRoll generates a minute-long video with continuous subject identity across six events.} The sequence shows the same woman performing successive actions in one studio setting, with a new prompt every 10s. Each frame represents one event, and the text below summarizes its prompt.}
    \label{fig:teaser}
\end{figure*}

\section{Introduction}
\label{sec:intro}

\begin{center}
\textit{\large\fontencoding{T1}\fontfamily{ppl}\selectfont A few seconds can capture a moment, but a story needs time to unfold.}\\
\textsc{--- MovieRoll Team}
\end{center}

Visual generative models now support realistic video synthesis~\citep{open_sora,open_sora2,open_sora_plan,osp_next,cogvideox,wan,hunyuanvideo,hunyuanvideo1.5,ltx,ltx2,movie_gen,kandinsky5}, synchronized audio-visual generation~\citep{ltx2,kandinsky6}, and increasingly flexible content control~\citep{flashi2v,consisid,skyreels_v3,swapanyone,uniworld}. However, bidirectional video models jointly denoise the full sequence, delaying the first output and increasing computation as duration grows. Streaming generation instead produces chunks sequentially from previously generated content~\citep{diffusion_forcing,causvid,longlive,vidu_s1,hunyuan_worldplay,hunyuan_world1.5,matrix_game,wan_streamer_v1,wan_streamer_v2}, enabling continuous output and online interaction.

Sustaining quality over long sequences remains challenging. With teacher forcing, models learn from short videos conditioned on ground-truth history. At inference, the shift to self-generated history introduces exposure bias, while longer generation requires extrapolation beyond the training horizon. Growing KV caches further increase memory and computation costs. Self Forcing~\citep{self_forcing} addresses exposure bias through self-rollout distillation, and LongLive~\citep{longlive} extends this approach to long sequences. These approaches primarily target few-step generators, leaving capable multi-step streaming bases less explored. Resampling Forcing~\citep{resampling_forcing} and Helios-Base~\citep{helios} address multi-step generation, but the former evaluates durations up to 15s, while the latter uses bidirectional continuation without KV caching. History management introduces further tradeoffs. Sliding windows discard intermediate context~\citep{self_forcing,causal_forcing,rolling_forcing,longlive}, while retrieval~\citep{moc}, evolving memory~\citep{echo_infinity}, and context compression~\citep{deep_forcing,context_forcing} introduce additional mechanisms.

We present \textbf{MovieRoll}, a \textbf{base-to-distilled} framework that builds a multi-step base through streaming fine-tuning and a few-step generator through long self-rollouts. As illustrated in Fig.~\ref{fig:teaser}, MovieRoll supports long-form generation with sequential event prompts. At the base stage, Window-wise Teacher Forcing (\textbf{WinTF}) adapts a pretrained bidirectional model using rolling windows~\citep{rolling_diffusion,magi1,rolling_forcing}. Each window jointly denoises chunks at progressive noise levels with bidirectional attention and causal access to preceding clean history. Parallel window training improves long-horizon robustness over chunk-wise training. Hierarchical KV-cache compression (\textbf{HieraKV}) retains full history in shallow layers and only sink--recent history in deeper layers. Compared with full-history caching across all layers, it reduces KV-cache memory by up to 78\% and peak denoising-step latency with CFG by approximately 53\%, while improving generation quality.

Building on this base, cumulative long tuning with distribution matching distillation (\textbf{CuLT-DMD}) trains a few-step generator on long self-rollouts under DMD supervision~\citep{dmd,dmd2}. Whereas LongLive-style training updates from individual supervision windows, CuLT-DMD holds parameters fixed throughout each rollout and averages local gradients before updating. Each update combines supervision across time, keeping one local computation graph at a time and improving long-horizon generation over LongLive-style training.

At inference, threshold-triggered history selection and KV recomputation rebuild the cache from representative generated chunks for both MovieRoll-Base and MovieRoll-Distilled, bounding the retained context and improving extrapolation beyond the training horizon. Classifier-free guidance annealing (\textbf{AnnealCFG}) further stabilizes MovieRoll-Base by reducing guidance strength~\citep{cfg} toward the clean endpoint, limiting late-stage extrapolation before generated chunks become history. The Streaming Video Prompt Rewriting Skill expresses sustained actions and maintains shared entities across event prompts.

MovieRoll also provides infrastructure for efficient training and streaming output. Flash Block-wise Causal Attention (FBCA) packs visible keys and values for FlashAttention, accelerating window-parallel training across input shapes. Parallel Streaming VAE (PS-VAE) preserves temporal state across decoded chunks and distributes spatial computation across devices, lowering latency and memory use for incremental decoding.

On HeliosBench~\citep{helios}, MovieRoll-Base remains competitive at 30s after training on videos of up to 15s. MovieRoll-Distilled achieves the highest total scores among the compared methods for single-event generation at 10s, 30s, and 60s and multi-event generation at 30s and 60s, combining sustained naturalness with active motion. A user study further shows majority preference for MovieRoll-Distilled over every evaluated baseline in both settings.

In summary, MovieRoll makes four contributions. First, it establishes a robust multi-step streaming base through Window-wise Teacher Forcing and HieraKV. Second, it introduces CuLT-DMD to train a few-step generator by aggregating supervision across long self-rollouts. Third, it supports long-horizon inference through KV recomputation for both models and AnnealCFG for MovieRoll-Base. Finally, it provides FBCA and PS-VAE for efficient training and streaming decoding.

\section{Related Work}
\label{sec:related_work}

\subsection{Bidirectional Video Generation Models}
Bidirectional video diffusion jointly denoises a latent sequence, allowing each frame to use both past and future context. Together with spatiotemporal compression and large-scale training, this formulation supports high-quality video synthesis~\citep{open_sora,open_sora_plan,cogvideox,hunyuanvideo,hunyuanvideo1.5,wan}. Joint refinement lets the model coordinate appearance and motion across the clip throughout denoising.

Scaling to longer clips increases attention memory and computation. Hybrid linear--softmax attention~\citep{sana_video2} and sparse attention in pretrained video transformers~\citep{sparse_video_gen,sparse_video_gen2} reduce this cost while retaining joint denoising across the clip. A complementary direction changes the generation order so that completed chunks can be emitted before the entire sequence is available.

\subsection{Autoregressive Video Generation Models}
To support sequential output, temporal noise schedules separate the denoising progress of different frames. Diffusion Forcing~\citep{diffusion_forcing} permits independent noise levels, while SkyReels-V2~\citep{skyreels_v2} uses non-decreasing schedules. FIFO-Diffusion~\citep{fifo_diffusion} applies staggered denoising to a frame queue at inference, while Rolling Forcing~\citep{rolling_forcing} trains a rolling-window generator that jointly refines chunks at ordered noise levels. Robust continuation further benefits from guidance with variable-length history~\citep{diffusion_forcing_transformer} and training with corrupted history~\citep{helios}.

Few-step distillation reduces per-chunk latency, as demonstrated by CausVid~\citep{causvid}, Helios-Distilled~\citep{helios}, and Krea-Realtime~\citep{krea}. Beyond distribution matching objectives~\citep{dmd,dmd2}, initialization and rollout exposure shape generation quality. Causal Forcing~\citep{causal_forcing} uses an autoregressive teacher for ODE initialization, while Self Forcing~\citep{self_forcing} trains on self-generated histories. For longer rollouts, LongLive~\citep{longlive} applies DMD supervision to each newly generated clip. Self-Forcing++~\citep{self_forcing_pp} uses a short-video teacher to supervise sampled segments of long self-generated videos. Context Forcing~\citep{context_forcing} additionally gives the teacher access to long-range history.

Using longer context also raises the cost of storing and processing history. Temporal pyramids~\citep{pyram_flow} and coarser tokens for distant frames~\citep{long_context_ar} reduce the historical token count, while block linear attention accumulates a fixed-size state~\citep{sana_video}. Within KV caches, compression can follow the distinct roles of layers and heads. PyramidKV~\citep{pyramidkv} varies cache allocation across language-model layers. In video models, head-specific pruning~\citep{forcing_kv} and calibrated layer--head token merging~\citep{calm} adapt compression to attention structure.

\subsection{Interactive Video Generation Models}
Beyond sustaining a sequence, interactive generation must respond to changing user input while preserving scene continuity. Applications range from navigable environments~\citep{genie3,hunyuanworld} to camera-controlled video~\citep{sana_wm}. For streaming video, history management must therefore account for the active condition as well as memory cost. LongLive~\citep{longlive} retains recent frames and a persistent sink, refreshing cached KV states under the new prompt at prompt switches. For sink-based histories, LoL~\citep{lol} uses multi-head RoPE jitter to reduce repeated returns to sink-frame content. Adaptive memory policies retrieve prompt-relevant history~\citep{memflow} or learn an evolving memory as frames leave the local window~\citep{echo_infinity}.

Meeting interactive latency targets requires both model and system design. On-policy distillation enables fast multimodal avatar generation~\citep{livetalk}, while deadline-aware scheduling reduces delays in live streams~\citep{stream_diffusion_v2}. SANA-Streaming~\citep{sana_streaming} combines model and system optimization for temporally consistent video editing.

\section{MovieRoll}
\label{sec:method}

\begin{figure}[!t]
    \centering
    \includegraphics[width=\linewidth]{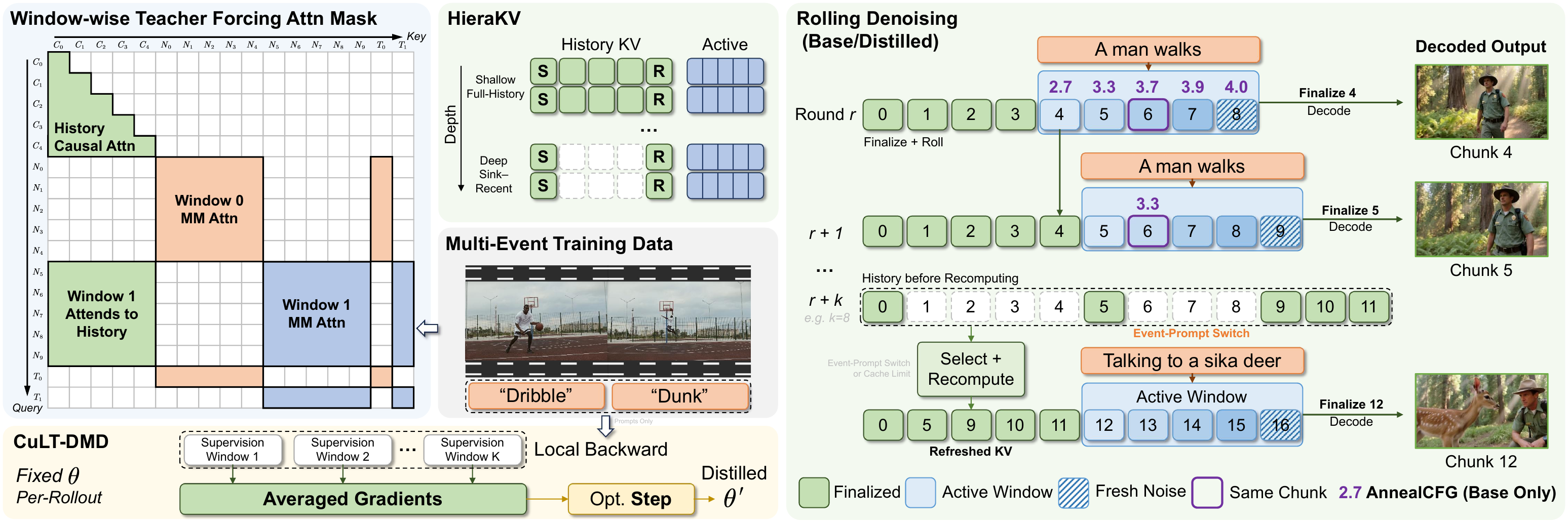}
    \caption{\textbf{Overview of MovieRoll.} Window-wise Teacher Forcing (WinTF) trains MovieRoll-Base through parallel window supervision conditioned on clean history and event prompts. After ODE initialization, Cumulative Long Tuning with Distribution Matching Distillation (CuLT-DMD) trains a few-step generator by averaging local gradients across long self-rollouts. Both models jointly denoise a rolling window, use hierarchical KV-cache compression (HieraKV) to retain history across layers, and rebuild the cache at a history threshold or prompt switch. Classifier-free guidance annealing (AnnealCFG) gradually reduces guidance during base-model denoising.}
    \label{fig:method_overview}
\end{figure}

\subsection{Overview}
\label{sec:method_overview}

MovieRoll jointly denoises a rolling window of latent chunks at progressively increasing noise levels. Fully denoised chunks are emitted, and their KV states are cached for subsequent generation. Attention is bidirectional within the active window and causal with respect to fully denoised historical chunks.

As outlined in Fig.~\ref{fig:method_overview}, we first adapt a pretrained bidirectional video model into MovieRoll-Base using \textbf{Window-wise Teacher Forcing (WinTF)}. Each training pass supervises multiple windows conditioned on preceding ground-truth chunks and event-specific prompts, enabling MovieRoll-Base to learn transitions across events.

We initialize a five-step student using ODE trajectories from MovieRoll-Base, then train it on long self-generated rollouts using \textbf{Cumulative Long Tuning with DMD (CuLT-DMD)}. A frozen bidirectional model provides real-score estimates. CuLT-DMD updates the generator with local gradients accumulated over each rollout.

At inference, \textbf{HieraKV} stores all available historical KV states in shallow blocks and retains only sink--recent history in deeper blocks. Threshold-triggered KV recomputation periodically rebuilds the historical KV cache from selected fully denoised chunks. \textbf{AnnealCFG} decreases the guidance scale during MovieRoll-Base denoising. The Streaming Video Prompt Rewriting Skill converts user instructions into descriptions of ongoing actions and consistent entities for single-event and multi-event generation.

\subsection{WinTF: Window-wise Teacher Forcing}
\label{sec:method_wintf}
\begin{figure}[!t]
    \centering
    \includegraphics[width=\linewidth]{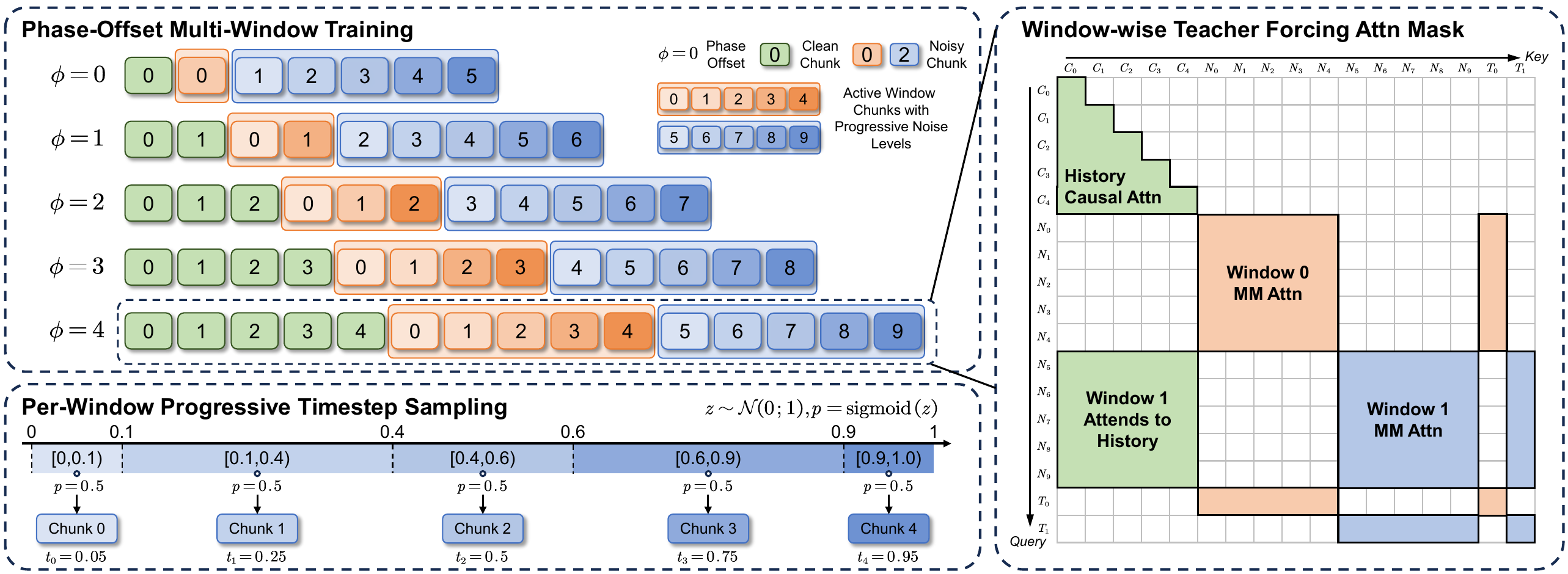}
    \caption{\textbf{WinTF learns rolling generation through parallel supervision of windows with progressive noise levels.} Phase offsets expose the model to both short prefill windows and full decoding windows. Within each window, later chunks receive higher noise levels. The attention mask combines bidirectional interactions within a noisy window with access to clean preceding history, while isolating noisy windows from one another. This allows multiple windows to be trained in parallel, each conditioned on its own event prompt and preceding clean history.}
    \label{fig:method_winTF}
\end{figure}

With chunk-wise teacher forcing, denoising each new chunk from pure noise encourages excessive reliance on preceding content, allowing historical errors to accumulate during generation. Rolling Forcing~\citep{rolling_forcing} mitigates this problem through joint denoising within a rolling window, although its training formulation focuses on few-step distillation. PA-VDM~\citep{pa_vdm} and Rolling Diffusion~\citep{rolling_diffusion} also train rolling-window models, but their bidirectional attention couples historical representations to the active window, preventing reuse of historical KV states and limiting access to earlier content as the window advances. Magi-1~\cite{magi1} supports KV caching through causal attention across chunks, yet imposes the same constraint within the active window, restricting mutual refinement among neighboring chunks. Building on these designs, WinTF extends rolling-window denoising to base-model training, where bidirectional attention within each window allows neighboring chunks to refine one another and causal access to preceding history preserves context. This attention structure also enables multiple windows to be supervised in parallel under their respective event prompts and clean histories, making training efficient while retaining the temporal consistency of joint denoising.

A full window contains $W$ chunks. Regular chunks contain $C$ latent frames. Let $i\in\{0,\ldots,M-1\}$ index training windows and $j\in\{0,\ldots,m_i-1\}$ index chunks within window $i$, where $m_i\leq W$. We denote the clean window by $\mathbf{x}_0^i$, its preceding ground-truth history by $\mathbf{x}_0^{<i}$, and its text condition by $\mathbf{c}_i$. All chunks in a window share one prompt. As shown in Fig.~\ref{fig:method_winTF}, noisy chunks attend bidirectionally within their own window and to preceding history. Attention between noisy windows is masked.

Frames in the same chunk share a timestep. We use normalized timesteps $t\in[0,1]$, with $0$ denoting clean data and $1$ pure noise. To assign progressively higher noise levels to later chunks, we split $[0,1]$ using $m_i-1$ sorted independent uniform samples, with boundaries $b_{i,0}=0$ and $b_{i,m_i}=1$. A shared relative position $p_i=\operatorname{sigmoid}(z_i)$, where $z_i\sim\mathcal{N}(0,1)$, then gives
\begin{equation}
    q_{i,j}=b_{i,j}+p_i(b_{i,j+1}-b_{i,j}),
    \qquad
    t_{i,j}=\frac{s\,q_{i,j}}{1+(s-1)q_{i,j}},
    \label{eq:wintf_timestep}
\end{equation}
where $s>0$ is the timestep-shift factor. This monotonic transformation preserves the ordering of noise levels and supplies the timesteps for both noising and model conditioning. Training occasionally assigns a shared timestep to all chunks, as described in App.~\ref{app:implementation}.

Inference starts with a growing window during prefill and then advances a full window during decoding. To cover both phases, each data-parallel group samples an offset $\phi\in\{0,\ldots,W-1\}$. The first training window contains up to $\phi+1$ available chunks, and subsequent windows contain $W$ chunks. Varying the offset exposes the model to different window sizes and chunk positions.

For text-to-video training, the latent sequence supplies the training windows and their histories. For image-to-video training, the first latent frame serves as shared history, and training windows are formed from the remaining frames. Let $\mathbf{h}_i$ denote the history conditioning derived from $\mathbf{x}_0^{<i}$, including the history noise perturbation specified in Tab.~\ref{tab:training_hyperparameters}. Joint flow matching~\citep{flow_matching,rectified_flow} with $\mathbf{t}_i=(t_{i,0},\ldots,t_{i,m_i-1})$ gives
\begin{equation}
    \mathcal{L}_{\mathrm{FM}}
    =
    \mathbb{E}\left[
        \frac{1}{N_{\mathrm{lat}}}
        \sum_{i=0}^{M-1}
        \left\|
            v_\theta
            \left(
                \mathbf{x}_{\mathbf{t}_i}^{i},
                \mathbf{t}_i,
                \mathbf{h}_i,
                \mathbf{c}_i
            \right)
            -
            \mathbf{v}^{i}
        \right\|_2^2
    \right],
    \label{eq:wintf_fm}
\end{equation}
where $N_{\mathrm{lat}}$ is the total number of supervised latent entries across all windows of a sample. The linear path uses $\mathbf{x}_{\mathbf{t}_i}^{i}=(1-\mathbf{t}_i)\odot\mathbf{x}_0^i+\mathbf{t}_i\odot\boldsymbol{\epsilon}^i$ and target velocity $\mathbf{v}^i=\boldsymbol{\epsilon}^i-\mathbf{x}_0^i$, with timesteps broadcast over chunk entries. The expectation covers training videos, Gaussian noise, phase offsets, timesteps, and history perturbations.

\subsection{HieraKV: Hierarchical KV-Cache Compression across Layers}
\label{sec:method_hierakv}
\begin{figure}[!t]
    \centering
    \includegraphics[width=\linewidth]{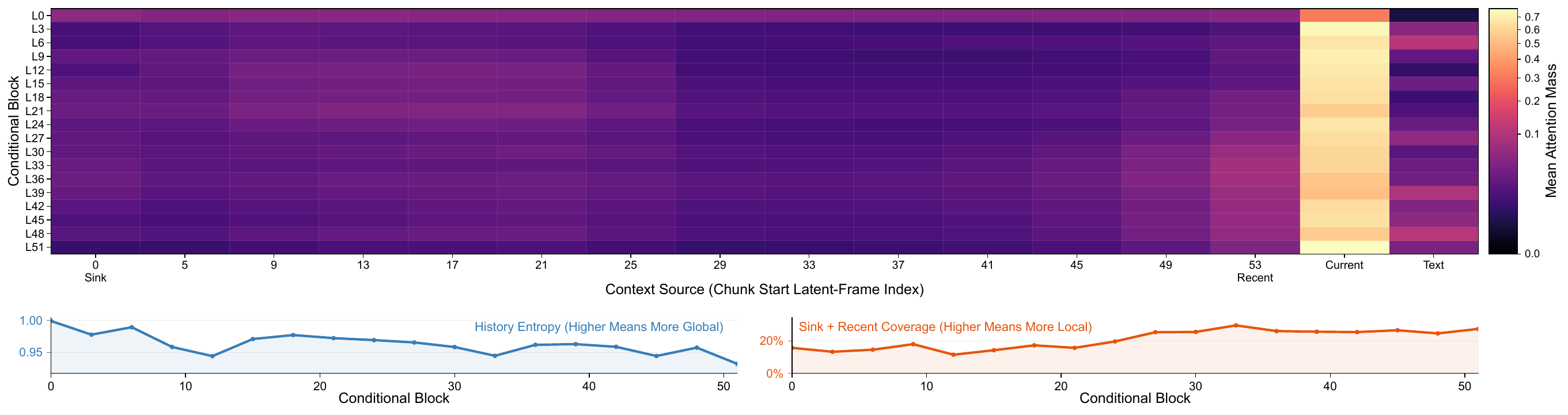}
    \caption{\textbf{Layer-dependent attention patterns motivate hierarchical history retention.} In the WinTF-trained model, shallow layers distribute attention broadly across historical chunks, while deeper layers place relatively more attention on the sink and most recent chunks. The heatmap and lower entropy and coverage plots show this shift with depth. HieraKV follows these patterns by retaining full historical context in shallow layers and restricting deeper layers to the sink and most recent chunks.}
    \label{fig:method_hierakv}
\end{figure}

Full-history KV caching incurs increasing memory and computation costs as video length grows, whereas uniform sliding-window caching discards intermediate history at every layer. Our inference-time analysis of the WinTF-trained model shows that attention to history varies with network depth, as illustrated in Fig.~\ref{fig:method_hierakv}. Shallow blocks distribute attention relatively evenly across historical chunks, while deeper blocks concentrate more attention on sink and recent chunks. Historical-attention entropy also decreases modestly with depth.

Guided by these attention patterns, \textbf{HieraKV} varies history retention across depth. The first $L_s$ blocks retain all available historical KV states, while deeper blocks cache only the sink chunk and the most recent historical chunk. The sink anchors visual content, and recent history provides immediate temporal context. Intermediate history remains accessible through the shallow-block outputs, while the active window retains bidirectional attention at every depth.

\subsection{CuLT-DMD: Cumulative Long Tuning with DMD}
\label{sec:method_rolling_longtuning_dmd}
\begin{figure}[!t]
    \centering
    \includegraphics[width=\linewidth]{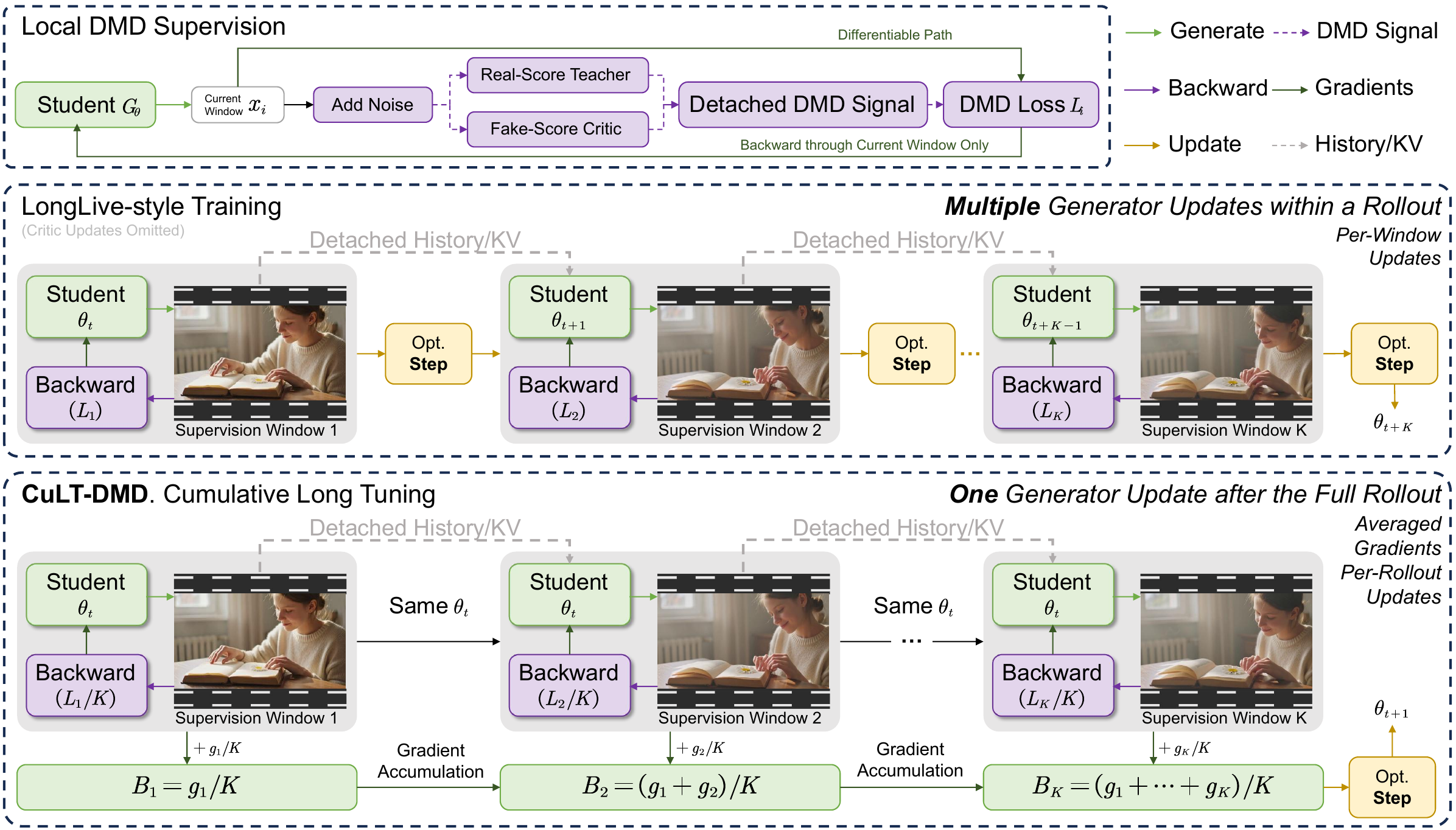}
    \caption{\textbf{CuLT-DMD combines supervision across a long rollout in each generator update.} The top panel shows local distillation supervision from teacher and critic predictions. LongLive-style training updates the generator from individual windows while carrying detached history forward. CuLT-DMD instead keeps parameters fixed throughout the rollout and averages the local gradients before updating. This incorporates supervision from multiple temporal positions while retaining one local computation graph at a time.}
    \label{fig:method_cult_dmd}
\end{figure}

\subsubsection{Cumulative Long Tuning}
Self Forcing~\citep{self_forcing} uses self-rollout distillation to narrow the training--inference gap. LongLive~\citep{longlive} extends training to long rollouts, retaining history and KV states across local parameter updates. \textbf{CuLT-DMD} instead keeps parameters fixed throughout a rollout and combines supervision across temporal positions in one update, as illustrated in Fig.~\ref{fig:method_cult_dmd}.

For generator update $r$, we fix $\theta=\theta_r$ and denote the local loss at supervision window $k$ by $\mathcal{L}_{\mathrm{DMD}}^{(k)}$. A rollout contains $K$ supervision windows. We backpropagate each $\mathcal{L}_{\mathrm{DMD}}^{(k)}/K$ and release its graph, retaining the accumulated gradients and detached history. After all $K$ windows, the generator update is
\begin{equation}
    \mathbf{g}_{\mathrm{roll}}
    = \frac{1}{K}\sum_{k=1}^{K}
    \left.
        \nabla_{\theta}\mathcal{L}_{\mathrm{DMD}}^{(k)}
    \right|_{\theta=\theta_r},
    \qquad
    \theta_{r+1}
    = \operatorname{OptStep}(
        \theta_r,\mathbf{g}_{\mathrm{roll}}
    ).
    \label{eq:rolling_gradient_accumulation}
\end{equation}
The critic likewise averages local gradients before updating at the rollout boundary. This aggregates supervision over long sequences while retaining one local computation graph at a time.

Score-clip and history/KV budgets bound the context used by each local graph. The training schedule is detailed in App.~\ref{app:implementation}.

\subsubsection{Window-Level Supervision}
The local DMD loss supervises the current window using generated history as context. At window $k$, the generator predicts clean latents $\hat{\mathbf{x}}_0^{(k)}$ from a noisy window with chunk timesteps $\boldsymbol{\tau}$ and retained history/KV state $\hat{\mathbf{H}}^{(k)}$. We prepend a score prefix $\hat{\mathbf{h}}^{(k)}$ from the nearest preceding generated frames and add noise to form the DMD score clip
\begin{equation}
    \begin{aligned}
    \hat{\mathbf{x}}_0^{(k)}
    &= G_{\theta}\!\left(
        \operatorname{sg}(\mathbf{x}_{\boldsymbol{\tau}}^{(k)}),
        \boldsymbol{\tau};
        \operatorname{sg}(\hat{\mathbf{H}}^{(k)})
    \right), \\
    \hat{\mathbf{z}}_0^{(k)}
    &= [
        \operatorname{sg}(\hat{\mathbf{h}}^{(k)}),
        \hat{\mathbf{x}}_0^{(k)}
    ], \qquad
    \hat{\mathbf{z}}_t^{(k)}
    = \alpha_t\operatorname{sg}(\hat{\mathbf{z}}_0^{(k)})
      + \sigma_t\boldsymbol{\epsilon}.
    \end{aligned}
    \label{eq:rolling_dmd_prediction}
\end{equation}
Here, $\operatorname{sg}$ stops gradient flow, $\boldsymbol{\epsilon}\sim\mathcal{N}(\mathbf{0},\mathbf{I})$, and the linear noising path uses $\alpha_t=1-t$ and $\sigma_t=t$. The timesteps $\boldsymbol{\tau}$ condition the generator, whereas $t$ controls the noise for score evaluation. Prompt conditioning is omitted from the notation.

DMD~\citep{dmd,dmd2} uses the difference between teacher and critic predictions on this noisy clip. The frozen teacher predicts clean latents $D_{\mathrm{real}}^\gamma$ with CFG scale $\gamma$, while the conditional critic predicts $D_{\mathrm{fake}}$ without CFG. We normalize their difference by $a_k=\max(\operatorname{mean}|\hat{\mathbf{z}}_0^{(k)}-D_{\mathrm{real}}^\gamma|,10^{-4})$, where the mean includes all latent entries in each score clip.

Let $\mathcal{I}_k$ index the supervised latent entries in the current window and $N_k=|\mathcal{I}_k|>0$. The detached update signal and local surrogate loss are
\begin{equation}
    \begin{aligned}
    \mathbf{d}_k
    &=\operatorname{sg}\!\left[
        \frac{D_{\mathrm{fake}}-D_{\mathrm{real}}^\gamma}{a_k}
    \right],\\
    \mathcal{L}_{\mathrm{DMD}}^{(k)}
    &=\frac{1}{2N_k}\sum_{j\in\mathcal{I}_k}
      \left[
        \hat{z}_{0,j}^{(k)}
        -\operatorname{sg}\!\left(\hat{z}_{0,j}^{(k)}-d_{k,j}\right)
      \right]^2.
    \end{aligned}
    \label{eq:rolling_dmd_loss}
\end{equation}
The derivative with respect to each supervised entry is $d_{k,j}/N_k$, while the history remains detached. Averaging the resulting local gradients gives the rollout update in Eq.~\ref{eq:rolling_gradient_accumulation}.

\subsection{AnnealCFG: Mitigating Long-Horizon Degradation with CFG Annealing}
\label{sec:method_annealcfg}
\begin{figure}[!t]
    \centering
    \includegraphics[width=\linewidth]{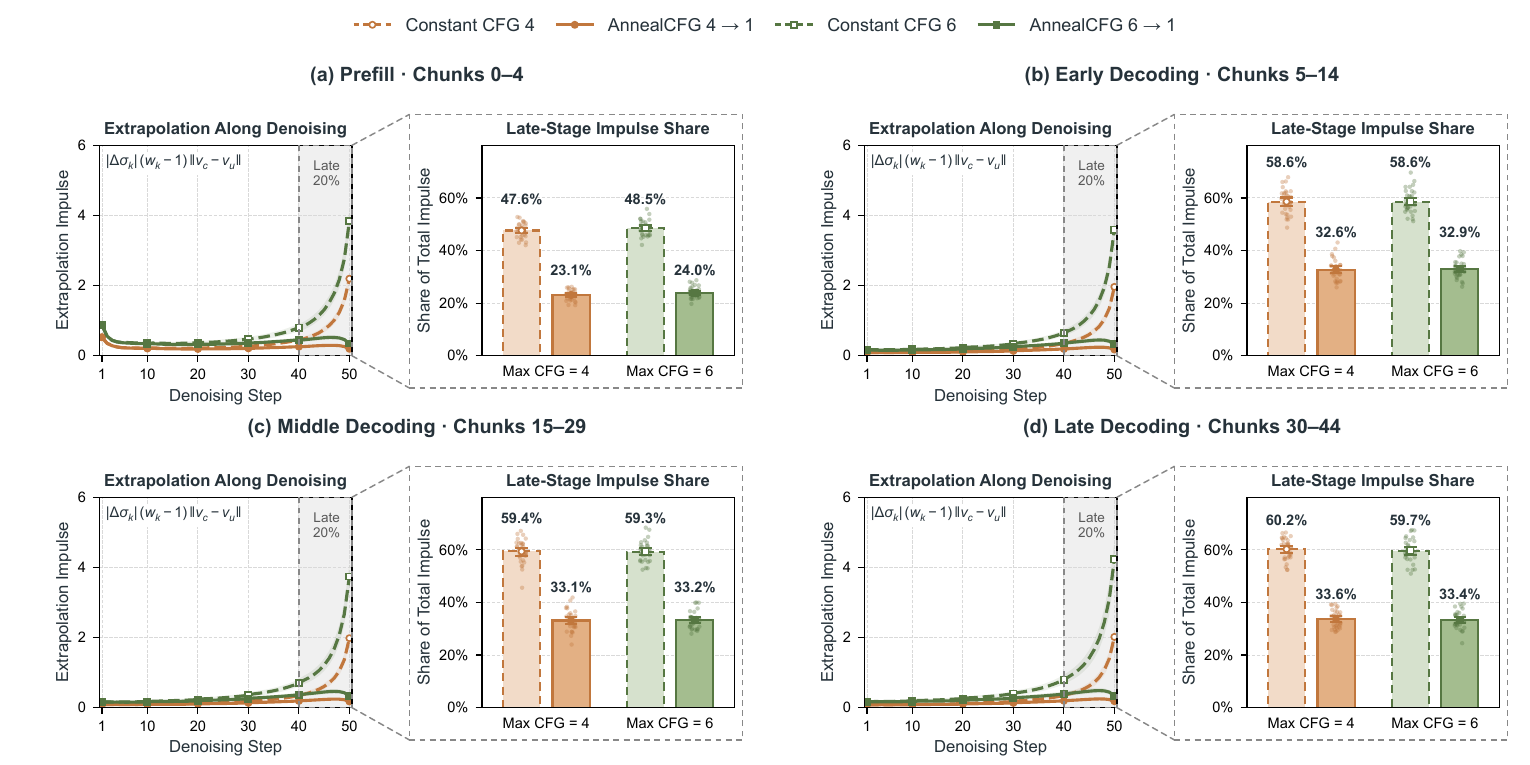}
    \caption{\textbf{AnnealCFG reduces the late-denoising extrapolation produced by constant guidance throughout 30s generation.} Across prefill and early, middle, and late decoding, the left plots show the additional latent-update magnitude induced by guidance. The right plots show its share in the final 20\% of denoising steps. Annealing lowers both the late-step peaks and their contribution to the total update. Results aggregate 32 paired prompts, with 95\% bootstrap confidence intervals shown by bands and whiskers.}
    \label{fig:method_annealcfg}
\end{figure}

Classifier-free guidance (CFG)~\citep{cfg} combines conditional and unconditional velocity predictions as
\begin{equation}
    v_{\mathrm{cfg}}
    = v_{\mathrm{uncond}} + w(t)
    \left(v_{\mathrm{cond}} - v_{\mathrm{uncond}}\right).
    \label{eq:cfg}
\end{equation}
Guidance with $w(t)>1$ can improve visual quality and semantic adherence~\citep{imagen} but may introduce a train--inference mismatch~\citep{cfg_gap} and off-manifold sampling~\citep{cfg++}. CFG-Zero*~\citep{cfg_zero_star} adjusts guidance scaling and the initial solver steps to improve flow-matching trajectories. In streaming generation, each fully denoised chunk becomes context for subsequent chunks, so deviations introduced by guidance can propagate through the generated history. We therefore reduce extrapolation near the clean endpoint before caching the chunk.

To quantify this extrapolation, let $v_{c,k,f}$ and $v_{u,k,f}$ denote the conditional and unconditional velocity predictions for latent frame $f$ at its local denoising step $k$. At normalized noise level $\sigma_k=t_k$, the guidance scale is $w_k=w(\sigma_k)\geq1$. For an Euler increment $\Delta\sigma_k$, the magnitude of the additional guided update is
\begin{equation}
    I_{k,f}
    = |\Delta\sigma_k|\,(w_k-1)
      \left\|v_{c,k,f}-v_{u,k,f}\right\|.
    \label{eq:cfg_impulse}
\end{equation}
The \emph{extrapolation impulse} $I_{k,f}$ measures the additional latent update from guidance at a fixed model input.

AnnealCFG retains strong guidance at high noise and reduces the guidance scale as denoising approaches the clean endpoint. Using the normalized timestep convention of WinTF, we set
\begin{equation}
    w(t)=1+(w_{\max}-1)t,\qquad t\in[0,1],
    \label{eq:lcat}
\end{equation}
where $w_{\max}>1$ is the maximum guidance scale. As denoising proceeds from $t=1$ to $t=0$, the guidance scale decreases from $w_{\max}$ to $1$, and the guided velocity approaches the conditional velocity prediction.

Fig.~\ref{fig:method_annealcfg} compares guidance schedules during 30s generation with 32 paired prompts and matched seeds. AnnealCFG reduces late-stage extrapolation impulses throughout generation, limiting guidance-induced deviations as chunks approach the clean endpoint.

\subsection{Threshold-Triggered KV Recomputation for Extrapolation}
\label{sec:method_recompute}
\begin{figure}[!t]
    \centering
    \includegraphics[width=\linewidth]{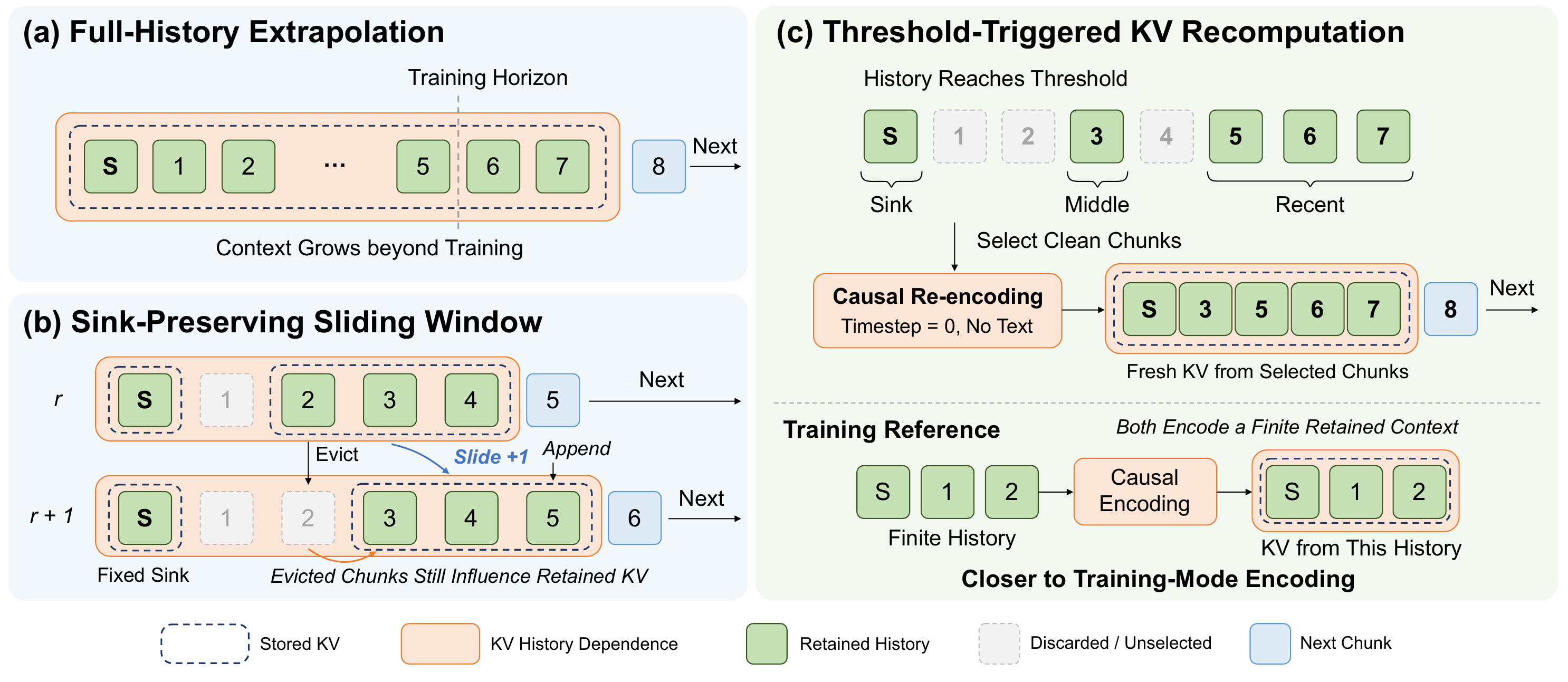}
    \caption{\textbf{KV recomputation refreshes historical context for generation beyond the training horizon.} Full-history caching expands the context as video length grows, while sliding-window caching retains dependencies on evicted chunks. At the history threshold, recomputation selects sink, intermediate, and recent chunks and causally re-encodes them into a fresh cache. This bounds the retained context and removes cached dependencies on discarded chunks before generation continues.}
    \label{fig:method_recompute}
\end{figure}

Streaming video models are trained with finite historical contexts but often need to generate beyond that horizon. Retaining all KV states increases memory usage and computational cost while exposing the model to longer contexts than those seen during training. Sliding-window extrapolation limits KV cache growth, but retained KV states may still depend on previously attended chunks that have since been evicted~\citep{krea}, as illustrated in Fig.~\ref{fig:method_recompute}.

When history reaches a configured threshold, we retain the sink chunk, representative event-middle chunks, and recent chunks. A causal-attention forward pass re-encodes them in temporal order at $t=0$, without text conditioning. Generation resumes from this rebuilt cache, bounding storage and removing cached dependencies on discarded chunks. Event-prompt switches also trigger recomputation.
 
\subsection{Streaming Video Prompt Rewriting Skill}
\label{sec:method_rewrite_skill}

Streaming generation requires sustained actions and consistent entities across prompts. Instructions with natural endpoints, such as a zoom-in, can be difficult to sustain over long sequences.

Our Streaming Video Prompt Rewriting Skill expands short user instructions into sustained actions and consistent scene descriptions. For single-event generation, it develops the action within one scene. For multi-event generation, it uses preceding rewritten prompts to maintain continuity. See App.~\ref{app:prompt_rewriting} for details.

\section{Infrastructure}
\label{sec:infra}

\subsection{FBCA: Flash Block-wise Causal Attention}

\begin{figure}[!t]
    \centering
    \includegraphics[width=\linewidth]{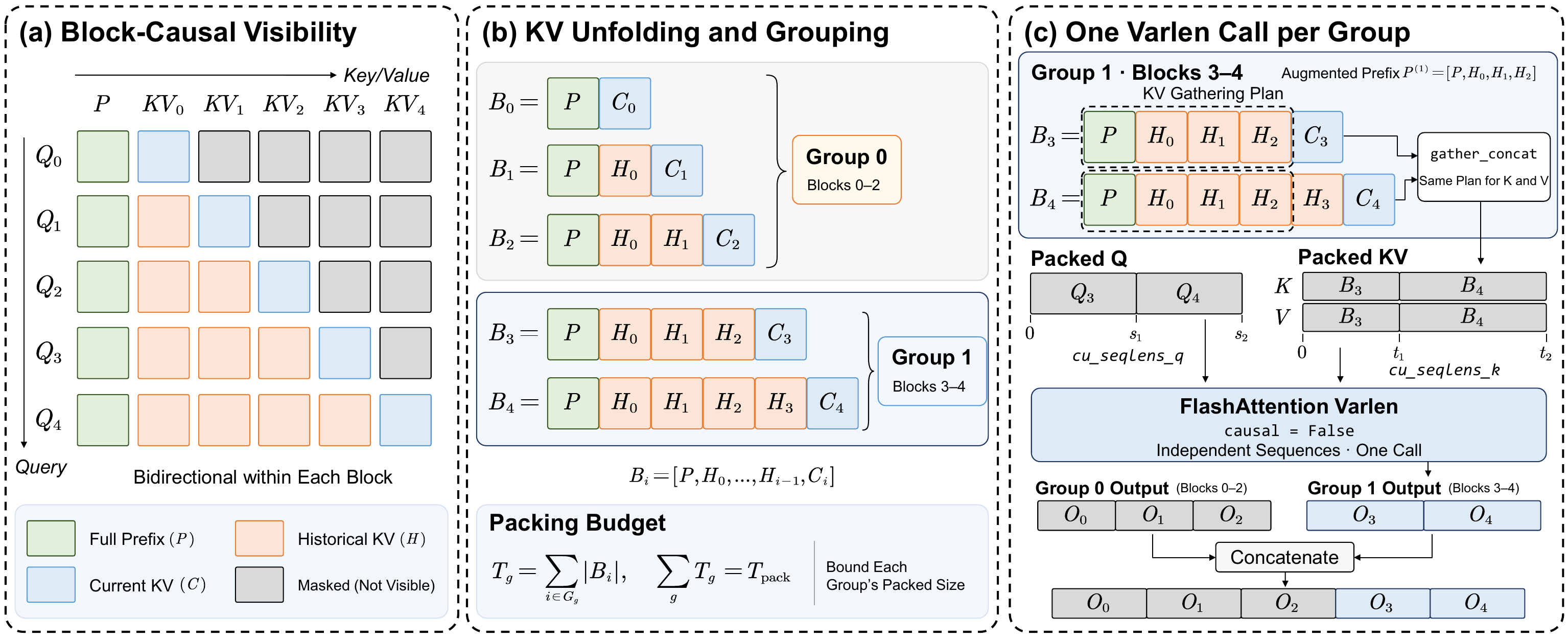}
    \caption{\textbf{Flash Block-wise Causal Attention (FBCA) implements block-causal attention with efficient FlashAttention kernels.} Each query block gathers its visible prefix, historical states, and current states into an independent attention sequence. KV unfolding and grouping organize these sequences within a per-call packing budget, and variable-length FlashAttention processes them before restoring the original block order. This preserves the required attention pattern while controlling the size of each packed buffer.}
    \label{fig:infra_flashCBA}
\end{figure}

WinTF uses a structured attention mask to connect each noisy window to its preceding history while isolating it from other noisy windows. FBCA implements this structure by gathering the keys and values visible to each query block for FlashAttention~\citep{flash_attn2,flash_attn3}, as shown in Fig.~\ref{fig:infra_flashCBA}. Each attention block corresponds to one training window of $m_i$ chunks.

Let $N_{\mathrm{blk}}$ denote the number of attention blocks. Block $i$ contains video queries $\mathbf{Q}_i^v$ and current keys and values $\mathbf{K}_i^v,\mathbf{V}_i^v$. The shared prefix consists of $\mathbf{K}_{\mathrm{full}}^v,\mathbf{V}_{\mathrm{full}}^v$, while $\mathbf{K}_{\mathrm{hist},j}^v,\mathbf{V}_{\mathrm{hist},j}^v$ denote the historical states of block $j$. Providing historical and current keys and values separately allows noisy training windows to attend to clean history.
The video key and value sequences visible to block $i$ share the form
\begin{equation}
    \overline{\mathbf{X}}_i^v
    =\operatorname{concat}\!\left(
        \mathbf{X}_{\mathrm{full}}^v,
        \mathbf{X}_{\mathrm{hist},0}^v,\ldots,
        \mathbf{X}_{\mathrm{hist},i-1}^v,
        \mathbf{X}_i^v
    \right),
    \qquad \mathbf{X}\in\{\mathbf{K},\mathbf{V}\}.
    \label{eq:fbca-visible-sequences}
\end{equation}

Each query in block $i$ attends to this same visible sequence, so key selection encodes the block-causal mask. Let $\mathbf{M}_i$ assign zero to visible source positions and $-\infty$ elsewhere. Then
\begin{equation}
    \operatorname{Attn}\left(\mathbf{Q}_i^v,
    \overline{\mathbf{K}}_i^v,\overline{\mathbf{V}}_i^v\right)
    =
    \operatorname{Attn}\left(\mathbf{Q}_i^v,
    \mathbf{K}_{\mathrm{src}}^v,\mathbf{V}_{\mathrm{src}}^v;\mathbf{M}_i\right).
\label{eq:fbca-mask-equivalence}
\end{equation}
Both expressions normalize over the same keys and aggregate the same values, preserving block-wise attention.

FBCA packs the visible sequences into a contiguous buffer, and variable-length FlashAttention evaluates each query--key sequence pair independently. In multimodal attention, video queries also attend to the valid text tokens in their window, while text queries attend only within that window. App.~\ref{app:fbca_details} describes how these sequences are packed and their boundaries are recorded for attention.

\begin{table}[!t]
    \centering
    \caption{\textbf{Efficiency comparison of block-causal attention implementations.} The comparison evaluates full-history attention with three or five windows on one NVIDIA H20 GPU. FBCA achieves the lowest forward and backward runtimes across fixed and varying input shapes, with preparation taking less than 0.1s when shapes vary. In multiple-shape tests, it also reduces peak memory relative to FlexAttention, supporting efficient training across inputs with different spatial resolutions.}
    \label{tab:fbca_efficiency}
    \fontsize{9}{11.4}\selectfont
    \setlength{\tabcolsep}{3.5pt}
    \renewcommand{\arraystretch}{1.2}
    \begin{tabular}{cccccccc}
        \toprule
        \multirow{3}{*}[-1pt]{\tablehead{Windows}}
        & \multirow{3}{*}[-1pt]{\tablehead{Implementation}}
        & \multicolumn{3}{c}{\tablehead{Single Shape}}
        & \multicolumn{3}{c}{\tablehead{Multiple Shapes}} \\
        \cmidrule(lr){3-5}\cmidrule(l){6-8}
        & & Forward & Fwd.+Bwd. & Mem Peak
        & Preparation & Total Time & Mem Peak \\
        & & (ms) $\downarrow$ & (ms) $\downarrow$ & (GiB) $\downarrow$
        & (s) $\downarrow$ & (s) $\downarrow$ & (GiB) $\downarrow$ \\
        \midrule
        \multirow{3}{*}{3}
        & SDPA (cuDNN) & 1678.64 & 6802.64 & 64.19 & 41.76 & 249.83 & 62.14 \\
        & FlexAttention & 699.81 & 2789.95 & \tablenum{10.37} & 57.48 & 141.73 & 28.61 \\
        \rowcolor{MovieRollColor}
        \cellcolor{white}
        & \tablenum{FBCA} & \tablenum{541.44} & \tablenum{1878.47} & 13.35
        & $<0.10$ & \tablenum{56.82} & \tablenum{13.19} \\
        \midrule
        \multirow{3}{*}{5}
        & SDPA (cuDNN) & & OOM & & & OOM & \\
        & FlexAttention & 1891.92 & 7644.83 & \tablenum{18.50} & 138.58 & 369.94 & 87.13 \\
        \rowcolor{MovieRollColor}
        \cellcolor{white}
        & \tablenum{FBCA} & \tablenum{1491.18} & \tablenum{5204.46} & 31.51
        & $<0.10$ & \tablenum{156.94} & \tablenum{31.12} \\
        \bottomrule
    \end{tabular}
\end{table}

Packing duplicates shared historical states. For equal-sized blocks containing $T_{\mathrm{blk}}$ video tokens and a full prefix containing $P$ tokens, the packed key/value length per sample is
\begin{equation}
    T_{\mathrm{pack}}
    =
    \sum_{i=0}^{N_{\mathrm{blk}}-1}\bigl(P+(i+1)T_{\mathrm{blk}}\bigr)
    =
    N_{\mathrm{blk}}P+\frac{N_{\mathrm{blk}}(N_{\mathrm{blk}}+1)}{2}T_{\mathrm{blk}}.
\label{eq:fbca-packed-length}
\end{equation}
The packed buffer grows quadratically with the number of blocks. FBCA therefore groups consecutive blocks and supplies each group with the preceding history as an augmented prefix. This bounds the buffer for each attention call while preserving the keys visible to each query and the total packed length. App.~\ref{app:fbca_details} gives the grouping and memory analysis.

FBCA achieves the lowest measured forward and training times in Tab.~\ref{tab:fbca_efficiency}. Its preparation cost remains below 0.1s across multiple input shapes, avoiding the repeated kernel-compilation overhead of FlexAttention~\citep{flex_attn}. These gains support efficient mixed-resolution training.

\subsection{PS-VAE: Parallel Streaming Variational Autoencoder}
\label{sec:infra_PSVAE}
\begin{figure}[!t]
    \centering
    \includegraphics[width=\linewidth]{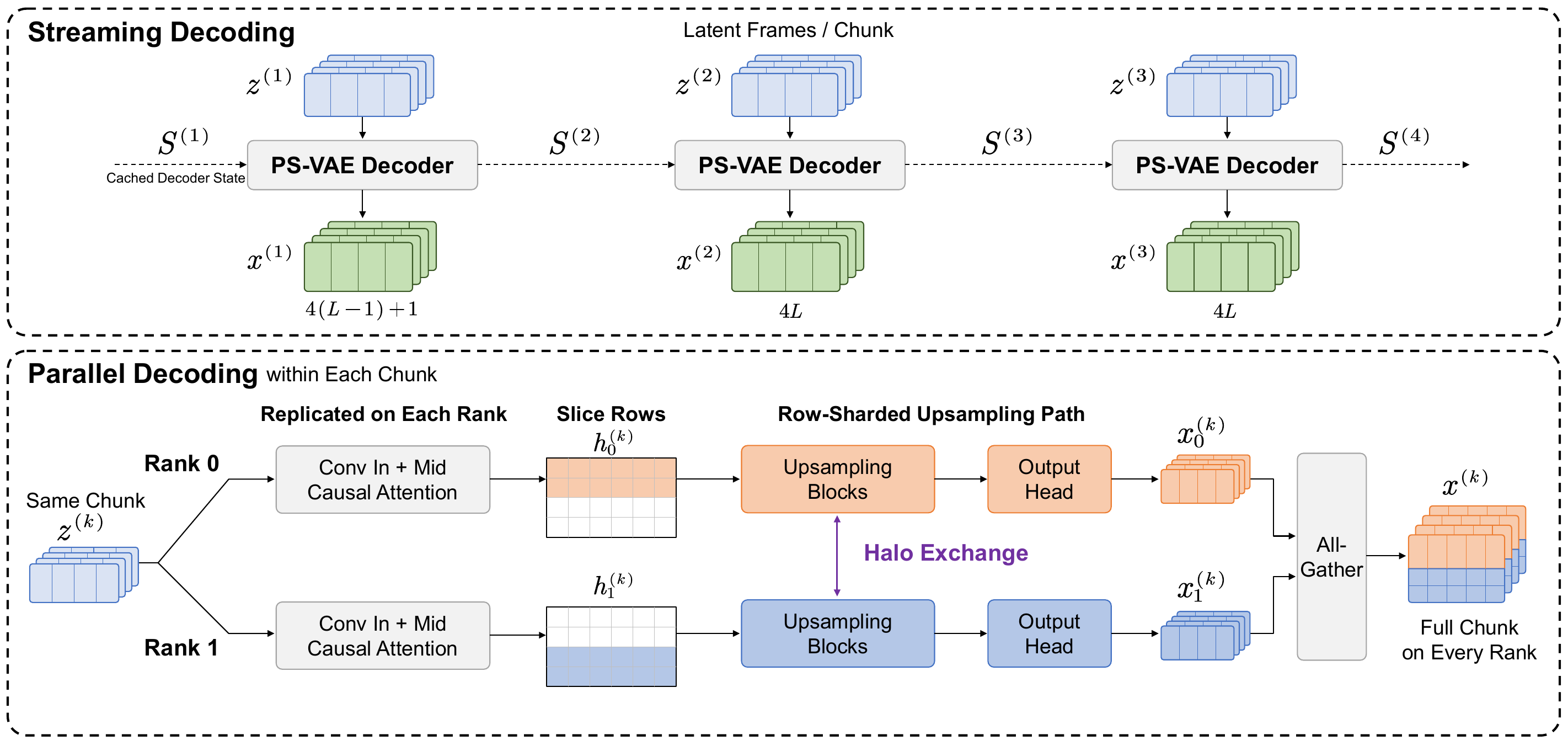}
    \caption{\textbf{The Parallel Streaming Variational Autoencoder (PS-VAE) supports continuous video decoding through temporal caching and spatial parallelism.} The upper panel shows how cached decoder states carry temporal context across arriving latent chunks, allowing each chunk to be decoded as it becomes available. The lower panel distributes high-resolution decoding across GPUs through row partitioning and boundary exchange. Together, these designs support streaming output and reduce per-GPU decoding cost.}
    \label{fig:infra_ps_vae}
\end{figure}

A streaming denoiser produces latent chunks before the full sequence is available. Waiting for all chunks delays the first decoded frame, while decoding each chunk independently loses temporal context from the VAE. PS-VAE preserves temporal state across chunks and distributes local spatial operations across devices, as illustrated in Fig.~\ref{fig:infra_ps_vae}.

Given latent chunk $z^{(k)}$ and decoder state $\mathcal{S}^{(k)}$, decoding and state updates are
\[
\begin{aligned}
x^{(k)}
    &= \mathcal{D}_{\mathrm{stream}}\left(z^{(k)};\mathcal{S}^{(k)}\right),\\
\mathcal{S}^{(k+1)}
    &= \mathcal{U}\left(\mathcal{S}^{(k)},z^{(k)}\right).
\end{aligned}
\]
The decoded chunk is emitted immediately. The decoder state $\mathcal{S}^{(k)}$ stores input suffixes for causal convolutions, first-frame indicators for temporal upsampling, and projected keys and values for bottleneck attention.

Temporal caches preserve convolution inputs and track the first-frame rule in temporal upsampling. An initial chunk of $L$ latent frames produces $4(L-1)+1$ pixel frames, while each subsequent chunk produces $4L$. Bottleneck attention is causal across frames and bidirectional within each frame. With full attention history, these states preserve the temporal inputs and visible keys used by full-sequence decoding.

For spatial parallelism, each rank computes the input convolution and middle block on the same latent chunk, then decodes a contiguous row band through the upsampling blocks. Neighboring ranks exchange boundary rows for local convolutions, and an all-gather reconstructs each decoded chunk. This distributes high-resolution computation while preserving single-device streaming decoding under the same cache policy, up to numerical differences. State handling and spatial decomposition are detailed in App.~\ref{app:psvae_details}.

Tab.~\ref{tab:psvae_efficiency} shows that PS-VAE enables long-video decoding when full-sequence decoding runs out of memory. On a single GPU, PS-VAE is faster and achieves better reconstruction quality than the tested tiling methods. Spatial parallelism further improves throughput and reduces per-GPU memory use, while maintaining reconstruction quality and close numerical agreement with the single-GPU output.

\begin{table}[!t]
    \centering
    \caption{\textbf{Comparison of video decoding speed, memory use, and reconstruction quality.} Single-GPU PS-VAE outperforms the tested tiling methods in runtime and reconstruction PSNR. Spatial parallelism further reduces decoding time and per-GPU memory. With eight GPUs, PS-VAE achieves a 3.97$\times$ decoding speedup over its single-GPU version while preserving high numerical agreement. Measurements decode 1441 frames at $480\times864$ on NVIDIA H20 GPUs in FP16.}
    \label{tab:psvae_efficiency}
    \fontsize{9.5}{11.4}\selectfont
    \setlength{\tabcolsep}{2.4pt}
    \renewcommand{\arraystretch}{1.15}
    \begin{tabular}{@{}lccccccc@{}}
        \toprule
        \multirow{2}{*}{\tablehead{Method}}
        & \multirow{2}{*}{\tablehead{GPUs}}
        & \tablehead{Time}
        & \tablehead{Throughput}
        & \multirow{2}{*}{\tablehead{Speedup} $\uparrow$}
        & \tablehead{Memory}
        & \tablehead{Reconstruction}
        & \tablehead{Agreement} \\
        & & (s) $\downarrow$ & (FPS) $\uparrow$ & & (GiB) $\downarrow$
        & \tablehead{PSNR} (dB) $\uparrow$
        & \tablehead{PSNR} (dB) $\uparrow$ \\
        \midrule
        VAE (Full Sequence)       & 1 & OOM             & --    & --            & --   & --    & --    \\
        Temporal Tiling (32 f)   & 1 & 219.61           & 6.56  & $0.80\times$  & 37.11 & 38.43 & 42.82 \\
        Spatial Tiling (64 px)   & 1 & 273.90           & 5.26  & $0.64\times$  & 20.67 & 38.69 & 43.20 \\
        \begin{tabular}[c]{@{}l@{}}Temporal + Spatial Tiling\\(64 f, 256 px)\end{tabular} & 1 & 355.11          & 4.06  & $0.50\times$  & 16.15 & 39.91 & 49.15 \\
        \midrule
        PS-VAE                  & 1 & 176.15           & 8.18  & $1.00\times$  & 28.84 & \tablenum{40.46} & Ref. \\
        PS-VAE                  & 2 & 101.96           & 14.13 & $1.73\times$  & 18.46 & \tablenum{40.46} & 86.21 \\
        PS-VAE                  & 4 & 64.79            & 22.24 & $2.72\times$  & 15.51 & \tablenum{40.46} & 86.22 \\
        PS-VAE                  & 8 & \tablenum{44.37}   & \tablenum{32.47} & $\mathbf{3.97\times}$ & \tablenum{14.83} & \tablenum{40.46} & 86.22 \\
        \bottomrule
    \end{tabular}
\end{table}

\section{Experiment}
\label{sec:experiment}

\subsection{Implementation Details}
\label{sec:experiment_implementation_detail}

MovieRoll-Base is initialized from HunyuanVideo-1.5-480P-T2V~\citep{hunyuanvideo1.5} with 6 full-history and 48 sink--recent blocks. During decoding, the rolling window contains five chunks of four latent frames each. During streaming fine-tuning, we perturb clean histories with probability $p=0.9$ and noise level $\sigma\sim\mathcal{U}(0,0.15)$. MovieRoll-Distilled undergoes ODE initialization followed by CuLT-DMD on single- and two-event prompts. All stages use Muon~\citep{muon}. At inference, MovieRoll-Base uses 50 steps with CFG annealed from 4 to 1, while MovieRoll-Distilled uses five steps without CFG. Both use an Euler solver with a timestep shift of 5 and a KV recomputation threshold of 18 chunks. App.~\ref{app:implementation} provides the full configurations.

We evaluate 200 randomly sampled single-event prompts from MovieGen~\citep{movie_gen} and VBench~\citep{vbench,vbench2} category sets. Multi-event evaluation uses 100 MemFlow~\citep{memflow} prompts. We use our Streaming Video Prompt Rewriting Skill to rewrite all single-event and multi-event evaluation prompts.

Video evaluation spans visual quality, temporal consistency, and prompt alignment~\citep{vbench_pp}. We report all HeliosBench metrics~\citep{helios}. The evaluators use CLIP~\citep{clip} for aesthetics, Farneb{\"a}ck optical flow~\citep{farneback} for dynamics, AMT-S~\citep{amt} for motion smoothness, and ViCLIP~\citep{viclip} for semantic alignment. Naturalness is evaluated with three Gemini-2.5-Pro~\citep{gemini2.5} queries per video. Multi-event semantic and drifting-semantic scores are averaged across events. App.~\ref{app:heliosbench} details the evaluation protocol.

\subsection{Main Comparison}
\label{sec:experiment_main_comparison}

\subsubsection{Single-Event Comparison}

\begin{figure}[!t]
    \centering
    \includegraphics[width=\linewidth]{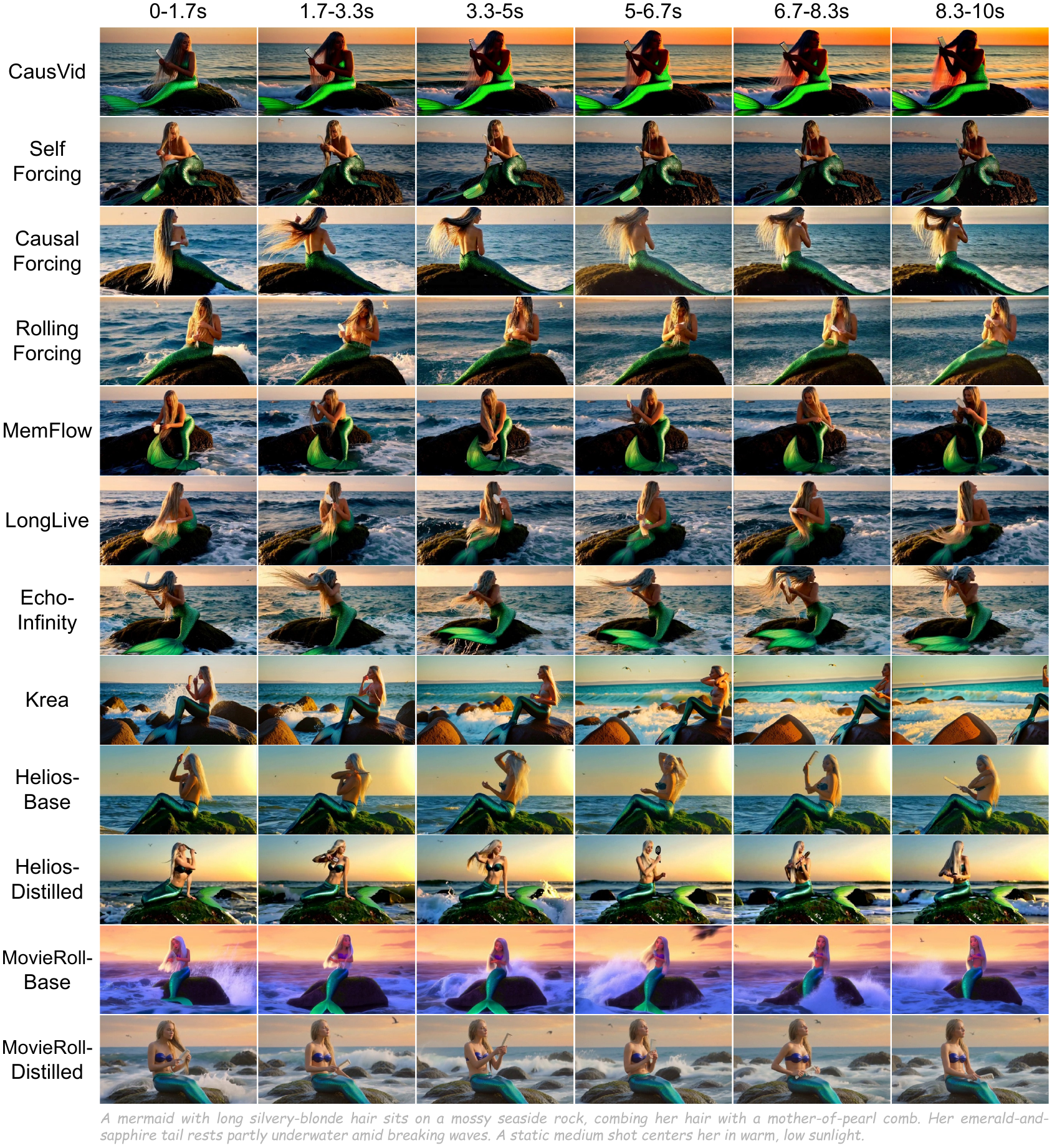}
    \caption{\textbf{Qualitative comparison of 10s single-event video generation.} Under the same prompt of a mermaid combing her hair on a seaside rock, MovieRoll-Base and MovieRoll-Distilled retain recognizable subjects and consistent surroundings as poses and waves change. Rows compare methods evaluated under the same prompt, and columns show successive temporal intervals within each 241-frame video.}
    \label{fig:experiment_single_event_241f}

\end{figure}

\begin{table*}[!t]

    \centering
    \caption{\textbf{Quantitative comparison of single-event video generation at 10, 30, and 60s.} MovieRoll-Distilled achieves the highest total HeliosBench score at all three durations, maintaining a naturalness rating of 10 while sustaining active motion. MovieRoll-Base ranks second in total score at both 10s and 30s. Higher metric ratings indicate better performance, including less temporal drift.}
    \label{tab:experiment_single_event}

    \small
    \setlength{\tabcolsep}{2.2pt}
    \renewcommand{\arraystretch}{1.08}

    \resizebox{\textwidth}{!}{%
        \begin{tabular}{@{}l*{10}{c}@{}}
            \toprule

            \tablehead{Method}
            & \shortstack{\tablehead{Aesthetic}\\$\uparrow$}
            & \shortstack{\tablehead{Dynamic}\\$\uparrow$}
            & \shortstack{\tablehead{Smoothness}\\$\uparrow$}
            & \shortstack{\tablehead{Semantic}\\$\uparrow$}
            & \shortstack{\tablehead{Naturalness}\\$\uparrow$}
            & \shortstack{\tablehead{Drifting}\\\tablehead{Aesthetic} $\uparrow$}
            & \shortstack{\tablehead{Drifting}\\\tablehead{Smoothness} $\uparrow$}
            & \shortstack{\tablehead{Drifting}\\\tablehead{Semantic} $\uparrow$}
            & \shortstack{\tablehead{Drifting}\\\tablehead{Naturalness} $\uparrow$}
            & \shortstack{\tablehead{Total}\\\tablehead{Score} $\uparrow$}
            \\

            \midrule

            \multicolumn{11}{c}{\textit{241-Frame Evaluation}} \\
            \midrule

            CausVid
            & 8       
            & 4       
            & 9       
            & 4       
            & 1       
            & 7       
            & 10      
            & 8       
            & 2       
            & 4.578   
            \\

            Self Forcing
            & 8       
            & 4       
            & 10      
            & 5       
            & 4       
            & 7       
            & 10      
            & 9       
            & 1       
            & 5.628   
            \\

            Causal Forcing
            & 8       
            & 10      
            & 9       
            & 5       
            & 4       
            & 6       
            & 10      
            & 8       
            & 1       
            & 5.580   
            \\

            Rolling Forcing
            & 8       
            & 2       
            & 10      
            & 5       
            & 7       
            & 8       
            & 10      
            & 10      
            & 2       
            & 6.630   
            \\

            MemFlow
            & 8       
            & 4       
            & 10      
            & 5       
            & 8       
            & 8       
            & 10      
            & 10      
            & 4       
            & 7.143   
            \\

            LongLive
            & 8       
            & 3       
            & 10      
            & 5       
            & 7       
            & 8       
            & 10      
            & 9       
            & 4       
            & 6.759   
            \\

            Echo-Infinity
            & 8       
            & 5       
            & 9       
            & 5       
            & 6       
            & 7       
            & 10      
            & 9       
            & 2       
            & 6.237   
            \\

            Krea
            & 8       
            & 10      
            & 9       
            & 6       
            & 1       
            & 4       
            & 10      
            & 7       
            & 1       
            & 4.773   
            \\

            Helios-Base
            & 8       
            & 4       
            & 10      
            & 5       
            & 8       
            & 8       
            & 10      
            & 9       
            & 5       
            & 7.143   
            \\

            Helios-Distilled
            & 8       
            & 5       
            & 9       
            & 5       
            & 8       
            & 8       
            & 10      
            & 9       
            & 4       
            & 7.044   
            \\

            \addlinespace[1.5pt]

            \rowcolor{MovieRollColor}
            \tablenum{MovieRoll-Base}
            & 8       
            & 5       
            & 10      
            & 5       
            & 10      
            & 7       
            & 10      
            & 9       
            & 4       
            & 7.485   
            \\

            \rowcolor{MovieRollColor}
            \tablenum{MovieRoll-Distilled}
            & 8       
            & 6       
            & 10      
            & 5       
            & 10      
            & 6       
            & 10      
            & 8       
            & 6       
            & 7.515   
            \\

            \midrule

            \multicolumn{11}{c}{\textit{721-Frame Evaluation}} \\
            \midrule

            CausVid
            & 8       
            & 4       
            & 9       
            & 4       
            & 1       
            & 6       
            & 10      
            & 8       
            & 6       
            & 4.875   
            \\

            Self Forcing
            & 7       
            & 3       
            & 9       
            & 5       
            & 1       
            & 1       
            & 10      
            & 5       
            & 1       
            & 3.783   
            \\

            Causal Forcing
            & 7       
            & 6       
            & 9       
            & 5       
            & 1       
            & 5       
            & 10      
            & 8       
            & 1       
            & 4.566   
            \\

            Rolling Forcing
            & 8       
            & 2       
            & 10      
            & 5       
            & 4       
            & 7       
            & 10      
            & 9       
            & 1       
            & 5.568   
            \\

            MemFlow
            & 8       
            & 5       
            & 10      
            & 5       
            & 6       
            & 8       
            & 10      
            & 10      
            & 5       
            & 6.762   
            \\

            LongLive
            & 8       
            & 3       
            & 10      
            & 5       
            & 5       
            & 8       
            & 10      
            & 10      
            & 5       
            & 6.447   
            \\

            Echo-Infinity
            & 8       
            & 5       
            & 9       
            & 5       
            & 4       
            & 5       
            & 10      
            & 8       
            & 2       
            & 5.430   
            \\

            Krea
            & 7       
            & 10      
            & 9       
            & 5       
            & 1       
            & 1       
            & 10      
            & 5       
            & 1       
            & 3.993   
            \\

            Helios-Base
            & 8       
            & 4       
            & 10      
            & 5       
            & 8       
            & 7       
            & 10      
            & 9       
            & 2       
            & 6.747   
            \\

            Helios-Distilled
            & 8       
            & 6       
            & 9       
            & 5       
            & 4       
            & 4       
            & 10      
            & 7       
            & 1       
            & 5.163   
            \\

            \addlinespace[1.5pt]

            \rowcolor{MovieRollColor}
            \tablenum{MovieRoll-Base}
            & 8       
            & 4       
            & 10      
            & 5       
            & 9       
            & 6       
            & 10      
            & 9       
            & 2       
            & 6.903   
            \\

            \rowcolor{MovieRollColor}
            \tablenum{MovieRoll-Distilled}
            & 8       
            & 6       
            & 10      
            & 4       
            & 10      
            & 6       
            & 10      
            & 8       
            & 6       
            & 7.260   
            \\

            \midrule

            \multicolumn{11}{c}{\textit{1441-Frame Evaluation}} \\
            \midrule

            CausVid
            & 8       
            & 3       
            & 9       
            & 3       
            & 1       
            & 6       
            & 10      
            & 7       
            & 8       
            & 4.689   
            \\

            Self Forcing
            & 7       
            & 3       
            & 9       
            & 3       
            & 1       
            & 1       
            & 10      
            & 1       
            & 1       
            & 2.877   
            \\

            Causal Forcing
            & 7       
            & 5       
            & 9       
            & 5       
            & 1       
            & 3       
            & 10      
            & 7       
            & 1       
            & 4.239   
            \\

            Rolling Forcing
            & 8       
            & 2       
            & 10      
            & 5       
            & 1       
            & 5       
            & 10      
            & 8       
            & 1       
            & 4.506   
            \\

            MemFlow
            & 8       
            & 4       
            & 10      
            & 5       
            & 4       
            & 8       
            & 10      
            & 10      
            & 2       
            & 5.925   
            \\

            LongLive
            & 8       
            & 3       
            & 10      
            & 5       
            & 4       
            & 9       
            & 10      
            & 10      
            & 3       
            & 6.093   
            \\

            Echo-Infinity
            & 7       
            & 6       
            & 9       
            & 5       
            & 4       
            & 5       
            & 10      
            & 8       
            & 3       
            & 5.529   
            \\

            Krea
            & 7       
            & 10      
            & 9       
            & 4       
            & 1       
            & 1       
            & 10      
            & 4       
            & 2       
            & 3.738   
            \\

            Helios-Base
            & 8       
            & 4       
            & 10      
            & 6       
            & 6       
            & 6       
            & 10      
            & 9       
            & 2       
            & 6.393   
            \\

            Helios-Distilled
            & 7       
            & 6       
            & 9       
            & 4       
            & 3       
            & 1       
            & 10      
            & 5       
            & 1       
            & 4.128   
            \\

            \addlinespace[1.5pt]

            \rowcolor{MovieRollColor}
            \tablenum{MovieRoll-Distilled}
            & 7       
            & 6       
            & 10      
            & 4       
            & 10      
            & 6       
            & 10      
            & 8       
            & 4       
            & 7.032   
            \\

            \bottomrule
        \end{tabular}%
    }

\end{table*}

\begin{figure}[!t]
    \centering
    \includegraphics[width=\linewidth]{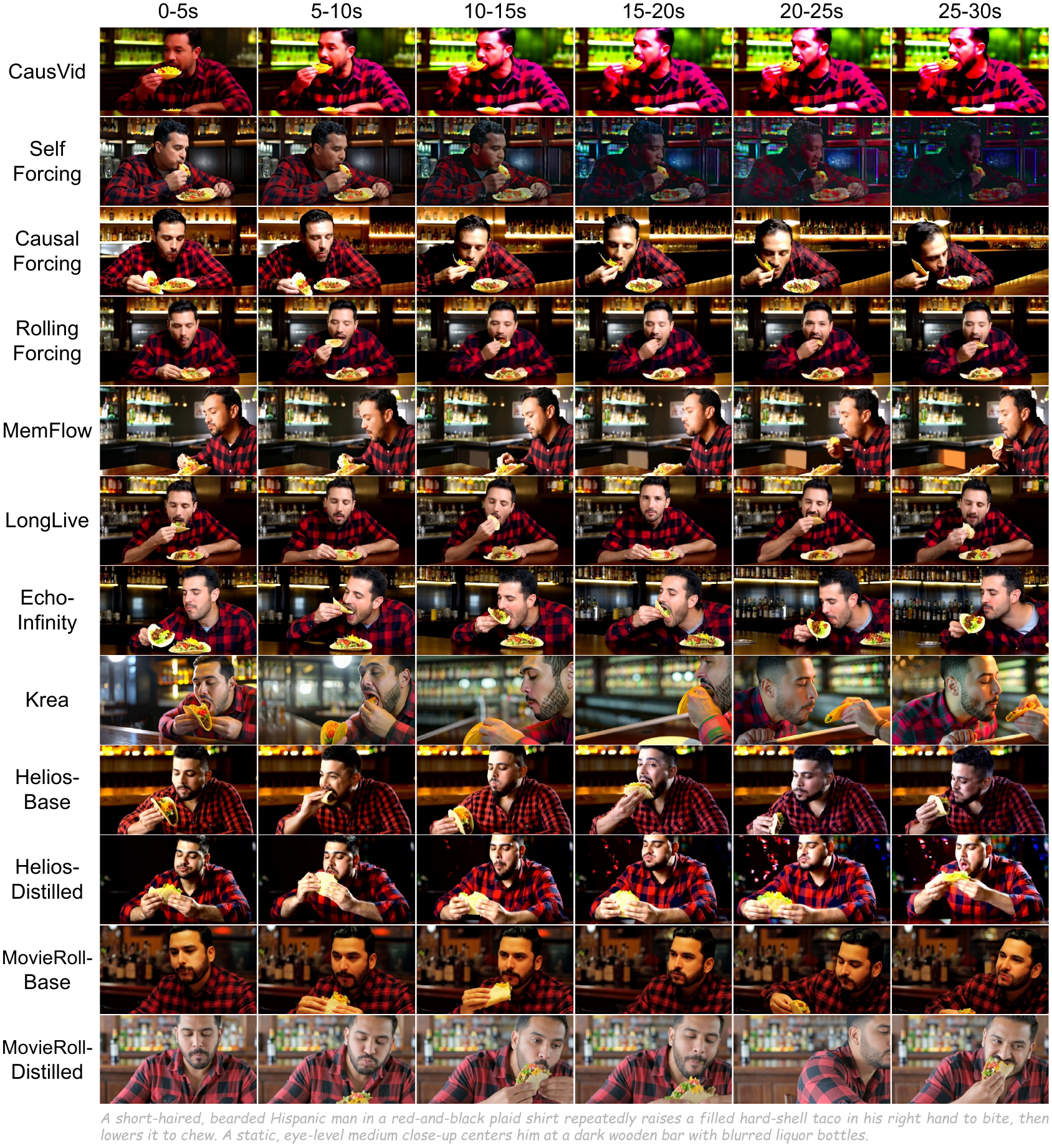}
    \caption{\textbf{Qualitative comparison of 30s single-event video generation.} MovieRoll-Base and MovieRoll-Distilled preserve the facial features, clothing, and surroundings of a man eating at a bar as the action continues. Rows compare methods under the same prompt, and columns follow the 721-frame sequence, showing how subject and background consistency evolve during generation.}
    \label{fig:experiment_single_event}

\end{figure}

\begin{figure}[!t]
    \centering
    \includegraphics[width=\linewidth]{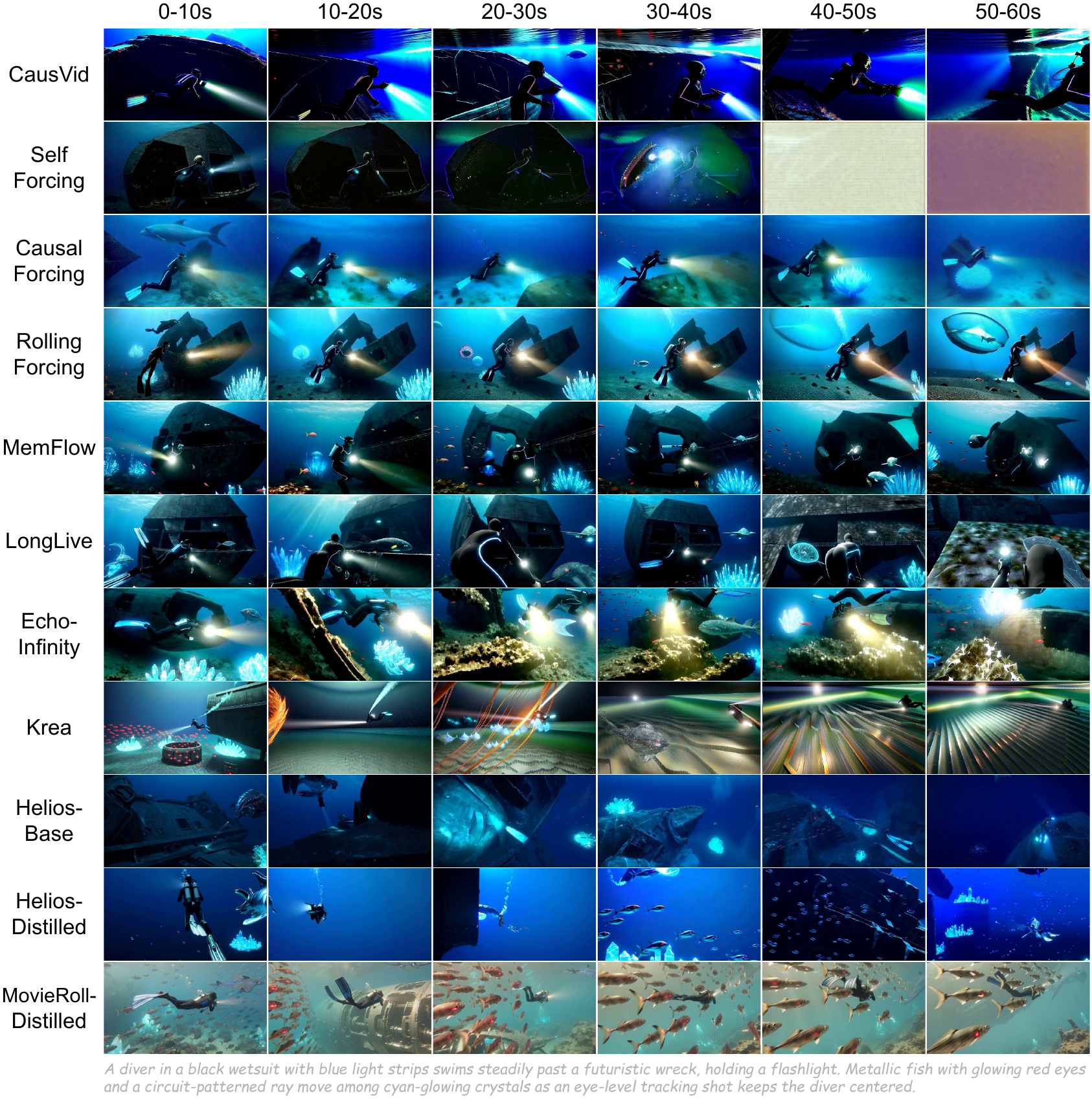}
    \caption{\textbf{Qualitative comparison of 60s single-event video generation.} MovieRoll-Distilled preserves the diver and underwater environment with relatively stable colors as generation continues. Rows compare methods under the same diver-and-wreck prompt, and columns follow successive 10s intervals of the 1441-frame videos.}
    \label{fig:experiment_single_event_1441f}

\end{figure}

We compare MovieRoll with CausVid~\citep{causvid}, Self Forcing~\citep{self_forcing}, Causal Forcing~\citep{causal_forcing}, Rolling Forcing~\citep{rolling_forcing}, MemFlow~\citep{memflow}, LongLive~\citep{longlive}, Echo-Infinity~\citep{echo_infinity}, Krea~\citep{krea}, and the two Helios variants~\citep{helios} at 10s (241 frames), 30s (721 frames), and 60s (1441 frames). At 60s, we evaluate only MovieRoll-Distilled because MovieRoll-Base remains insufficiently stable.

\textbf{Quantitative Results.} MovieRoll-Distilled achieves the highest total score at every duration, as shown in Tab.~\ref{tab:experiment_single_event}. Its naturalness rating remains stable as video length increases, and its total score declines more gradually than those of Self Forcing and Helios-Distilled. It also exceeds Helios-Base in dynamics and matches or exceeds Helios-Distilled, combining sustained naturalness with active motion. MovieRoll-Base also remains competitive on the 10s and 30s benchmarks, supporting the effectiveness of the streaming fine-tuning and inference strategies used in MovieRoll.

\textbf{Qualitative Results.} MovieRoll maintains coherent subjects and scenes across the three evaluated durations. In the 10s example shown in Fig.~\ref{fig:experiment_single_event_241f}, both MovieRoll variants preserve the mermaid and seaside setting, as do LongLive and the Helios variants. At 30s, both MovieRoll variants retain consistent subject appearance and the bar setting, as illustrated in Fig.~\ref{fig:experiment_single_event}. LongLive also remains coherent, whereas Helios-Distilled shows increasing color saturation. At 60s, MovieRoll-Distilled preserves the diver and underwater environment with relatively stable colors, whereas LongLive shifts toward close-up views of the wreck and the Helios variants produce darker or more saturated scenes, as shown in Fig.~\ref{fig:experiment_single_event_1441f}. These comparisons complement the sustained naturalness ratings reported in Tab.~\ref{tab:experiment_single_event}.

\subsubsection{Multi-Event Comparison}
\begin{table*}[!t]

    \centering
    \caption{\textbf{Quantitative comparison of multi-event video generation at 30 and 60s.} Videos contain three or six events, with prompt changes every 10s. MovieRoll-Distilled achieves the highest total score, dynamics rating, and drifting naturalness rating at both durations, with a naturalness rating of 10. These results show sustained naturalness and active motion across event transitions. Higher metric ratings indicate better performance, with higher drift ratings corresponding to smaller changes between the beginning and end of a video.}
    \label{tab:experiment_multi_event}

    \small
    \setlength{\tabcolsep}{2.2pt}
    \renewcommand{\arraystretch}{1.08}

    \resizebox{\textwidth}{!}{%
        \begin{tabular}{@{}l*{10}{c}@{}}
            \toprule

            \tablehead{Method}
            & \shortstack{\tablehead{Aesthetic}\\$\uparrow$}
            & \shortstack{\tablehead{Dynamic}\\$\uparrow$}
            & \shortstack{\tablehead{Smoothness}\\$\uparrow$}
            & \shortstack{\tablehead{Semantic}\\$\uparrow$}
            & \shortstack{\tablehead{Naturalness}\\$\uparrow$}
            & \shortstack{\tablehead{Drifting}\\\tablehead{Aesthetic} $\uparrow$}
            & \shortstack{\tablehead{Drifting}\\\tablehead{Smoothness} $\uparrow$}
            & \shortstack{\tablehead{Drifting}\\\tablehead{Semantic} $\uparrow$}
            & \shortstack{\tablehead{Drifting}\\\tablehead{Naturalness} $\uparrow$}
            & \shortstack{\tablehead{Total}\\\tablehead{Score} $\uparrow$}
            \\

            \midrule

            \multicolumn{11}{c}{\textit{721-Frame (3-Event) Evaluation}} \\
            \midrule

            MemFlow
            & 8       
            & 3       
            & 10      
            & 3       
            & 7       
            & 6       
            & 10      
            & 9       
            & 1       
            & 5.754   
            \\

            LongLive
            & 9       
            & 1       
            & 10      
            & 3       
            & 5       
            & 6       
            & 10      
            & 9       
            & 1       
            & 5.214   
            \\

            Echo-Infinity
            & 8       
            & 3       
            & 9       
            & 3       
            & 7       
            & 5       
            & 10      
            & 9       
            & 1       
            & 5.625   
            \\

            Helios-Base
            & 8       
            & 1       
            & 10      
            & 2       
            & 10      
            & 7       
            & 10      
            & 9       
            & 2       
            & 6.402   
            \\

            Helios-Distilled
            & 8       
            & 3       
            & 10      
            & 2       
            & 7       
            & 5       
            & 10      
            & 9       
            & 1       
            & 5.400   
            \\

            \addlinespace[1.5pt]

            \rowcolor{MovieRollColor}
            \tablenum{MovieRoll-Base}
            & 8       
            & 2       
            & 10      
            & 2       
            & 10      
            & 6       
            & 10      
            & 9       
            & 1       
            & 6.234   
            \\

            \rowcolor{MovieRollColor}
            \tablenum{MovieRoll-Distilled}
            & 8       
            & 4       
            & 10      
            & 3       
            & 10      
            & 5       
            & 10      
            & 8       
            & 8       
            & 7.044   
            \\

            \midrule

            \multicolumn{11}{c}{\textit{1441-Frame (6-Event) Evaluation}} \\
            \midrule

            MemFlow
            & 8       
            & 2       
            & 10      
            & 2       
            & 4       
            & 5       
            & 10      
            & 9       
            & 1       
            & 4.605   
            \\

            LongLive
            & 8       
            & 1       
            & 10      
            & 3       
            & 4       
            & 5       
            & 10      
            & 9       
            & 1       
            & 4.830   
            \\

            Echo-Infinity
            & 8       
            & 3       
            & 9       
            & 3       
            & 5       
            & 3       
            & 10      
            & 9       
            & 1       
            & 4.917   
            \\

            Helios-Base
            & 8       
            & 2       
            & 10      
            & 2       
            & 9       
            & 5       
            & 10      
            & 9       
            & 1       
            & 5.880   
            \\

            Helios-Distilled
            & 8       
            & 3       
            & 10      
            & 1       
            & 5       
            & 1       
            & 10      
            & 8       
            & 1       
            & 4.140   
            \\

            \addlinespace[1.5pt]

            \rowcolor{MovieRollColor}
            \tablenum{MovieRoll-Distilled}
            & 8       
            & 4       
            & 10      
            & 2       
            & 10      
            & 4       
            & 10      
            & 8       
            & 6       
            & 6.492   
            \\

            \bottomrule
        \end{tabular}%
    }

\end{table*}

\begin{figure}[!t]
    \centering
    \includegraphics[width=\linewidth]{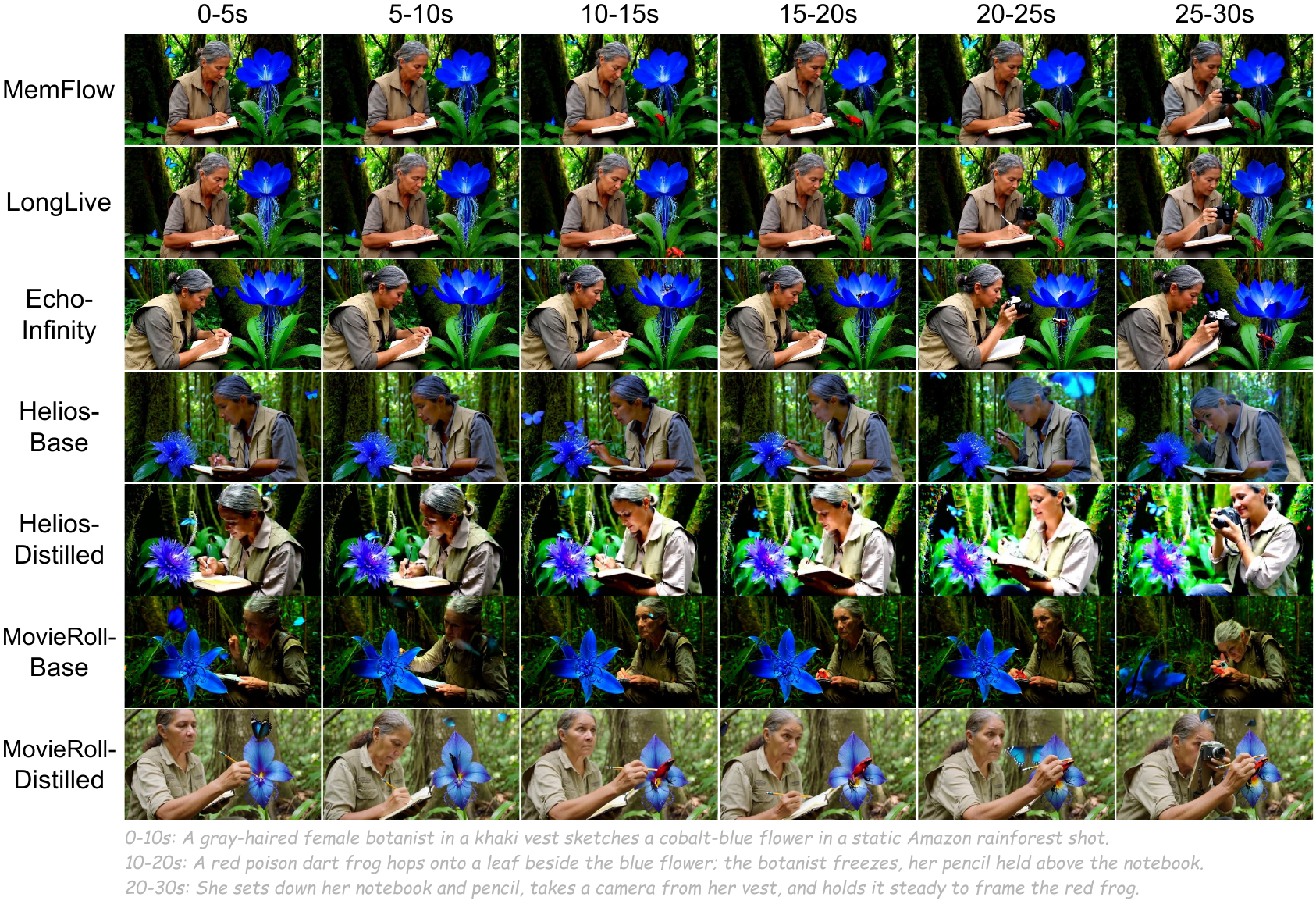}
    \caption{\textbf{Qualitative comparison of 30s multi-event video generation.} The sequence progresses from sketching a flower to observing a frog and raising a camera, with a prompt change every 10s. MovieRoll-Distilled depicts these action changes while retaining the botanist and rainforest setting. Each row shows one method, and each pair of columns corresponds to one event.}
    \label{fig:experiment_multi_event}

\end{figure}

\begin{figure}[!t]
    \centering
    \includegraphics[width=\linewidth]{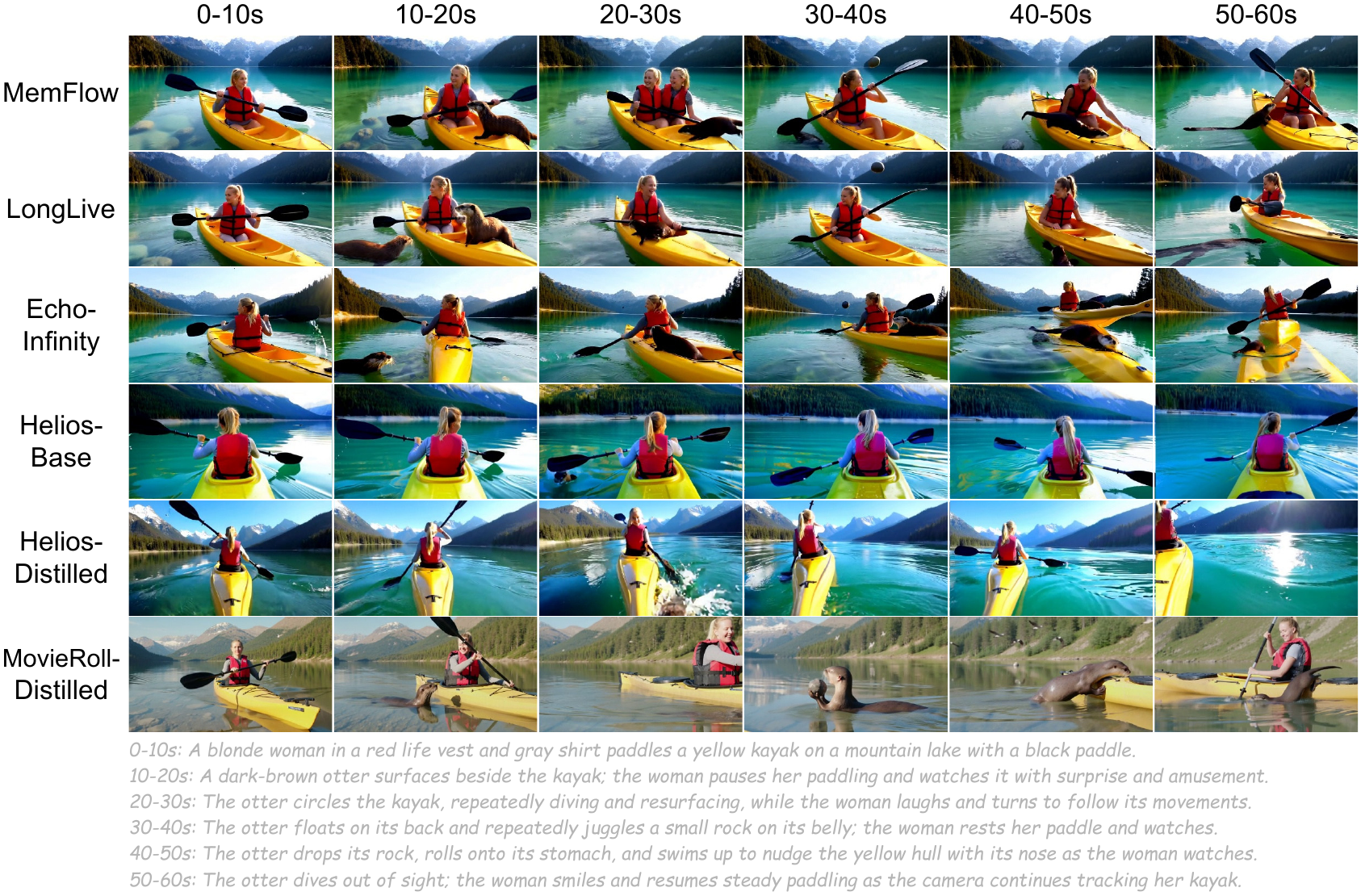}
    \caption{\textbf{Qualitative comparison of 60s multi-event video generation.} The sequence develops from paddling to successive interactions with an otter, including surfacing beside the kayak, handling a rock, and approaching the hull. MovieRoll-Distilled depicts this event progression while maintaining a consistent lake setting. Rows compare methods under the same six sequential prompts, and each column represents one 10s event.}
    \label{fig:experiment_multi_event_1441f}

\end{figure}

We compare multi-event generation with MemFlow~\citep{memflow}, LongLive~\citep{longlive}, Echo-Infinity~\citep{echo_infinity}, and both Helios variants~\citep{helios}. We evaluate three-event videos at 30s and six-event videos at 60s, with a new prompt every 10s. MovieRoll-Base is evaluated at 30s, and MovieRoll-Distilled at both durations. This tests visual continuity and adaptation across prompts.

\textbf{Quantitative Results.} MovieRoll-Distilled achieves the highest total score at both 30s and 60s, outperforming LongLive and both Helios variants, as reported in Tab.~\ref{tab:experiment_multi_event}. It ranks first in dynamics and drifting naturalness at both durations, while its naturalness rating remains at 10 as the sequence extends from three to six events. Thus, MovieRoll-Distilled sustains active motion and visual naturalness across events.

\textbf{Qualitative Results.} MovieRoll-Distilled responds to successive event prompts while preserving scene continuity. In the 30s example shown in Fig.~\ref{fig:experiment_multi_event}, it depicts the transition from sketching a flower to observing a frog and raising a camera, with consistent subject appearance and surroundings. Helios-Base largely continues the sketching action, while Helios-Distilled develops stronger color saturation. In the 60s example shown in Fig.~\ref{fig:experiment_multi_event_1441f}, MovieRoll-Distilled depicts distinct otter actions, from surfacing beside the kayak to handling a rock and approaching the hull, within a consistent lake setting. LongLive shows fewer changes in otter behavior, while both Helios variants largely continue the paddling scene with little visible otter interaction. These examples highlight the ability of MovieRoll-Distilled to combine scene continuity with a clear progression of events over longer durations.

\subsection{Ablation Study}
\label{sec:experiment_ablation_study}

\begin{table*}[!t]
    \centering
    \caption{\textbf{Quantitative ablations of training and inference designs.} WinTF outperforms chunk-wise formulations, while the selected HieraKV allocation, KV recomputation, long-rollout CuLT-DMD, and AnnealCFG schedule achieve the highest total scores within their respective comparisons. Shaded rows identify these choices. Autoregressive formulations are evaluated at 241 and 721 frames, and the remaining comparisons use 721 frames. Higher ratings indicate better quality and less drift.}
    \label{tab:experiment_ablation}

    \small
    \setlength{\tabcolsep}{2.2pt}
    \renewcommand{\arraystretch}{1.08}

    \resizebox{\textwidth}{!}{%
        \begin{tabular}{@{}l*{10}{c}@{}}
            \toprule

            \tablehead{Model Variants}
            & \shortstack{\tablehead{Aesthetic}\\$\uparrow$}
            & \shortstack{\tablehead{Dynamic}\\$\uparrow$}
            & \shortstack{\tablehead{Smoothness}\\$\uparrow$}
            & \shortstack{\tablehead{Semantic}\\$\uparrow$}
            & \shortstack{\tablehead{Naturalness}\\$\uparrow$}
            & \shortstack{\tablehead{Drifting}\\\tablehead{Aesthetic} $\uparrow$}
            & \shortstack{\tablehead{Drifting}\\\tablehead{Smoothness} $\uparrow$}
            & \shortstack{\tablehead{Drifting}\\\tablehead{Semantic} $\uparrow$}
            & \shortstack{\tablehead{Drifting}\\\tablehead{Naturalness} $\uparrow$}
            & \shortstack{\tablehead{Total}\\\tablehead{Score} $\uparrow$}
            \\

            \midrule

            \multicolumn{11}{c}{\textit{Autoregressive Formulation (241-Frame Evaluation)}} \\
            \midrule

            Chunk-wise Teacher Forcing
            & 7       
            & 4       
            & 10      
            & 5       
            & 9       
            & 6       
            & 10      
            & 9       
            & 1       
            & 6.774   
            \\

            Chunk-wise Teacher Forcing with Self-resampling
            & 7       
            & 6       
            & 10      
            & 5       
            & 10      
            & 5       
            & 10      
            & 8       
            & 1       
            & 6.891   
            \\

            \addlinespace[1.5pt]

            \rowcolor{MovieRollColor}
            \tablenum{Window-wise Teacher Forcing}
            & 8       
            & 5       
            & 10      
            & 5       
            & 10      
            & 7       
            & 10      
            & 9       
            & 2       
            & 7.287   
            \\

            \midrule

            \multicolumn{11}{c}{\textit{Autoregressive Formulation (721-Frame Evaluation)}} \\
            \midrule

            Chunk-wise Teacher Forcing
            & 7       
            & 3       
            & 9       
            & 4       
            & 6       
            & 2       
            & 10      
            & 6       
            & 1       
            & 5.001   
            \\

            Chunk-wise Teacher Forcing with Self-resampling
            & 6       
            & 5       
            & 10      
            & 4       
            & 5       
            & 1       
            & 10      
            & 3       
            & 1       
            & 4.410   
            \\

            \addlinespace[1.5pt]

            \rowcolor{MovieRollColor}
            \tablenum{Window-wise Teacher Forcing}
            & 7       
            & 4       
            & 10      
            & 5       
            & 8       
            & 5       
            & 10      
            & 8       
            & 1       
            & 6.321   
            \\

            \midrule
            \multicolumn{11}{c}{\textit{Long-Horizon Extrapolation Strategy}} \\
            \midrule

            Full History
            & 7       
            & 4       
            & 10      
            & 5       
            & 8       
            & 6       
            & 10      
            & 9       
            & 1       
            & 6.519   
            \\

            Sliding Window
            & 7       
            & 4       
            & 10      
            & 5       
            & 8       
            & 6       
            & 10      
            & 9       
            & 1       
            & 6.519   
            \\

            \addlinespace[1.5pt]

            \rowcolor{MovieRollColor}
            \tablenum{KV Recomputation}
            & 7       
            & 4       
            & 10      
            & 5       
            & 9       
            & 6       
            & 10      
            & 9       
            & 2       
            & 6.873   
            \\

            \midrule

            \multicolumn{11}{c}{\textit{HieraKV}} \\
            \midrule

            Global 0 + Local 54 (All Local)
            & 7       
            & 5       
            & 10      
            & 5       
            & 9       
            & 6       
            & 10      
            & 9       
            & 1       
            & 6.804   
            \\

            Global 12 + Local 42
            & 7       
            & 4       
            & 10      
            & 5       
            & 9       
            & 6       
            & 10      
            & 9       
            & 1       
            & 6.774   
            \\

            Global 18 + Local 36
            & 7       
            & 3       
            & 10      
            & 5       
            & 9       
            & 6       
            & 10      
            & 8       
            & 1       
            & 6.645   
            \\

            Global 54 + Local 0 (All Global)
            & 7       
            & 4       
            & 10      
            & 5       
            & 8       
            & 5       
            & 10      
            & 8       
            & 1       
            & 6.321   
            \\

            \addlinespace[1.5pt]

            \rowcolor{MovieRollColor}
            \tablenum{Global 6 + Local 48}
            & 7       
            & 4       
            & 10      
            & 5       
            & 9       
            & 6       
            & 10      
            & 9       
            & 2       
            & 6.873   
            \\

            \midrule

            \multicolumn{11}{c}{\textit{CuLT-DMD} } \\
            \midrule

            LongLive-style Streaming Long Tuning
            & 8       
            & 5       
            & 10      
            & 4       
            & 10      
            & 6       
            & 10      
            & 8       
            & 5       
            & 7.131   
            \\

            Cumulative Long Tuning Only on 241 Frames
            & 8       
            & 6       
            & 10      
            & 5       
            & 10      
            & 5       
            & 10      
            & 8       
            & 4       
            & 7.218   
            \\

            \addlinespace[1.5pt]

            \rowcolor{MovieRollColor}
            \tablenum{Cumulative Long Tuning}
            & 8       
            & 6       
            & 10      
            & 4       
            & 10      
            & 6       
            & 10      
            & 8       
            & 6       
            & 7.260   
            \\

            \midrule

            \multicolumn{11}{c}{\textit{AnnealCFG} } \\
            \midrule

            Constant CFG = 6
            & 7       
            & 5       
            & 9       
            & 5       
            & 5       
            & 3       
            & 8       
            & 8       
            & 1       
            & 5.160   
            \\

            Constant CFG = 4
            & 7       
            & 4       
            & 9       
            & 5       
            & 8       
            & 6       
            & 10      
            & 9       
            & 1       
            & 6.489   
            \\

            Linear Increase ($1 \rightarrow 6$)
            & 7 & 3 & 10 & 4 & 10 & 6 & 10 & 9 & 1 & 6.744 \\

            Linear Increase ($1 \rightarrow 4$)
            & 7 & 2 & 10 & 4 & 10 & 5 & 10 & 8 & 1 & 6.516 \\

            Linear Decay ($6 \rightarrow 1$)
            & 7       
            & 5       
            & 9       
            & 5       
            & 7       
            & 5       
            & 10      
            & 9       
            & 2       
            & 6.264   
            \\

            \addlinespace[1.5pt]

            \rowcolor{MovieRollColor}
            \tablenum{Linear Decay ($4 \rightarrow 1$)}
            & 7       
            & 4       
            & 10      
            & 5       
            & 9       
            & 6       
            & 10      
            & 9       
            & 2       
            & 6.873   
            \\

            \bottomrule
        \end{tabular}%
    }

\end{table*}

\subsubsection{Autoregressive Formulation}

\begin{figure}[!t]
    \centering
    \includegraphics[width=\linewidth]{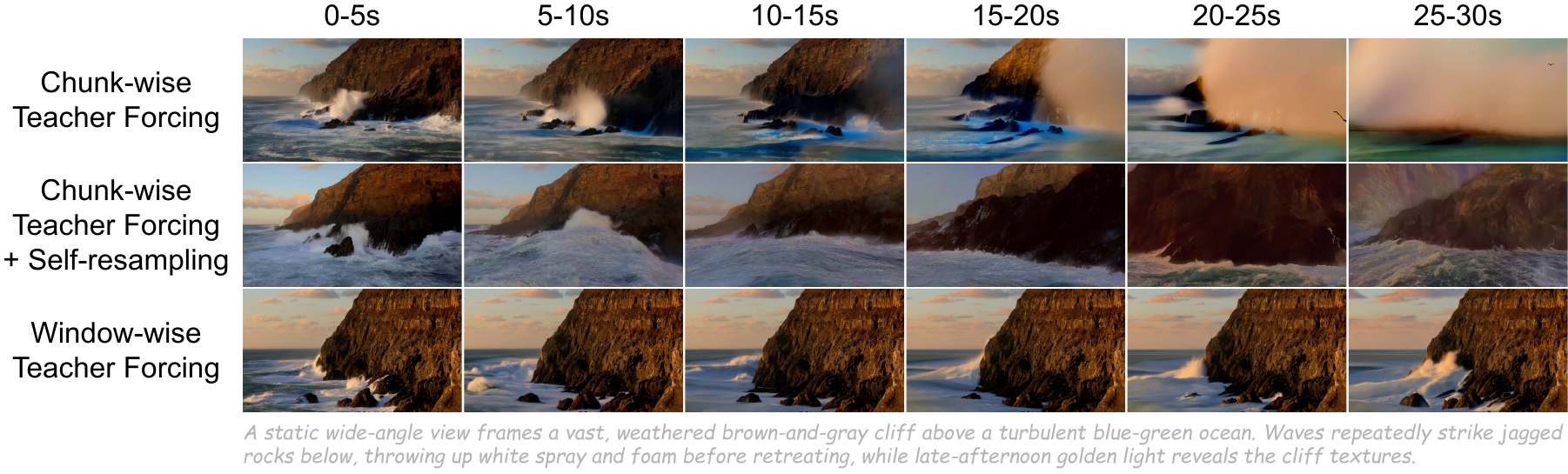}
    \caption{\textbf{Ablation of autoregressive training strategies.} The comparison includes chunk-wise teacher forcing, its self-resampling extension, and WinTF under the same coastal-scene prompt. As generation progresses, WinTF retains the cliff boundary and wave structure, while the chunk-wise variants lose scene detail or develop stronger changes in the colors of the coastal scene.}
    \label{fig:experiment_ablation_autoregressive}

\end{figure}

We compare Window-wise and Chunk-wise Teacher Forcing after 10000 training steps on the same data, together with a chunk-wise variant trained for another 10000 steps using self-resampling~\citep{resampling_forcing}. All variants use full-history attention without HieraKV and KV recomputation.

Window-wise Teacher Forcing achieves the highest total score at both 10s and 30s, with a larger lead at 30s, as reported in Tab.~\ref{tab:experiment_ablation}. At the longer duration, it also improves naturalness and semantic alignment over both chunk-wise variants, including the one with additional self-resampling training. This advantage is also visible in Fig.~\ref{fig:experiment_ablation_autoregressive}, where WinTF preserves the cliff boundary and wave detail throughout the 30s sequence, while the chunk-wise variants lose scene detail or develop stronger color changes.

\subsubsection{Long-Horizon Extrapolation Strategy}

\begin{figure}[!t]
    \centering
    \includegraphics[width=\linewidth]{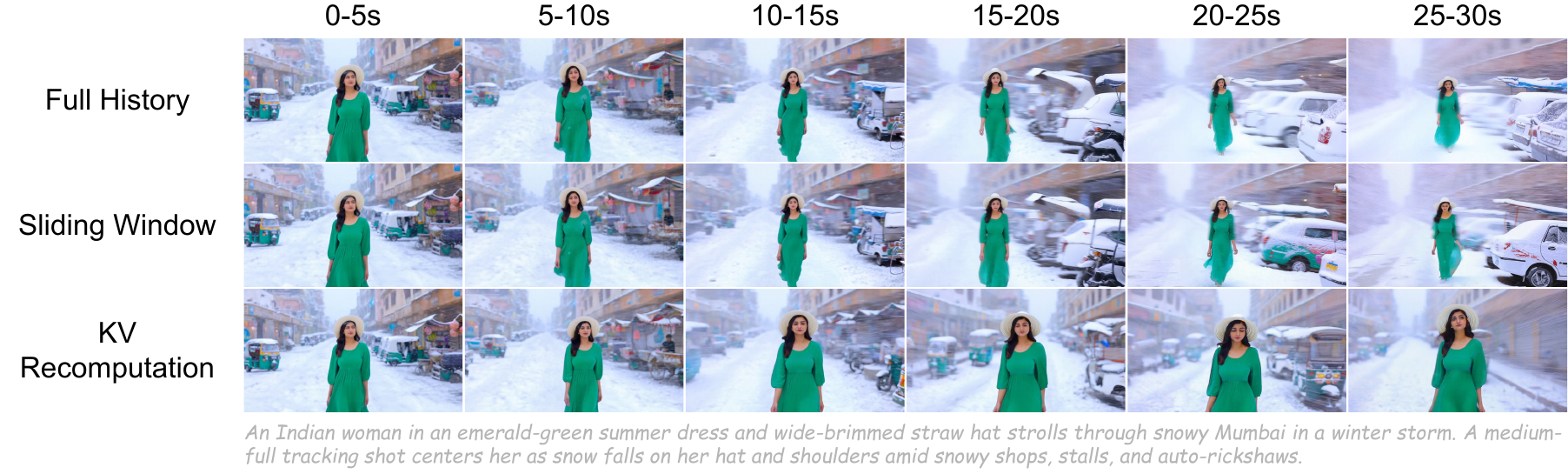}
    \caption{\textbf{Ablation of long-horizon extrapolation strategies.} All three strategies use the same Stage~1 MovieRoll-Base checkpoint to extend generation from 10s training clips to 30s videos. Full-history and sliding-window caching produce increasingly washed-out backgrounds, whereas KV recomputation preserves clearer subject and street details through the end of the sequence.}
    \label{fig:experiment_ablation_extrapolation}

\end{figure}

Using the same Stage~1 MovieRoll-Base checkpoint trained on videos of up to 10s, we compare full-history caching, sliding-window caching, and threshold-triggered KV recomputation at 30s. With the local-history policy fixed, the sliding window retains 15 chunks, and recomputation is triggered at the same cache length.

KV recomputation achieves the highest total score by improving both naturalness and drifting naturalness over full-history and sliding-window caching, as reported in Tab.~\ref{tab:experiment_ablation}. With the same model checkpoint, rebuilding the cached context therefore improves generation beyond the training horizon. Consistent with these gains, recomputation preserves clearer subject and street details in the 30s example shown in Fig.~\ref{fig:experiment_ablation_extrapolation}, while both alternatives produce increasingly washed-out backgrounds.

\subsubsection{HieraKV}

\begin{figure}[!t]
    \centering
    \includegraphics[width=\linewidth]{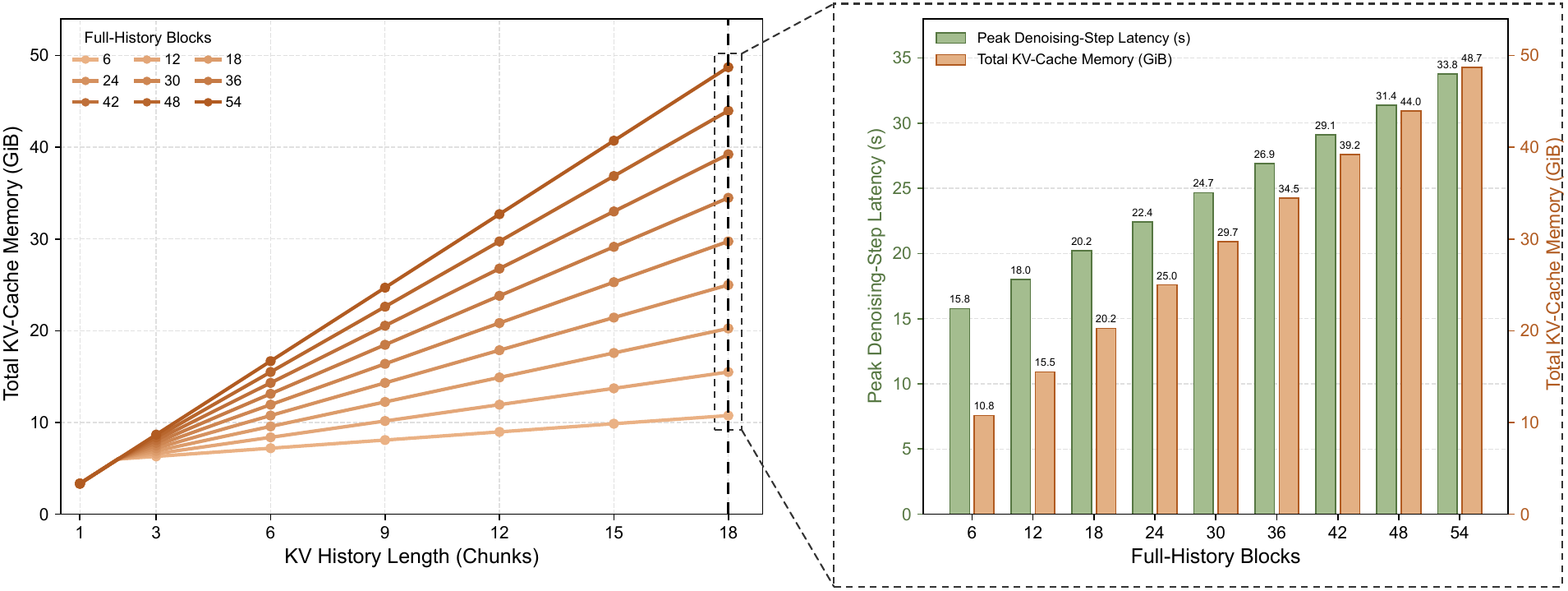}
    \caption{\textbf{Memory and latency comparison across full-history and sink--recent attention allocations.} The left plot shows slower cache-memory growth when fewer layers retain full history. At 18 cached chunks, the default HieraKV allocation of 6 global and 48 local layers reduces memory from 48.7 to 10.8 GiB and peak denoising-step latency from 33.8 to 15.8s relative to all-global attention. Measurements use MovieRoll-Base with CFG on one NVIDIA H20 GPU. Each timed denoising step includes both conditional and unconditional DiT evaluations. Timing excludes KV recomputation and VAE encoding/decoding.}
    \label{fig:hierakv_vary_global_blocks}

\end{figure}

\begin{figure}[!t]
    \centering
    \includegraphics[width=\linewidth]{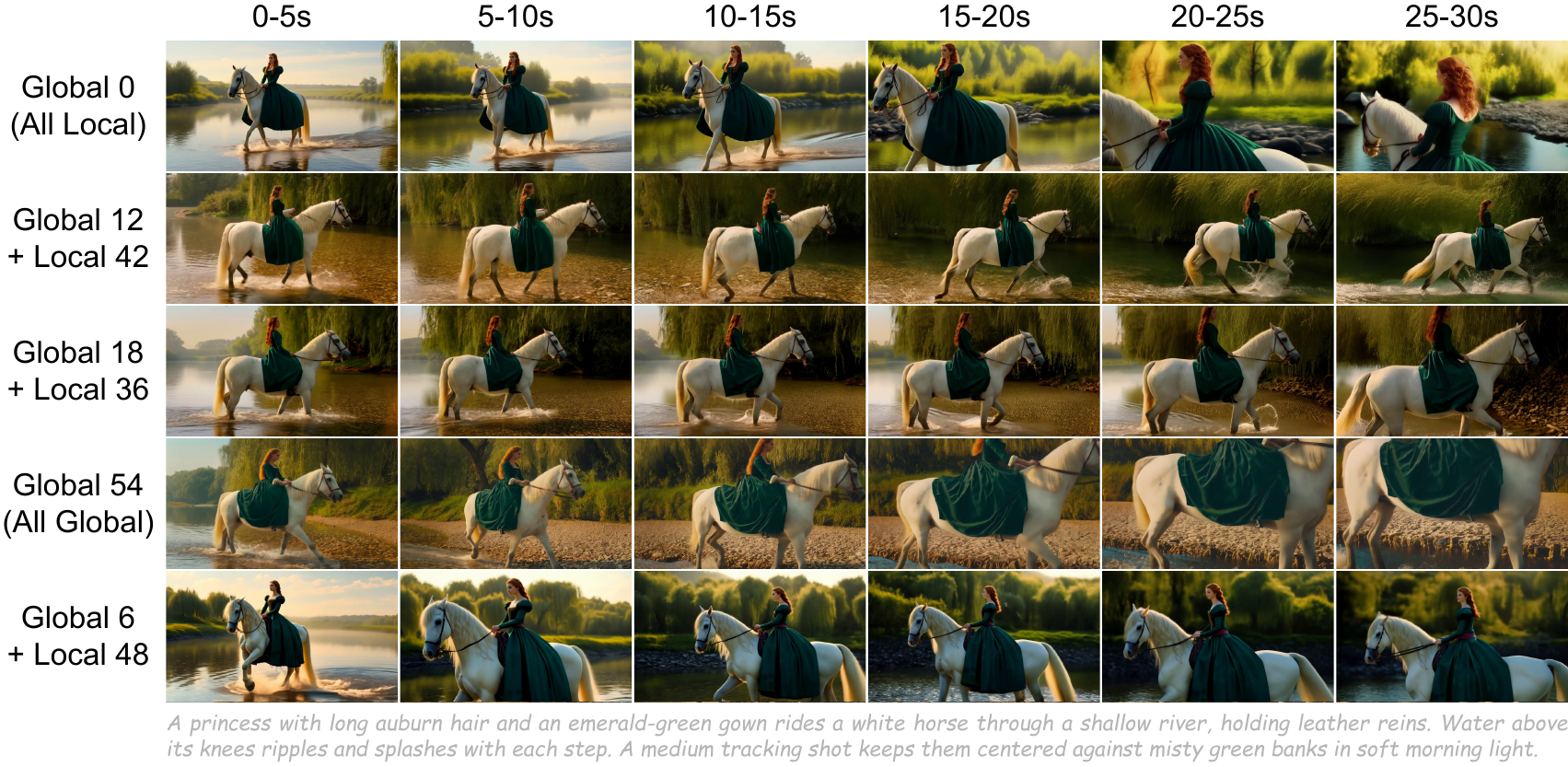}
    \caption{\textbf{Ablation of HieraKV attention allocations.} The comparison varies the allocation of 54 transformer blocks between full-history and sink--recent attention. With the default HieraKV allocation of 6 global and 48 local blocks, the rider and horse remain recognizable throughout the sequence. All-global attention progressively crops the subjects, while all-local attention shifts toward a close rear view.}
    \label{fig:experiment_ablation_hierakv}

\end{figure}

We compare five allocations of full-history (global) and sink--recent (local) blocks, including all-global and all-local attention, using the same training data and recomputation threshold, as detailed in App.~\ref{app:implementation}.

The allocation of 6 full-history and 48 sink--recent blocks achieves the highest total score and drifting naturalness rating among the tested configurations, as reported in Tab.~\ref{tab:experiment_ablation}. Beyond these quality gains, this allocation reduces KV-cache memory and peak denoising-step latency with CFG in MovieRoll-Base relative to all-global attention, as shown in Fig.~\ref{fig:hierakv_vary_global_blocks}. The quality advantage is further illustrated in Fig.~\ref{fig:experiment_ablation_hierakv}, where the default allocation keeps the rider and horse in view, while all-global attention progressively crops them and all-local attention shifts toward a close rear view. These results support retaining full history in a small subset of layers.

\subsubsection{CuLT-DMD}

\begin{figure}[!t]
    \centering
    \includegraphics[width=\linewidth]{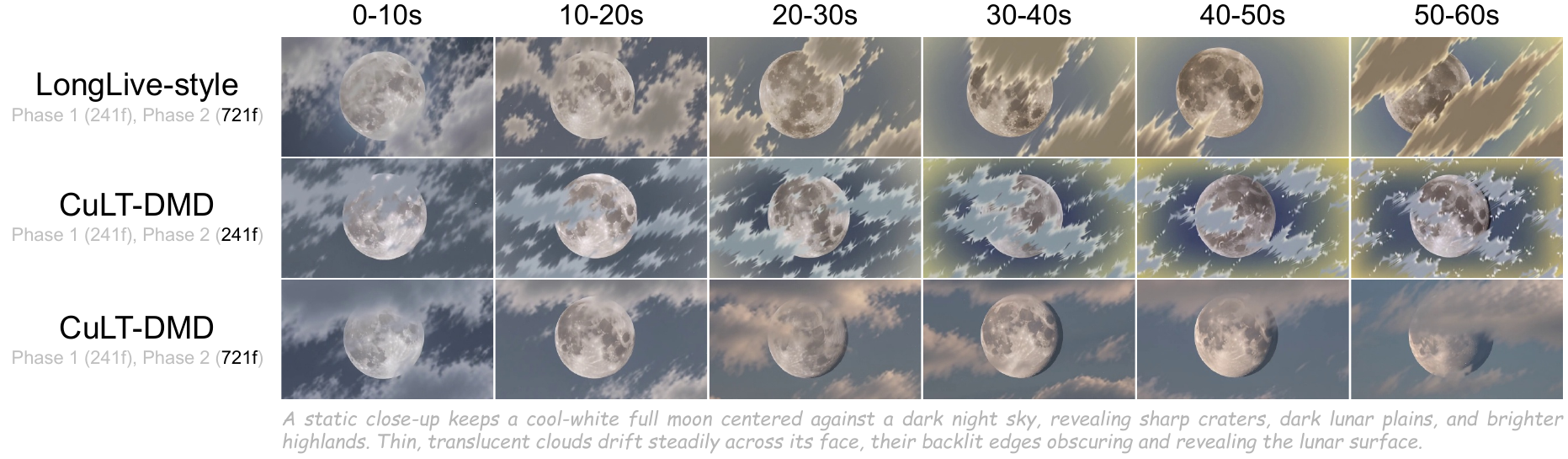}
    \caption{\textbf{Ablation of CuLT-DMD training strategies.} The 60s sequences compare LongLive-style training with CuLT-DMD using 241-frame or 721-frame Phase~2 rollouts. LongLive-style training shows larger changes in moon position and scale, while short-rollout CuLT-DMD develops stronger cloud-texture changes. Long-rollout CuLT-DMD maintains a more consistent moon-and-sky appearance in this example.}
    \label{fig:experiment_ablation_cult_dmd}

\end{figure}

We compare CuLT-DMD with LongLive-style training~\citep{longlive}, starting from the same MovieRoll-Base checkpoint. Both use 241-frame rollouts in Phase~1 and 721-frame rollouts in Phase~2. CuLT-DMD accumulates gradients across each rollout, whereas the baseline updates from individual supervision windows. A third variant uses CuLT-DMD with 241-frame rollouts in both phases.

CuLT-DMD achieves the highest total score and improves dynamics and drifting naturalness over LongLive-style training, as reported in Tab.~\ref{tab:experiment_ablation}. Moreover, extending Phase~2 from 241-frame to 721-frame rollouts improves drifting aesthetics and drifting naturalness over short-rollout CuLT-DMD, supporting the use of longer self-rollouts for sustained generation. This improved temporal consistency is reflected in Fig.~\ref{fig:experiment_ablation_cult_dmd}, where long-rollout CuLT-DMD preserves a more consistent moon-and-sky appearance over 60s. In comparison, LongLive-style training produces larger changes in moon position and scale, while short-rollout CuLT-DMD develops stronger changes in cloud texture.

\subsubsection{AnnealCFG}

\begin{figure}[!t]
    \centering
    \includegraphics[width=\linewidth]{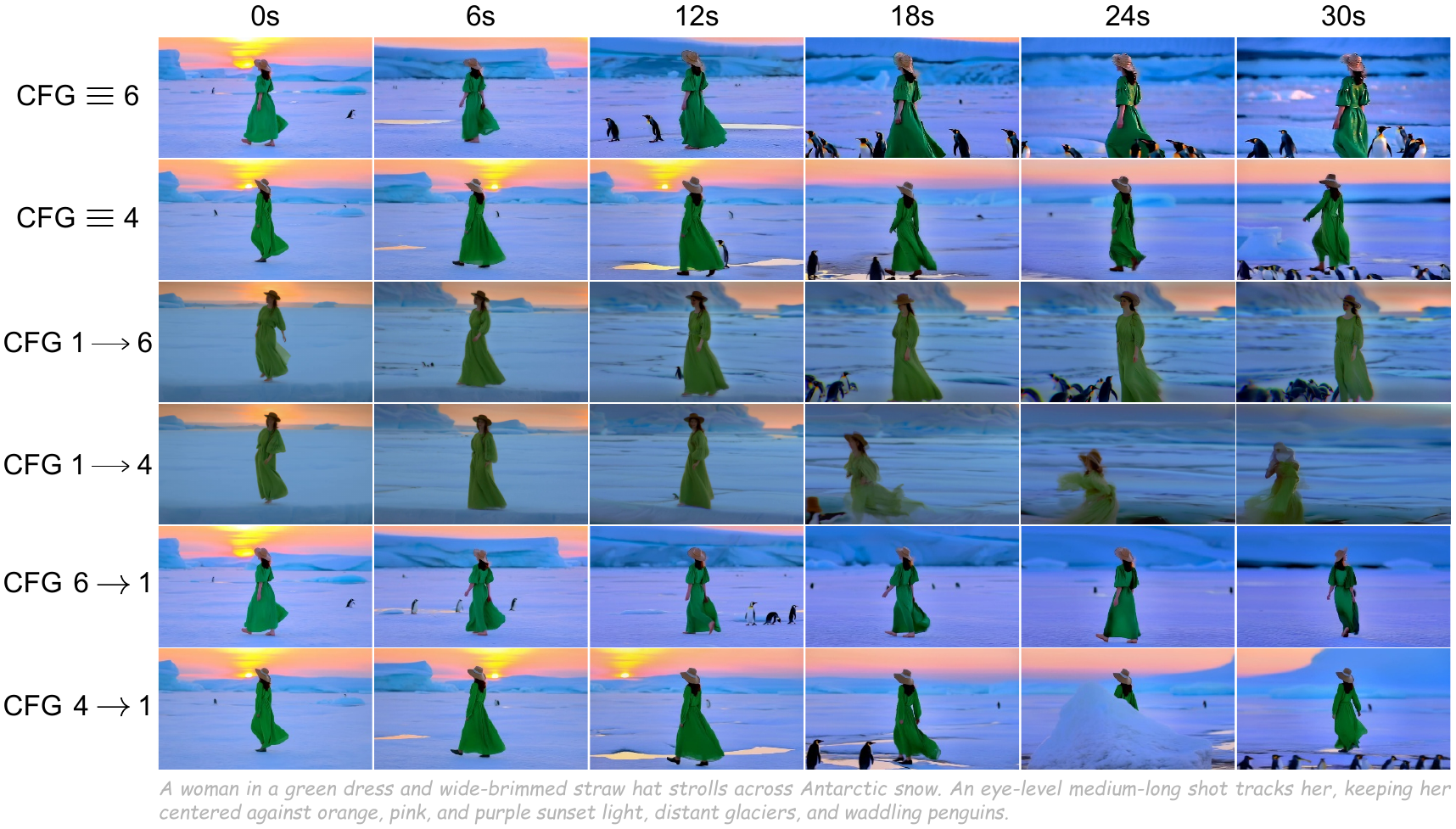}
    \caption{\textbf{Ablation of CFG schedules.} Rows compare constant, increasing, and decreasing guidance using the same MovieRoll-Base checkpoint. Guidance changes within each denoising trajectory. The default $4\rightarrow1$ schedule retains clearer background detail, while constant guidance increases color saturation and increasing guidance from 1 to 4 distorts the subject in later frames.}
    \label{fig:experiment_ablation_annealcfg}

\end{figure}

Using MovieRoll-Base, we compare constant, linearly increasing, and linearly decreasing guidance at scales of 4 and 6. 

Annealing CFG from 4 to 1 achieves the highest total score, as reported in Tab.~\ref{tab:experiment_ablation}. It improves naturalness and drifting naturalness over constant guidance at scale 4, and preserves stronger dynamics, semantic alignment, and drifting naturalness than either increasing schedule. Consistent with these quantitative gains, the annealing schedule preserves clearer background detail in the 30s comparison shown in Fig.~\ref{fig:experiment_ablation_annealcfg}, while constant guidance produces stronger color saturation and increasing guidance from 1 to 4 distorts the subject in later frames. These results support the annealing schedule for maintaining both visual quality and temporal stability during generation.

\subsection{User Study}
\label{sec:experiment_user_study}

\begin{figure}[!t]

    \centering
    \includegraphics[width=0.87\linewidth]{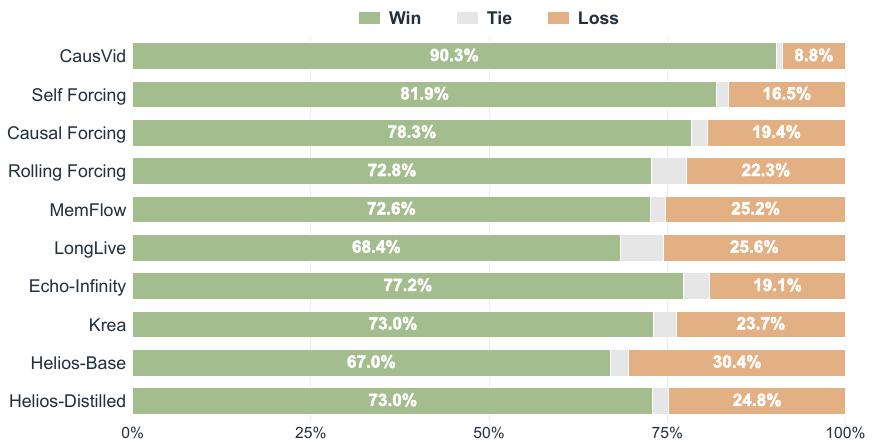}
    \caption{\textbf{User preference comparison for single-event video generation.} MovieRoll-Distilled wins a majority of judgments against every evaluated baseline, with win rates from 67.0\% to 90.3\%. Each bar reports win, tie, and loss rates from the MovieRoll-Distilled perspective in same-prompt comparisons, pooling individual judgments across 10, 30, and 60s videos with ties included in the denominator.}
    \label{fig:experiment_user_study_single_event}
\end{figure}

\begin{figure}[!t]
    \centering
    \includegraphics[width=0.87\linewidth]{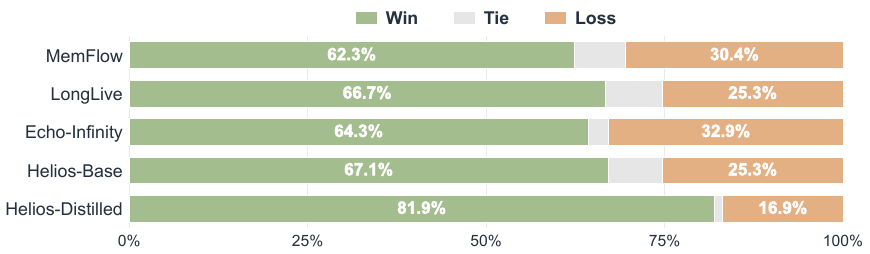}
    \caption{\textbf{User preference comparison for multi-event video generation.} MovieRoll-Distilled wins a majority of judgments against every evaluated baseline, with win rates from 62.3\% to 81.9\%. Each bar reports win, tie, and loss rates from the MovieRoll-Distilled perspective under the same prompt sequences, pooling judgments across 30s three-event and 60s six-event videos with ties included in the denominator.}
    \label{fig:experiment_user_study_multi_event}

\end{figure}

We conduct a pairwise preference study with 59 professionals, with at least three experts evaluating each video pair. Participants compare videos generated from the same prompt and choose a preferred video or a tie. The study covers single-event generation at 10, 30, and 60s and multi-event generation at 30 and 60s. We aggregate win, tie, and loss rates across durations within each setting. App.~\ref{app:user_study} provides the sample sizes and explains how judgments are aggregated for each baseline and generation setting.

MovieRoll-Distilled wins a majority of judgments against every evaluated baseline in both settings, as shown in Figs.~\ref{fig:experiment_user_study_single_event} and~\ref{fig:experiment_user_study_multi_event}. The closest comparisons are with Helios-Base for single-event generation and MemFlow for multi-event generation, consistent with their competitive automatic evaluation results.

\subsection{More Showcases}
\label{sec:experiment_showcase}
\label{sec:experiment_multi_event_showcase}

\begin{figure}[!t]

    \centering
    \includegraphics[width=0.94\linewidth,height=0.88\textheight,keepaspectratio]{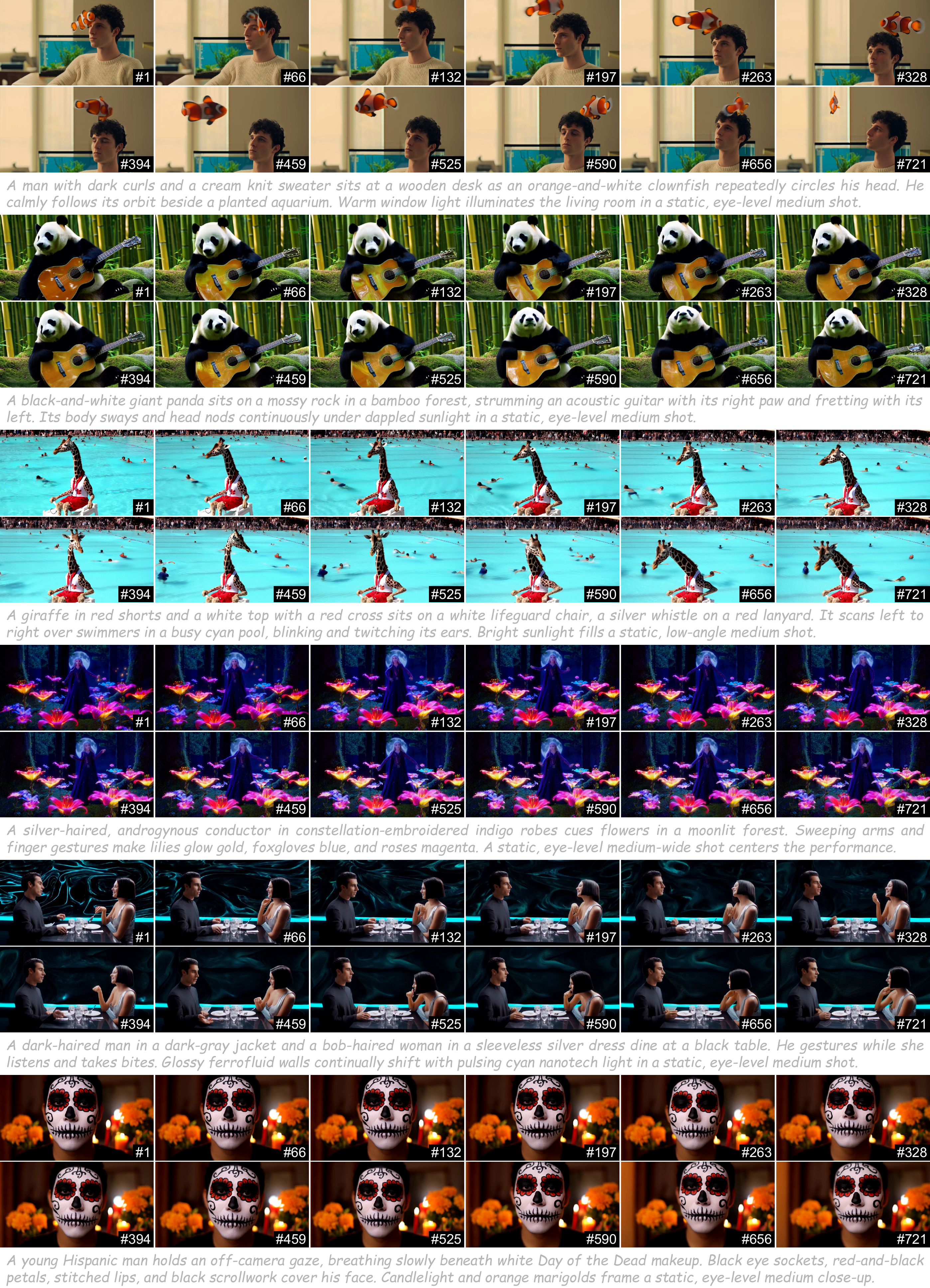}
    \caption{\textbf{Examples of 30s single-event videos generated by MovieRoll-Base.} The six examples span human portraits, anthropomorphic performances, and fantastical settings, showing recognizable subjects as actions develop. Each two-row group follows one 721-frame video from left to right and then top to bottom. Frame indices and the accompanying prompt identify the progression within each example.}
    \label{fig:showcase_base}

\end{figure}

Fig.~\ref{fig:showcase_base} presents six 30s MovieRoll-Base examples spanning human portraits, anthropomorphic performances, and fantastical scenes. Alongside recurring actions such as playing a guitar, the examples retain recognizable faces, clothing, and scene layouts through changes in pose. This range illustrates how the base model supports detailed appearance and ongoing action across realistic and imaginative prompts.

\begin{figure}[!t]
    \centering
    \includegraphics[width=0.94\linewidth,height=0.88\textheight,keepaspectratio]{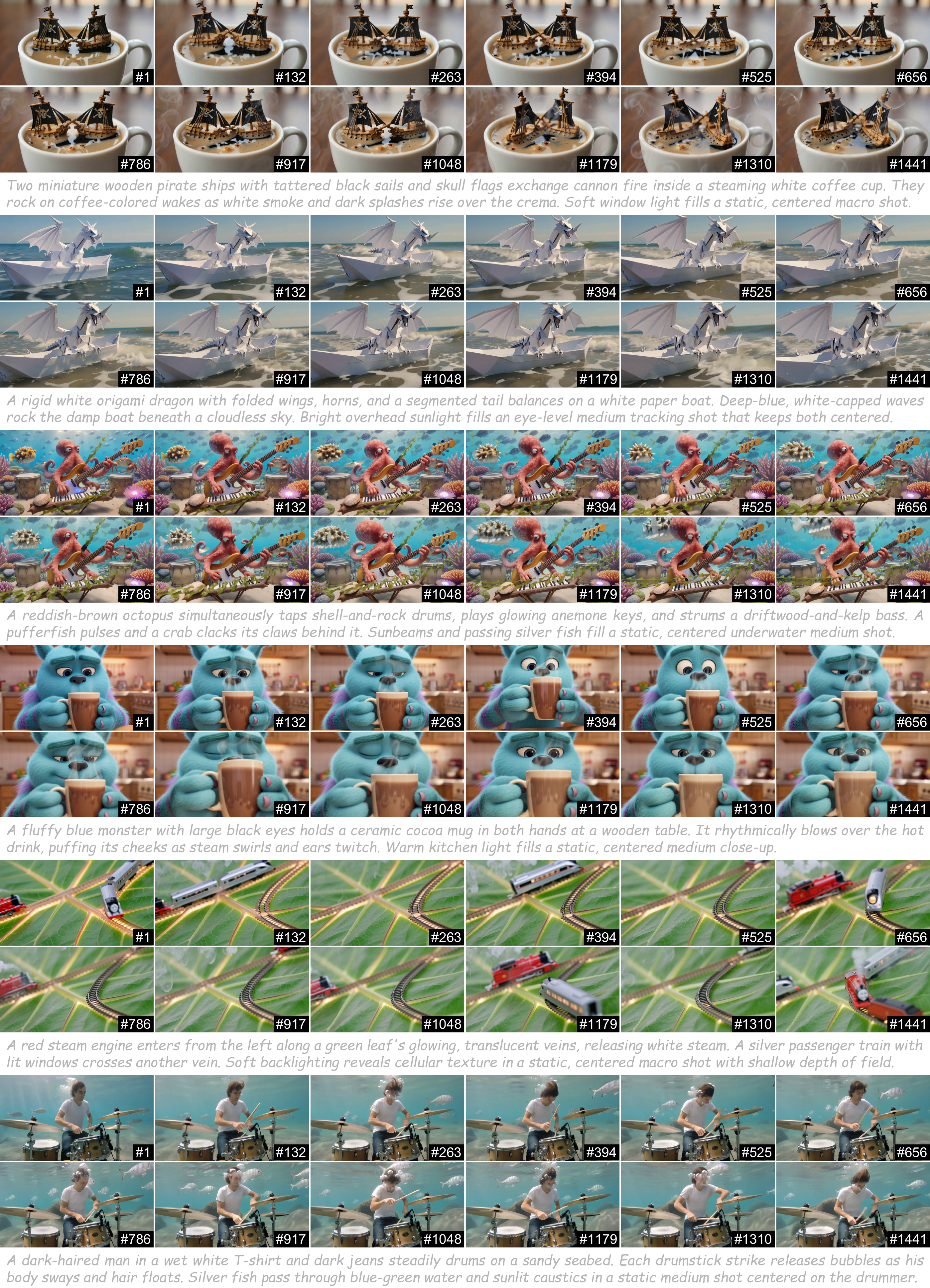}
    \caption{\textbf{Examples of 60s single-event videos generated by MovieRoll-Distilled.} This first set of six examples includes miniature ships, an origami dragon, and underwater performances. MovieRoll-Distilled preserves subject appearance as waves, steam, and gestures evolve. Each two-row group follows one 1441-frame video from left to right and then top to bottom, with frame indices overlaid and its prompt below.}
    \label{fig:showcase_distilled_1}

\end{figure}

The six 60s MovieRoll-Distilled examples in Fig.~\ref{fig:showcase_distilled_1} extend the showcases to minute-length generation. They combine persistent subject structure with motion in the surrounding scene, as illustrated by an origami dragon on a boat and an underwater drummer. The main characters and objects remain recognizable while waves, steam, and gestures continue to change, showing sustained activity alongside visual continuity.

\begin{figure}[!t]
    \centering
    \includegraphics[width=0.94\linewidth,height=0.88\textheight,keepaspectratio]{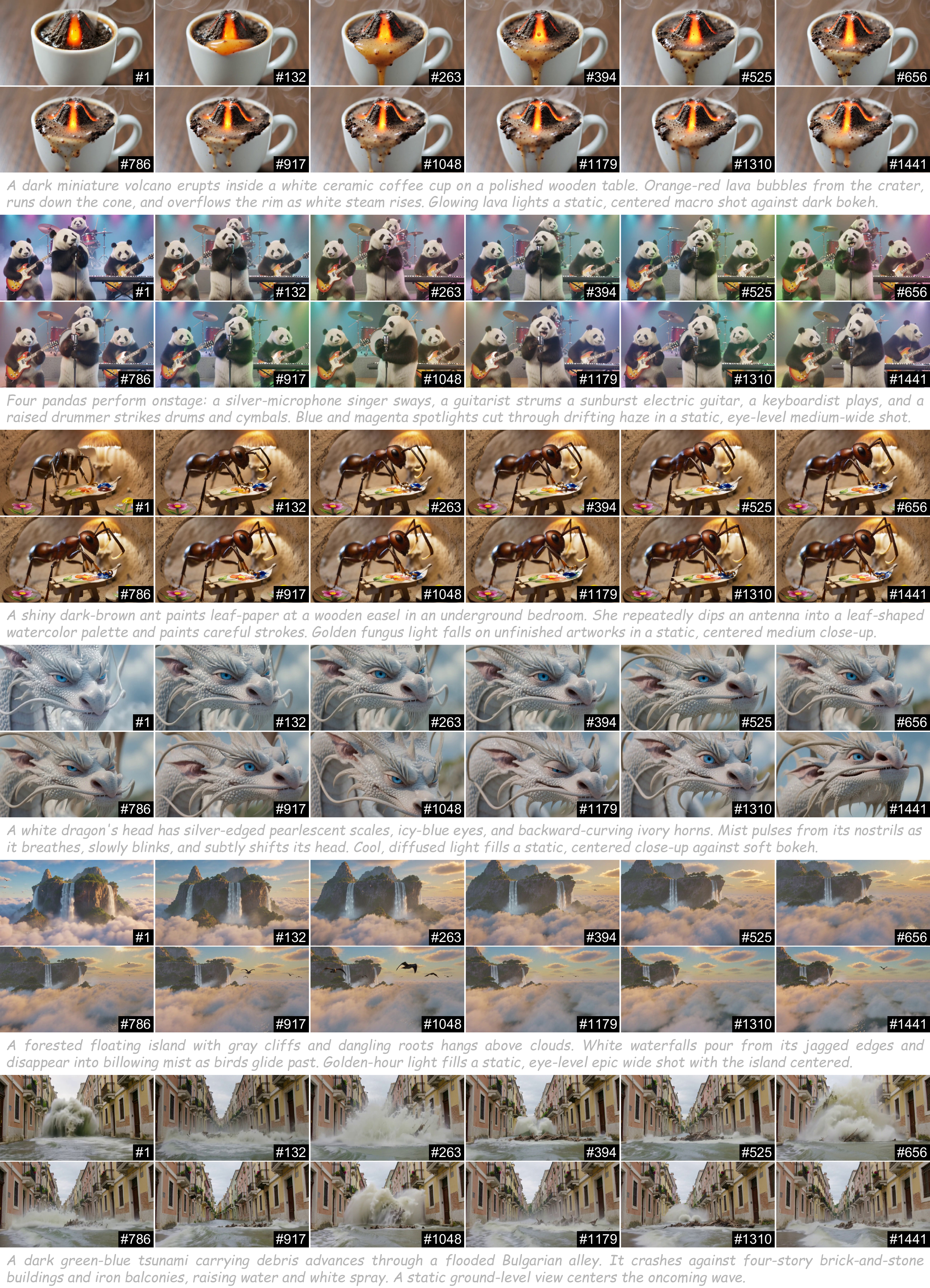}
    \caption{\textbf{Additional examples of 60s single-event videos generated by MovieRoll-Distilled.} This second set of six examples spans miniature eruptions, animal performances, dragon close-ups, and landscapes. MovieRoll-Distilled retains appearance details alongside sustained scene activity. Each two-row group follows one 1441-frame video from left to right and then top to bottom, with frame indices overlaid and its prompt below.}
    \label{fig:showcase_distilled_2}

\end{figure}

A further six 60s examples in Fig.~\ref{fig:showcase_distilled_2} cover a wider range of visual scales and appearances, including miniature eruptions, animal performances, and landscape scenes. The coffee-cup volcano combines a recognizable setting with changing lava and steam, while the dragon close-up retains fine facial detail through changes in expression and pose. These examples illustrate detail retention alongside sustained scene activity.

\begin{figure}[!t]
    \centering
    \includegraphics[width=0.94\linewidth,height=0.88\textheight,keepaspectratio]{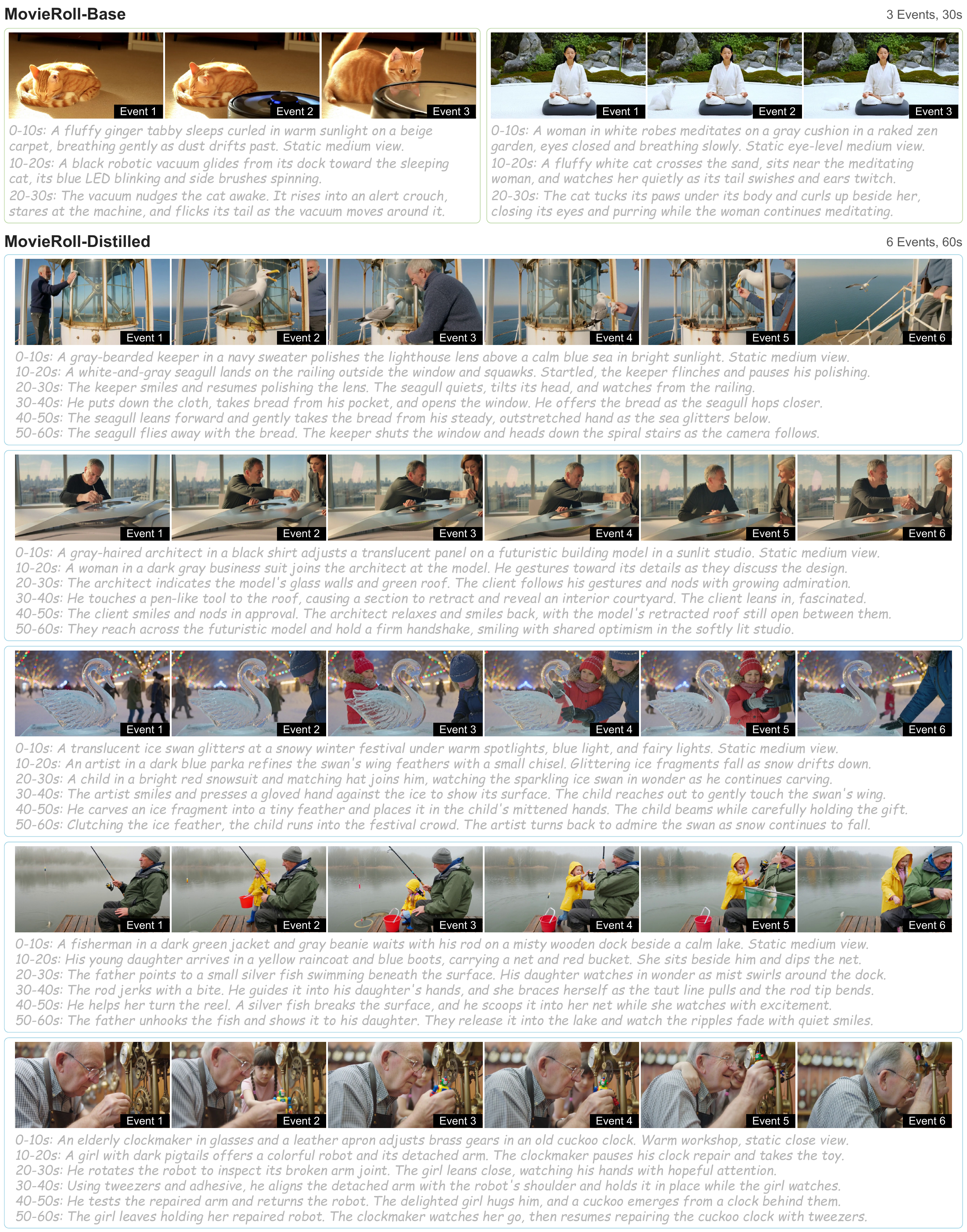}
    \caption{\textbf{Examples of multi-event video generation with sequential prompts.} The upper examples show three-event, 30s videos from MovieRoll-Base, and the lower examples show six-event, 60s videos from MovieRoll-Distilled. Prompts change every 10s, with one frame representing each event. Both models retain recurring characters and settings as new interactions develop, such as a cat reacting to a robotic vacuum and a keeper responding to a seagull.}
    \label{fig:showcase_multi_event}

\end{figure}

Fig.~\ref{fig:showcase_multi_event} complements the single-event examples with three-event MovieRoll-Base videos at 30s and six-event MovieRoll-Distilled videos at 60s. As prompts change every 10s, new participants enter established settings and ongoing actions develop into interactions. A cat reacts to a robotic vacuum, a lighthouse keeper interacts with a seagull, and a child joins a sculptor. Across these changes, recurring characters and objects remain recognizable, illustrating scene continuity through successive events.
\section{Conclusion and Limitations}
\label{sec:conclusion}
\subsection{Conclusion}
MovieRoll connects streaming fine-tuning, long-rollout distillation, and long-horizon inference in a base-to-distilled framework. WinTF establishes a robust multi-step streaming base, and CuLT-DMD builds on it to train a few-step generator through supervision across long self-rollouts. HieraKV and KV recomputation retain historical context efficiently for both models, while AnnealCFG further stabilizes MovieRoll-Base. FBCA and PS-VAE complement these methods with efficient training and streaming decoding. Together, these designs enable competitive 30s generation from MovieRoll-Base despite training on videos of up to 15s. MovieRoll-Distilled achieves the highest total HeliosBench scores among the compared methods across single- and multi-event evaluations up to 60s, combining sustained naturalness with active motion.

\subsection{Limitations}

\begin{table}[!t]
    
    \centering
    \caption{\textbf{Single-GPU inference efficiency for 60s video generation.} The table compares denoising and end-to-end costs across methods at the listed resolutions for 1441-frame videos on one NVIDIA H20 GPU. Switching from FlashAttention-3 to SageAttention increases MovieRoll-Distilled end-to-end throughput from 1.52 to 2.04 FPS. Single-GPU throughput remains below the 24 FPS playback rate required for real-time generation.}
    \label{tab:inference_speed_h20}

    \small
    \setlength{\tabcolsep}{3.0pt}
    \renewcommand{\arraystretch}{1.08}

    \resizebox{0.7\columnwidth}{!}{%
        \begin{tabular}{@{}lcc*{4}{c}@{}}
            \toprule

            \multirow{2}{*}{\tablehead{Method}}
            & \multirow{2}{*}{\tablehead{Params}}
            & \multirow{2}{*}{\makecell[c]{\tablehead{Resolution} \\ ($H\times W$)}}
            & \multicolumn{2}{c}{\tablehead{Inference Time (s)}}
            & \multicolumn{2}{c}{\tablehead{FPS} $\uparrow$}
            \\

            \cmidrule(lr){4-5}
            \cmidrule(lr){6-7}

            & & &
            \tablehead{E2E}
            & \tablehead{Denoising}
            & \tablehead{E2E}
            & \tablehead{Denoising}
            \\

            \midrule

            CausVid
            & 1.3B
            & $480\times832$
            & 232
            & 131
            & 6.21
            & 11.00
            \\

            Self Forcing
            & 1.3B
            & $480\times832$
            & 455
            & 361
            & 3.17
            & 3.99
            \\

            Causal Forcing
            & 1.3B
            & $480\times832$
            & 429
            & 336
            & 3.36
            & 4.29
            \\

            Rolling Forcing
            & 1.3B
            & $480\times832$
            & 500
            & 407
            & 2.88
            & 3.54
            \\

            MemFlow
            & 1.3B
            & $480\times832$
            & 338
            & 253
            & 4.26
            & 5.70
            \\

            LongLive
            & 1.3B
            & $480\times832$
            & 264
            & 175
            & 5.46
            & 8.23
            \\

            Echo-Infinity
            & 1.3B
            & $480\times832$
            & 304
            & 210
            & 4.74
            & 6.86
            \\

            Krea
            & 14B
            & $480\times832$
            & 1270
            & 1161
            & 1.13
            & 1.24
            \\

            Helios-Distilled
            & 14B
            & $384\times640$
            & 470
            & 407
            & 3.07
            & 3.54
            \\

            \addlinespace[1.5pt]

            \rowcolor{MovieRollColor}
            \tablenum{MovieRoll-Distilled (FA3)}
            & \tablenum{8.3B}
            & $480\times864$
            & 948
            & 763
            & 1.52
            & 1.89
            \\

            \rowcolor{MovieRollColor}
            \tablenum{MovieRoll-Distilled (Sage)}
            & \tablenum{8.3B}
            & $480\times864$
            & 707
            & 521
            & 2.04
            & 2.77
            \\

            \bottomrule
        \end{tabular}%
    }
    
\end{table}

\begin{table}[!t]
    \centering
    \caption{\textbf{Inference scaling across GPU counts and attention backends.} Increasing the GPU count improves MovieRoll-Distilled throughput with both backends, and SageAttention is faster at every tested count. On eight NVIDIA H20 GPUs, the best setting generates a 60s video in 152.2s, reaching 9.47 FPS. Real-time generation at 24 FPS remains a limitation even with eight GPUs. Videos contain 1441 frames at $480\times864$.}
    \label{tab:inference_speed_scaling}
    \fontsize{9.5}{11.4}\selectfont
    \begin{tabular}{@{}ccccccc@{}}
        \toprule
        \multirow{3}{*}{\tablehead{GPUs}}
        & \multicolumn{3}{c}{\tablehead{FlashAttention-3}}
        & \multicolumn{3}{c}{\tablehead{SageAttention}} \\
        \cmidrule(lr){2-4}\cmidrule(l){5-7}
        & \tablehead{Denoising} & \tablehead{Total} & \tablehead{Throughput}
        & \tablehead{Denoising} & \tablehead{Total} & \tablehead{Throughput} \\
        & (s) $\downarrow$ & (s) $\downarrow$ & (FPS) $\uparrow$
        & (s) $\downarrow$ & (s) $\downarrow$ & (FPS) $\uparrow$ \\
        \midrule
        1 & 763 & 948 & 1.52 & 521 & 707 & 2.04 \\
        2 & 423 & 538 & 2.68 & 293 & 407 & 3.54 \\
        4 & 229 & 304 & 4.75 & 158 & 236 & 6.11 \\
        8 & 131 & 185 & 7.77 & 95 & 152 & 9.47 \\
        \bottomrule
    \end{tabular}
\end{table}

MovieRoll-Distilled has not yet achieved real-time generation. Tab.~\ref{tab:inference_speed_h20} reports single-GPU comparisons, while Tab.~\ref{tab:inference_speed_scaling} evaluates sequence parallelism and attention backends. Even with SageAttention and eight H20 GPUs, generating a 60s video takes 152.2s, yielding 9.47 FPS against the 24 FPS required for real-time playback. HieraKV reduces cache memory and denoising-step latency in MovieRoll-Base with CFG relative to all-global attention, but joint window denoising, periodic KV recomputation, and video decoding remain costly.

Long-horizon stability also remains a limitation. MovieRoll-Base is insufficiently stable at 60s, so only MovieRoll-Distilled is evaluated at this duration. Despite high full-video naturalness, the distilled model receives lower semantic and drifting-naturalness ratings at the longest evaluated duration than at the shortest.

\section{Discussion and Future Work}
\label{sec:future_work}

\subsection{Agentic Streaming Generation}
\label{sec:future_work_agentic}

In our experience, streaming models approach bidirectional quality in visually constrained settings, such as digital-human animation and game-world scenes with limited visual variation, but struggle with complex motion and camera trajectories. Difficulty maintaining continuity as instructions change may contribute to this gap. Even with a fixed prompt, an unfolding event can leave the next action unspecified. These challenges motivate generation guided by both user intent and the current visual state.

An agent could provide this guidance by observing newly generated frames and translating user instructions into successive events. It could update the active prompt as the scene develops and plan transitions when new instructions conflict with existing objects or states. Such planning would connect immediate instructions to the generated history while preserving compatible scene content.

The Streaming Video Prompt Rewriting Skill in Sec.~\ref{sec:method_rewrite_skill} provides a starting point through its rules for event organization and textual continuity. Visual feedback could extend this process by guiding revisions to the event plan during generation. To support these finer adjustments, window-level captions could complement event-level conditions with supervision for local actions and state changes.

\subsection{Pretraining Foundation Models for Autoregressive Video Generation}

Agentic control also calls for a model that can continue an evolving scene under changing conditions. MovieRoll adapts a pretrained bidirectional video model for streaming generation. Native autoregressive pretraining would align the model with this setting from the outset and allow us to study whether temporal causal attention can match bidirectional visual quality under comparable data and compute budgets.

This direction could build on language-model infrastructure while retaining bidirectional attention within each active chunk or window. Optimized kernels can skip temporally masked interactions, with savings determined by the chunk or window size and kernel implementation. At inference, KV caching could reuse fully denoised history as the model repeatedly denoises the active video tokens.

A common conditional interface could then support text-to-video, image-to-video, and continuation with a shared backbone. Adding source videos, masks, and control inputs would extend the same framework to video transformation, inpainting, and controlled generation. Joint pretraining would let us study how data diversity, model capacity, and context length affect quality and transfer. Across these tasks, shared temporal attention rules would connect training to inference, while robustness to self-generated history would remain a learning objective for maintaining continuity over extended video rollouts.

\begin{figure}[!t]
    \centering
    \includegraphics[width=0.88\linewidth]{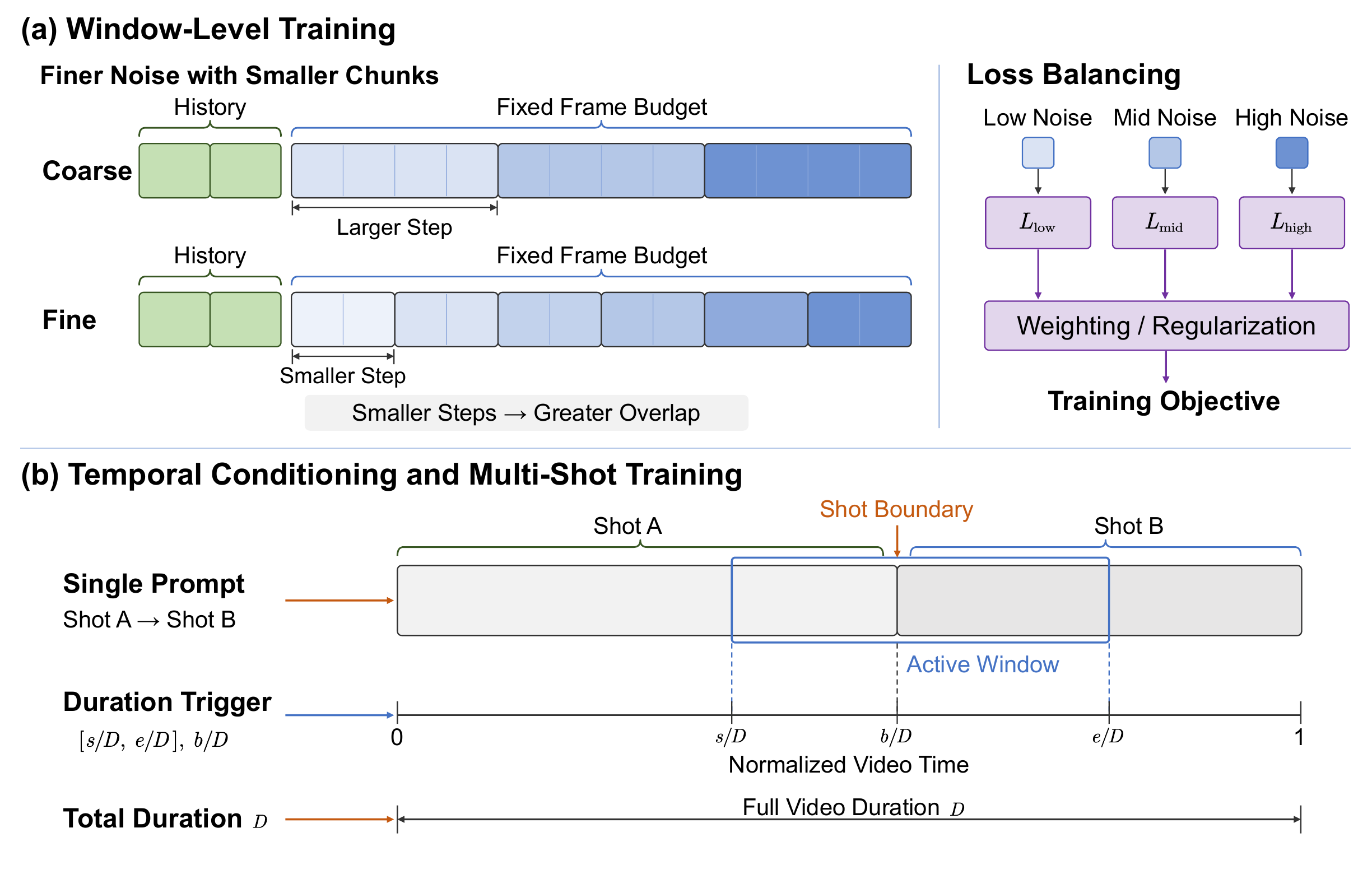}
    \caption{\textbf{Potential extensions of rolling-window video pretraining.} (a) Smaller chunks allow finer noise progression and temporal advances at a fixed frame budget, while loss weighting or regularization could balance learning across noise levels. (b) Total-duration conditioning could guide overall pacing, while duration triggers locate active windows and shot transitions. Together with multi-shot training, these conditions could support coordinated shot sequences under one prompt.}
    \label{fig:future_pretraining}
\end{figure}

WinTF offers a starting point for this training framework, as its ablation shows better preservation of long-horizon quality than chunk-wise teacher forcing in Tab.~\ref{tab:experiment_ablation}. Building on this result, Fig.~\ref{fig:future_pretraining} outlines two connected directions. Rolling-window optimization would address how generation advances through time, while explicit temporal conditions would specify its pace and shot structure.

\textbf{Finer Noise Levels with Smaller Chunks.} At a fixed active-frame budget, using more, smaller chunks would allow a denser noise progression and finer temporal advances. The chunk adjacent to clean history could therefore lie closer to the clean endpoint. Greater overlap between consecutive windows could further smooth the transition from clean history to noisy content and improve continuation.

\textbf{Regularization for Rolling-Window Losses.} Finer noise levels also motivate balancing their contributions to training. Loss weighting or regularization, guided by observed training dynamics, could account for differences in loss scale and optimization progress across the window.

\textbf{Duration Conditioning.} In bidirectional training, the temporal extent of the full attention window reveals the target duration, allowing the model to adjust motion speed and content planning accordingly. An autoregressive model sees only the available prefix and active window, which do not specify the total training-video length. Without this global duration cue, it tends to generate at the average motion speed learned for each position across training clips. Because text-to-video generation starts without history, it is biased toward the pacing of short clips, making it difficult to learn duration-dependent behavior from mixed-duration data. A \emph{total-duration condition} would address this ambiguity by pairing each training-window caption with the full video duration. At inference, a target duration, or pseudo-duration, would specify the time budget for motion and event planning.

While total duration sets the overall time budget, a \emph{duration trigger} would specify where the active window starts and ends and when shot transitions occur. These positions would be expressed as fractions of the total video duration. During training, they would come from the sampled window and observed shot boundaries. At inference, they would come from the current window and a specified shot schedule.

\textbf{Multi-Shot Data Training.} Multi-shot videos would provide supervision for applying these temporal conditions across changes in viewpoint or scene. A shared caption could describe the shots and their order across windows, while the duration trigger would locate transitions within each window. Training on windows that span shot boundaries would allow us to test whether window-wise denoising preserves within-shot content and follows the requested transition times.

Together, these temporal conditions could help the agent in Sec.~\ref{sec:future_work_agentic} translate event plans into paced sequences and shot transitions. Beyond controlling when events occur, a longer-term goal is to learn how scenes evolve under different actions, including the physical regularities and causal relationships reflected in diverse motions and interactions. Comparing predicted and observed consequences of different actions from the same initial visual state would provide one way to study this capability.

\clearpage
\bibliographystyle{plainnat}
\setlength{\bibhang}{0pt}
\setlength\bibindent{0pt}
\bibliography{main}

\newpage
\appendix
\section*{\centering
\Large Appendix
}

\section{Implementation Details of Main Experiments}
\label{app:implementation}

Sec.~\ref{sec:experiment_implementation_detail} summarizes the main training and inference settings. Tab.~\ref{tab:model_config}, Tab.~\ref{tab:inference_setting}, and Tab.~\ref{tab:training_hyperparameters} provide the model configuration and detailed hyperparameters.

\subsection{Training Data}
MovieRoll is trained on an internal video corpus that cannot be released publicly because of confidentiality restrictions. Tab.~\ref{tab:training_hyperparameters} reports the data scale, sample duration, resolution, and task composition for each stage.

\subsection{Streaming Fine-Tuning}
We assign increasing timesteps to chunks in temporal order within each rolling window. We first split $[0,1]$ at sorted random points, then sample a shared logit-normal relative position and use it to select a raw timestep within each sub-interval. Each raw timestep $q$ is shifted to $t=sq/[1+(s-1)q]$ before noising and model conditioning, with $s=3$ in both streaming fine-tuning stages, as specified in Tab.~\ref{tab:training_hyperparameters}. For each sampled phase, the first window uses up to the available number of chunks, and subsequent windows are full. A trailing remainder that cannot fill another window is omitted from that training pass. The flow-matching loss averages over all supervised latent entries. Occasionally, all chunks instead receive the same timestep. We also perturb clean history with noise to approximate the self-generated history encountered at inference. In Stage~2, all chunks in each training window share one prompt.

Base-stage clips are bucketed in steps of 8 frames. The batch size per data-parallel group decreases with clip length, from 6 clips at 33 frames to 1 clip at $\geq$121 frames.

\subsection{Streaming Architecture}
All 54 blocks read clean history from a KV cache. The first 6 blocks attend to all retained historical chunks, while the remaining 48 attend to the first chunk, which serves as an attention sink, and the most recent chunk. Within each layer group, cached chunks and the current window are indexed contiguously from zero, keeping temporal RoPE positions bounded as the video grows.

\subsection{Distillation}
For ODE initialization, the Stage~2 base model serves as a 50-step teacher with CFG linearly annealed from 4 to 1. The 5-step student learns to reproduce the clean endpoints of the rolling teacher trajectories, preserving the annealed guidance behavior with fewer sampling steps. After initialization, CuLT-DMD trains the generator on full videos produced with its 5-step schedule. For each predicted window, we prepend the nearest preceding generated latent frames to form a score clip of up to 31 latent frames. If fewer history frames are available, the score clip starts at the video beginning and remains unpadded. Only the current window contributes to the DMD loss. Motivated by the mixed training strategy of Rolling Forcing~\citep{rolling_forcing}, we regularize predictions across noise levels. With probability 0.5, we re-noise the predicted window at a shared timestep and run it through the generator again before computing the local DMD loss.

\subsection{Inference}
When the cached history reaches the recomputation threshold, we select chunks and rebuild the KV cache according to Tab.~\ref{tab:inference_setting}. This bounds memory usage and keeps positions within the RoPE range. For multi-event prompts, an event-prompt switch also triggers KV recomputation, with intermediate chunks retained only from the five most recent events.

\subsection{Ablation Settings}
The HieraKV quality ablation in Tab.~\ref{tab:experiment_ablation} trains each block allocation on the same data for 10K steps using the Stage~1 settings. These models use the main inference settings with the KV recomputation threshold changed to 15 chunks. The AnnealCFG ablation uses the Stage~1 MovieRoll-Base checkpoint and this same threshold, varying only the CFG schedule.

\begin{table}[!t]
  \centering
  \caption{\textbf{Shared architecture and history-cache configuration of the MovieRoll models.} Both variants use an 8.3-billion-parameter backbone with text conditioning, a causal video VAE, and a five-chunk active window. The first six transformer blocks retain full history, while the remaining 48 retain the sink and most recent chunks, preserving broad historical context through a small subset of layers.}
  \label{tab:model_config}
  \fontsize{9.5}{11.4}\selectfont
  \begin{tabular}{ll}
    \toprule
    Backbone             & MMDiT, 8.3B parameters, 54 dual-stream blocks \\
    Hidden Size / Heads  & 2048 / 16 \\
    Patch Size           & $1{\times}1{\times}1$ \\
    VAE                  & causal 3D VAE, $16{\times}16$ spatial and $4{\times}$ temporal compression, 32 channels \\
    Text Encoders        & Qwen2.5-VL-7B (up to 1000 tokens), Glyph-ByT5 (up to 256 tokens) \\
    Frame Rate           & 24\,fps \\
    \midrule
    Chunk Size           & 4 latent frames (16 frames), except 5 latent frames in the first chunk \\
    Rolling Window       & 5 chunks at staggered noise levels \\
    Full-History Blocks  & 1--6 \\
    Sink--Recent Blocks & 7--54 (first chunk + most recent chunk) \\
    \bottomrule
  \end{tabular}
\end{table}

\begin{table}[!t]
  \centering
  \caption{\textbf{Inference settings of MovieRoll-Base and MovieRoll-Distilled.} Distillation reduces sampling from 50 steps to 5 while preserving the shared rolling-generation pipeline. MovieRoll-Base uses guidance annealed from 4 to 1, whereas MovieRoll-Distilled operates without CFG. Both use the same chunk and window sizes and rebuild their KV caches at the same history threshold. Denoising passes are counted per output chunk, with the conditional and unconditional CFG branches contributing separately to the reported count.}
  \label{tab:inference_setting}
  \fontsize{9.5}{11.4}\selectfont
  \begin{tabular}{lcc}
    \toprule
                                & Base Model                        & Distilled Model \\
    \midrule
    Resolution / Frame Rate     & \multicolumn{2}{c}{$480{\times}864$ / 24\,fps} \\
    Chunk Size / Window Size    & \multicolumn{2}{c}{4 latent frames / 5 chunks} \\
    Sampling Steps              & 50 (10 per round)                 & 5 (1 per round) \\
    Timesteps                   & 50-step schedule, shift 5         & $\{1000,952,882,769,556\}$ \\
    CFG                         & Per-Frame, Linear $4\rightarrow1$ & None \\
    Denoising Passes per Chunk  & 20                                & 1 \\
    KV Context Noise $\sigma$   & 0.05                              & 0.15 \\
    KV Recomputation Threshold      & \multicolumn{2}{c}{18 chunks ($\approx$12s)} \\
    Chunks Kept after Recomputation & \multicolumn{2}{c}{sink + middle chunks of last 5 events + 3 most recent} \\
    \bottomrule
  \end{tabular}
\end{table}

\begin{table}[!t]
  \centering
  \caption{\textbf{Training stages and hyperparameters of MovieRoll.} The two base-model stages introduce single-event and multi-event streaming generation. ODE initialization then transfers the base model to a few-step student, and CuLT-DMD trains it on self-generated rollouts. The columns summarize the data and optimization settings supporting this progression. The two distillation phases are denoted by P1 and P2, while G and F identify the generator and fake-score critic in the optimization settings for each phase.}
  \label{tab:training_hyperparameters}
  \small
  \setlength{\tabcolsep}{5pt}
  \resizebox{\linewidth}{!}{%
    \begin{tabular}{lcccc}
      \toprule
                           & \multicolumn{2}{c}{Base Model} & \multicolumn{2}{c}{Distillation} \\
      \cmidrule(lr){2-3}\cmidrule(lr){4-5}
                           & Stage 1            & Stage 2            & ODE Init.          & CuLT-DMD \\
      \midrule
      Initialization       & Bidirectional T2V  & Stage 1 (EMA)      & Stage 2 (EMA)      & ODE Student \\
      Training Data        & 620K single-event clips
                           & \makecell{738K single-event clips\\+ 560K multi-event clips}
                           & 1.1K teacher trajectories
                           & \makecell{5K single-event prompts (P1)\\5K two-event prompts (P2)} \\
      Frames per Sample    & 33--241            & 33--361            & 241                & 241 (P1) / 721 (P2) \\
      Duration             & 1.4--10s         & 1.4--15s         & 10s              & 10s / 30s \\
      Resolution           & \multicolumn{2}{c}{multi-aspect, $\approx640^2$ pixels}
                           & $480{\times}864$, $864{\times}480$
                           & $480{\times}864$, $864{\times}480$ \\
      \midrule
      Teacher / Real Score & --                 & --
                           & \makecell{Stage 2, 50 steps\\CFG $4\rightarrow1$}
                           & \makecell{Bidirectional T2V\\CFG 4} \\
      Text Condition       & Single Event         & \makecell{Multiple Events (0.9)\\Single Event (0.1)}
                           & Single Event
                           & Single Event (P1) / Two Events (P2) \\
      Task (T2V/I2V) & 0.8 / 0.2    & 0.8 / 0.2    & T2V only           & 0.8 / 0.2 \\
      Context Noise        & \multicolumn{2}{c}{$\mathcal{U}(0,0.15)$ per chunk, $p{=}0.9$}
                           & $\mathcal{U}(0,0.15)$ per sample
                           & 0.15 (fixed) \\
      Null-Text Dropout    & \multicolumn{2}{c}{0.05 per sample + 0.05 per window} & -- & -- \\
      \midrule
      Objective            & \multicolumn{2}{c}{Flow Matching}       & ODE Regression     & DMD \\
      Timestep Shift       & 3                  & 3                  & 5                  & 5 \\
      Optimizer            & Muon               & Muon               & Muon               & Muon (G \& F) \\
      Learning Rate        & $2{\times}10^{-5}$ & $5{\times}10^{-6}$ & $1{\times}10^{-5}$ & $1{\times}10^{-5}$ (G \& F) \\
      Weight Decay         & 0.01               & 0.01               & 0.01               & 0.01 \\
      LR Schedule          & \multicolumn{4}{c}{Constant, No Warmup} \\
      Gradient Clipping    & 1.0                & 1.0                & 1.0                & 10.0 \\
      EMA Decay            & 0.9993             & 0.9993             & --                 & 0.99 (from step 200) \\
      Training Steps       & 10K                & 2K                 & 1K                 & 600 (P1) + 600 (P2) \\
      Batch Size (Videos)  & 32--192            & 16--96             & 32                 & 32 \\
      Precision            & \multicolumn{4}{c}{BF16 Mixed Precision} \\
      GPUs (SP Degree)     & 64 (2)             & 64 (4)             & 32 (1)             & 128 (4) \\
      \bottomrule
    \end{tabular}%
  }
    
\end{table}

\begin{table}[!t]
  \centering
  \caption{\textbf{Rollout settings for the two CuLT-DMD training phases.} Phase~1 uses single-event, 241-frame rollouts lasting 10s. Phase~2 extends training to two-event, 721-frame rollouts lasting 30s. Both phases average local gradients across each rollout, allowing one update to incorporate supervision from multiple temporal positions. Bounded score clips keep distillation supervision local as the generated sequence grows.}
  \label{tab:rolling_dmd_setting}
  \fontsize{9.5}{11.4}\selectfont
  \begin{tabular}{lcc}
    \toprule
                                    & Phase 1  & Phase 2 \\
    \midrule
    Frames                          & 241 (61) & 721 (181) \\
    Chunks / Supervision Windows $K$ & 15 / 3   & 45 / 9 \\
    Events per Video                & 1        & 2 (switch at frame 129--593) \\
    Global Steps / Critic Updates   & 600      & 600 \\
    Generator Updates               & 200      & 200 \\
    \midrule
    Generator Schedule              & \multicolumn{2}{c}{$\{1000,800,600,400,200\}$, shift 5, 1 step per round} \\
    Fake Score Initialization       & \multicolumn{2}{c}{Bidirectional T2V Model} \\
    Fake-Score / Generator Updates  & \multicolumn{2}{c}{3 to 1 (generator every third global step)} \\
    Score Clip                    & \multicolumn{2}{c}{up to 31 latent frames (121 frames)} \\
    Uniform-Timestep Regularization & \multicolumn{2}{c}{probability 0.5, shared $t\in\{952,882,769,556\}$} \\
    KV Recomputation (Training)         & \multicolumn{2}{c}{at 18 chunks, retaining sink + 1--5 random + 3 most recent} \\
    \bottomrule
  \end{tabular}
\end{table}

The HieraKV efficiency measurements in Fig.~\ref{fig:hierakv_vary_global_blocks} report peak denoising-step latency for MovieRoll-Base. Each timed step includes both conditional and unconditional DiT evaluations for CFG. Timing excludes KV recomputation and VAE encoding/decoding.

The AnnealCFG analysis in Fig.~\ref{fig:method_annealcfg} uses 32 paired prompts, 50 denoising steps, and a timestep shift of 5. Groups (a)--(d) cover prefill (chunks 0--4), early decoding (5--14), middle decoding (15--29), and late decoding (30--44). To obtain the curves, we average extrapolation impulse over latent frames within each emitted chunk, then over chunks within each group, and finally across prompts. For each prompt and group, the late-stage share is the sum of chunk-mean impulses over the final 10 steps divided by the corresponding sum over all 50 steps. Curves and bars show means over the 32 prompts. Dots represent individual prompts, and bands and whiskers indicate 95\% bootstrap confidence intervals.

The CuLT-DMD ablation compares three training strategies initialized from the same MovieRoll-Base checkpoint. CuLT-DMD and the LongLive-style baseline use 241-frame rollouts in Phase~1 and 721-frame rollouts in Phase~2. The short-rollout variant uses CuLT-DMD with 241-frame rollouts in both phases.

The LongLive-style baseline adapts the shared-rollout schedule of LongLive~\citep{longlive} to our architecture. Within a rollout, the generator and critic advance through supervision windows in temporal order, retaining detached history and KV states across parameter updates. Each generator update uses a single supervision window. Each phase contains 600 global steps, with one critic update per step and one generator update every third step, giving 600 critic and 200 generator updates. When a step updates both networks, they use different, consecutive supervision windows and complete both backward passes before either optimizer update. This update schedule is independent of rollout boundaries.

CuLT-DMD instead keeps both networks fixed throughout each rollout. The critic accumulates gradients over all $K$ supervision windows and updates once from their average at the rollout end, completing one global step. On every third global step, the generator also accumulates its DMD gradients across the rollout and updates from their average. The update counts and frequency match the LongLive-style baseline, but each update aggregates $K$ windows rather than one. We use $K=3$ for 241-frame rollouts and $K=9$ for 721-frame rollouts, as specified in Tab.~\ref{tab:rolling_dmd_setting}. All variants use the main inference settings.

\section{HeliosBench Evaluation Protocol}
\label{app:heliosbench}

We use long-video HeliosBench with five full-video quality metrics and four long-range drift metrics, as listed in Tab.~\ref{tab:heliosbench_models}. Full-video
metrics evaluate the average quality over the entire video. Drift metrics
measure the absolute score difference between the first and last
$\rho=0.15$ fraction of the video~\citep{framepack} as
\[
d_m(V)=
\left|q_m(V_{\mathrm{start}})-q_m(V_{\mathrm{end}})\right|,
\]
where $q_m$ denotes the base metric. Higher full-video scores and lower drift scores indicate better performance.

Motion smoothness retains every other frame and uses AMT-S to interpolate the omitted frames. The error $\Delta_{\mathrm{VFI}}$ is the mean absolute pixel difference between the interpolated frames and the original frames.

Aesthetic evaluation resizes and center-crops frames to $224\times224$ before CLIP encoding. Semantic evaluation resizes frames to width $640$ while preserving the aspect ratio. Naturalness evaluation resizes the long side to $512$ pixels and assigns an integer score $s\in\{1,\ldots,5\}$, with higher scores indicating more natural videos. We use three Gemini-2.5-Pro calls per video, while the original HeliosBench uses one GPT-5.2 call.

\begin{table}[!t]
\centering
\fontsize{9.5}{11.4}\selectfont
\setlength{\tabcolsep}{4pt}
\caption{\textbf{HeliosBench metrics, evaluators, and aggregation weights.} Five full-video metrics cover aesthetics, dynamics, motion smoothness, semantic alignment, and naturalness. Four drift metrics measure changes between the beginning and end of a video. The aggregation weights give greater emphasis to semantic alignment and naturalness while retaining contributions from all four metrics of temporal drift.}
\label{tab:heliosbench_models}
\begin{tabularx}{\linewidth}{@{}l l X c@{}}
\toprule
Metric & Model & Measurement & Dir./Weight \\
\midrule

Aesthetic
& CLIP ViT-L/14 + LAION Head
& All-frame aesthetic score divided by $10$ and averaged.
& $\uparrow$ / 0.030 \\

Dynamic
& Farneb{\"a}ck Optical Flow
& Absolute value after averaging downsampled flow components over pixels and frame pairs.
& $\uparrow$ / 0.030 \\

Motion Smoothness
& AMT-S
& Interpolation error at omitted frames, normalized as
$(255-\Delta_{\mathrm{VFI}})/255$.
& $\uparrow$ / 0.030 \\

Semantic Alignment
& ViCLIP (InternVid-10M-FLT)
& Cosine similarity between the prompt and eight middle-sampled frames.
& $\uparrow$ / 0.255 \\

Naturalness
& Gemini-2.5-Pro
& VLM score from 16 uniformly sampled frames, normalized as $(s-1)/4$ and averaged over
three calls.
& $\uparrow$ / 0.255 \\

Drifting Aesthetic
& Same as Aesthetic
& Absolute difference between the first and last $15\%$ window scores.
& $\downarrow$ / 0.099 \\

Drifting Motion Smoothness
& Same as Motion Smoothness
& Absolute difference between the first and last $15\%$ window scores.
& $\downarrow$ / 0.099 \\

Drifting Semantic
& Same as Semantic Alignment
& Absolute difference between the first and last $15\%$ window scores.
& $\downarrow$ / 0.099 \\

Drifting Naturalness
& Same as Naturalness
& Absolute difference between the first and last $15\%$ window scores.
& $\downarrow$ / 0.099 \\

\bottomrule
\end{tabularx}
\end{table}
\begin{table*}[!t]
\centering
\scriptsize
\setlength{\tabcolsep}{3.5pt}
\renewcommand{\arraystretch}{1.15}
\caption{\textbf{Conversion of HeliosBench quality and drift metrics to a common rating scale.} The listed intervals map raw metric values to ratings from 1 to 10 for weighted aggregation. Higher quality scores and lower drift scores receive higher ratings, so higher converted ratings consistently indicate better performance.}
\label{tab:heliosbench_scores}
\resizebox{\linewidth}{!}{%
\begin{tabular}{lcccccccccc}
\toprule
Metric
& Rating 10 & Rating 9 & Rating 8 & Rating 7 & Rating 6
& Rating 5 & Rating 4 & Rating 3 & Rating 2 & Rating 1
\\
\midrule

Aesthetic ($\uparrow$)
& $[0.70,\infty)$
& $[0.65,0.70)$
& $[0.60,0.65)$
& $[0.55,0.60)$
& $[0.50,0.55)$
& $[0.45,0.50)$
& $[0.40,0.45)$
& $[0.35,0.40)$
& $[0.30,0.35)$
& $(-\infty,0.30)$
\\

Dynamic ($\uparrow$)
& $[0.45,\infty)$
& $[0.40,0.45)$
& $[0.35,0.40)$
& $[0.30,0.35)$
& $[0.25,0.30)$
& $[0.20,0.25)$
& $[0.15,0.20)$
& $[0.10,0.15)$
& $[0.05,0.10)$
& $(-\infty,0.05)$
\\

Motion Smoothness ($\uparrow$)
& $[0.99,\infty)$
& $[0.98,0.99)$
& $[0.97,0.98)$
& $[0.96,0.97)$
& $[0.95,0.96)$
& $[0.94,0.95)$
& $[0.93,0.94)$
& $[0.92,0.93)$
& $[0.91,0.92)$
& $(-\infty,0.91)$
\\

Semantic Alignment ($\uparrow$)
& $[0.30,\infty)$
& $[0.29,0.30)$
& $[0.28,0.29)$
& $[0.27,0.28)$
& $[0.26,0.27)$
& $[0.25,0.26)$
& $[0.24,0.25)$
& $[0.23,0.24)$
& $[0.22,0.23)$
& $(-\infty,0.22)$
\\

Naturalness ($\uparrow$)
& $[0.65,\infty)$
& $[0.60,0.65)$
& $[0.55,0.60)$
& $[0.50,0.55)$
& $[0.45,0.50)$
& $[0.40,0.45)$
& $[0.30,0.40)$
& $[0.25,0.30)$
& $[0.20,0.25)$
& $(-\infty,0.20)$
\\

Drifting Aesthetic ($\downarrow$)
& $(-\infty,0.01]$
& $(0.01,0.02]$
& $(0.02,0.03]$
& $(0.03,0.04]$
& $(0.04,0.05]$
& $(0.05,0.06]$
& $(0.06,0.07]$
& $(0.07,0.08]$
& $(0.08,0.09]$
& $(0.09,\infty)$
\\

Drifting Motion Smoothness ($\downarrow$)
& $(-\infty,0.01]$
& $(0.01,0.02]$
& $(0.02,0.03]$
& $(0.03,0.04]$
& $(0.04,0.05]$
& $(0.05,0.06]$
& $(0.06,0.07]$
& $(0.07,0.08]$
& $(0.08,0.09]$
& $(0.09,\infty)$
\\

Drifting Semantic ($\downarrow$)
& $(-\infty,0.01]$
& $(0.01,0.02]$
& $(0.02,0.03]$
& $(0.03,0.04]$
& $(0.04,0.05]$
& $(0.05,0.06]$
& $(0.06,0.07]$
& $(0.07,0.08]$
& $(0.08,0.09]$
& $(0.09,\infty)$
\\

Drifting Naturalness ($\downarrow$)
& $(-\infty,0.06]$
& $(0.06,0.08]$
& $(0.08,0.10]$
& $(0.10,0.12]$
& $(0.12,0.14]$
& $(0.14,0.16]$
& $(0.16,0.18]$
& $(0.18,0.20]$
& $(0.20,0.22]$
& $(0.22,\infty)$
\\

\bottomrule
\end{tabular}%
}
\end{table*}
\begin{table}[!t]
\centering
\caption{\textbf{Comparison of GPT-5.2 and Gemini-2.5-Pro as video evaluators.} Both evaluate the same 30s single-event videos and rank MovieRoll-Distilled first and MovieRoll-Base second. Method rankings agree closely, with Spearman correlation $\rho=0.930$. We use Gemini-2.5-Pro because it provides greater score separation between methods, as shown by the statistical measures in the lower section.}
\label{tab:gpt_vs_gemini}

\footnotesize
\setlength{\tabcolsep}{2.0pt}
\renewcommand{\arraystretch}{1.05}

\scalebox{1.19}{
\begin{tabular}{@{}lcccc@{}}
    \toprule
    & \multicolumn{2}{c}{\tablehead{GPT-5.2}}
    & \multicolumn{2}{c}{\tablehead{Gemini-2.5-Pro}} \\
    \cmidrule(lr){2-3}
    \cmidrule(lr){4-5}
    \tablehead{Method}
    & \tablehead{Score}
    & \tablehead{Rank}
    & \tablehead{Score}
    & \tablehead{Rank} \\
    \midrule

    CausVid
    & 0.1446 & 11
    & 0.0208 & 12 \\

    Self Forcing
    & 0.1596 & 10
    & 0.0392 & 11 \\

    Causal Forcing
    & 0.2508 & 9
    & 0.1421 & 9 \\

    Rolling Forcing
    & 0.3217 & 8
    & 0.3625 & 8 \\

    MemFlow
    & 0.3325 & 7
    & 0.4612 & 4 \\

    LongLive
    & 0.3575 & 5
    & 0.4417 & 5 \\

    Echo-Infinity
    & 0.3412 & 6
    & 0.3667 & 7 \\

    Krea
    & 0.1117 & 12
    & 0.0617 & 10 \\

    Helios-Base
    & 0.4271 & 3
    & 0.5592 & 3 \\

    Helios-Distilled
    & 0.4179 & 4
    & 0.3862 & 6 \\

    \addlinespace[1.5pt]
    \rowcolor{MovieRollColor}
    \tablenum{MovieRoll-Base}
    & 0.4825 & 2
    & 0.6096 & 2 \\

    \rowcolor{MovieRollColor}
    \tablenum{MovieRoll-Distilled}
    & \tablenum{0.6096} & \tablenum{1}
    & \tablenum{0.7979} & \tablenum{1} \\

    \midrule
    \multicolumn{5}{c}{
        \textit{Method-Level Score Separation}
    } \\
    \midrule

    Mean Range
    & \multicolumn{2}{c}{0.498}
    & \multicolumn{2}{c}{\tablenum{0.777}} \\

    Mean Std
    & \multicolumn{2}{c}{0.141}
    & \multicolumn{2}{c}{\tablenum{0.235}} \\

    $\eta^2$
    & \multicolumn{2}{c}{0.202}
    & \multicolumn{2}{c}{\tablenum{0.286}} \\

    $F$ Statistic
    & \multicolumn{2}{c}{55.0}
    & \multicolumn{2}{c}{\tablenum{86.8}} \\

    \bottomrule
\end{tabular}
}

\end{table}

\subsection{Raw-Score Conversion}
We convert raw scores to ratings $r_m\in\{1,\ldots,10\}$ using the thresholds in Tab.~\ref{tab:heliosbench_scores}. The nine thresholds correspond to ratings $10$ through $2$ in order, and the first satisfied threshold determines the rating. The inequality follows whether higher or lower scores are better. If no threshold is satisfied, the rating is $1$.

The final HeliosBench score is the Total Weighted Rating (TWR), defined as
\[
\mathrm{TWR}=\sum_{m=1}^{9}w_m r_m,
\]
where the weights $w_m$, listed in Tab.~\ref{tab:heliosbench_models}, sum to approximately $0.996$, giving a theoretical maximum TWR of $9.96$ when all nine metrics receive the highest rating of $10$.

\subsection{Multi-Event Evaluation}
For multi-event videos, we compute semantic and drifting-semantic scores for each event using its prompt. The video-level scores average the event-level semantic scores and absolute beginning--end differences, respectively.

\subsection{Choice of the Naturalness Evaluator}
We compare GPT-5.2 and Gemini-2.5-Pro on the same 2400 videos from the 30s (721-frame) single-event benchmark, with 200 videos for each of 12 methods. Each evaluator scores every video three times, and we average the three scores.

The method-level rankings show strong agreement, with Spearman correlation $\rho = 0.930$, as reported in Tab.~\ref{tab:gpt_vs_gemini}. Both evaluators rank MovieRoll-Distilled first and the evaluated MovieRoll-Base checkpoint second.

We adopt Gemini-2.5-Pro because it provides greater between-method score separation on every statistical measure in Tab.~\ref{tab:gpt_vs_gemini}.

\section{User Study Protocol}
\label{app:user_study}

The user study compares MovieRoll-Distilled with the main-evaluation baselines. Each baseline is evaluated on 10 single-event or five multi-event prompt cases per duration. In total, 59 participants provide 2616 completed judgments on 350 video pairs. Each pair is evaluated by 3--12 distinct participants.

Tab.~\ref{tab:user_study_counts} reports completed judgments by baseline and duration. For each baseline, we pool judgments across durations separately for single- and multi-event generation. Each judgment contributes one vote, with ties included in the denominator of win, tie, and loss rates. Durations are therefore weighted by their numbers of completed judgments.

\begin{table}[!t]
    \centering
    \caption{\textbf{Distribution of user-study judgments across baselines and video durations.} The study collects 2616 preference judgments across single-event and multi-event comparisons. Each cell counts individual judgments for MovieRoll-Distilled versus the listed baseline at a given duration. Judgments are pooled across durations to calculate win, tie, and loss rates for each baseline, separately for single-event and multi-event generation. Each judgment contributes one vote, including ties.}
    \label{tab:user_study_counts}
    \fontsize{9.5}{11.4}\selectfont
    \begin{tabular}{lrrrrr}
        \toprule
        & \multicolumn{3}{c}{Single-Event} & \multicolumn{2}{c}{Multi-Event} \\
        \cmidrule(lr){2-4}\cmidrule(lr){5-6}
        Baseline & 10s & 30s & 60s & 30s & 60s \\
        \midrule
        CausVid          & 74 & 74 & 78 & -- & -- \\
        Self Forcing     & 79 & 81 & 89 & -- & -- \\
        Causal Forcing   & 76 & 72 & 69 & -- & -- \\
        Rolling Forcing  & 73 & 78 & 73 & -- & -- \\
        MemFlow          & 76 & 72 & 82 & 31 & 38 \\
        LongLive         & 77 & 75 & 63 & 38 & 37 \\
        Echo-Infinity    & 73 & 73 & 69 & 37 & 33 \\
        Krea             & 70 & 69 & 76 & -- & -- \\
        Helios-Base      & 73 & 78 & 76 & 40 & 39 \\
        Helios-Distilled & 79 & 66 & 77 & 42 & 41 \\
        \midrule
        Total Judgments & 750 & 738 & 752 & 188 & 188 \\
        \bottomrule
    \end{tabular}
\end{table}

\section{Infrastructure Details}
\label{app:infrastructure_details}

\subsection{FBCA Packing and Grouping}
\label{app:fbca_details}

Sec.~\ref{sec:infra} and Eq.~\ref{eq:fbca-visible-sequences} define the FBCA blocks and their visible key/value sequences. FBCA concatenates these sequences in batch--block order and records their cumulative lengths. Separate query boundaries match each query block to its visible keys and values. Variable-length FlashAttention evaluates these sequence pairs independently, while equal-length pairs can use the batched kernel.

For equal-sized blocks, each visible video sequence contains $P+(i+1)T_{\mathrm{blk}}$ tokens, giving the total packed length in Eq.~\ref{eq:fbca-packed-length}. If the first block has a different size $T_0$, it is processed separately. Its historical states then join the shared prefix for subsequent blocks, whose visible lengths are $P+T_0+iT_{\mathrm{blk}}$ for $i\geq1$.

In multimodal attention, each video block also attends to its own valid text states. Text queries attend only to video and valid text tokens within that block. Padded text tokens are removed before packing, and video and text queries use separate attention calls when their visible keys differ. Outputs are returned in the original batch--block order, with padded text positions restored.

The custom \texttt{gather\_concat} operator copies visible segments into a contiguous buffer. For a fixed packing plan, this is a linear gather with backward accumulation
\begin{equation}
    \mathbf{y}=\mathbf{A}\mathbf{x},
    \qquad
    \frac{\partial\mathcal{L}}{\partial\mathbf{x}}
    =\mathbf{A}^{\mathsf{T}}
      \frac{\partial\mathcal{L}}{\partial\mathbf{y}}.
\end{equation}
The transpose adds contributions from every packed occurrence of a source entry, preserving gradients through repeated prefix and history states. For FP16/BF16 inputs, the implementation accumulates in FP32 when the total referenced token count exceeds the source length, then casts back to the incoming gradient dtype.

To limit the packed buffer for each attention call, FBCA partitions consecutive blocks into groups. For a group $g=[s,e)$, the augmented prefix $\mathbf{X}_{\mathrm{full},g}^v$ concatenates the original prefix and historical blocks $0,\ldots,s-1$. Block $i\in g$ then sees
\begin{equation}
    \overline{\mathbf{X}}_{i,g}^v
    =\operatorname{concat}\!\left(
        \mathbf{X}_{\mathrm{full},g}^v,
        \mathbf{X}_{\mathrm{hist},s}^v,\ldots,
        \mathbf{X}_{\mathrm{hist},i-1}^v,
        \mathbf{X}_i^v
    \right)
    =\overline{\mathbf{X}}_i^v,
    \quad \mathbf{X}\in\{\mathbf{K},\mathbf{V}\}.
    \label{eq:fbca-group-equivalence}
\end{equation}
Empty history ranges are omitted. Grouping preserves both the attention function and its analytical gradients because every query retains the same ordered keys and values.

Let $T_g=\sum_{i\in g}|\overline{\mathbf{K}}_i^v|$ be the packed video length of group $g$. With memory budget $M_{\mathrm{budget}}$, we use
\begin{equation}
    T_g\leq T_{\max},
    \qquad
    T_{\max}=\left\lfloor
      \frac{M_{\mathrm{budget}}}{4BHD\,b}
    \right\rfloor,
\end{equation}
where $B$ is batch size, $H$ the head count on the current device, $D$ the head dimension, and $b$ the bytes per element. The factor four counts $K$, $V$, $dK$, and $dV$. For equal-sized blocks, the grouping minimizes the number of contiguous groups under this budget, then minimizes the largest group at that group count. A budget smaller than one visible sequence is rejected. A differently sized first block is handled separately before grouping the remaining blocks.

Grouping leaves $T_{\mathrm{pack}}=\sum_g T_g$ unchanged. Without recomputation, keys and values saved across the attention calls still require approximately $2BT_{\mathrm{pack}}HD\,b$ bytes. The budget covers video key/value positions, with text states, queries, and kernel workspace requiring additional memory. Grouping reduces individual buffers at the cost of more kernel launches and prefix assembly.

\subsection{PS-VAE State and Spatial Decomposition}
\label{app:psvae_details}

As described in Sec.~\ref{sec:infra_PSVAE}, PS-VAE combines temporal caching with spatial computation over row bands. Its state contains convolution input suffixes, temporal upsampling phase indicators, and projected bottleneck attention keys and values. Each layer retains its state across chunks of the same video, and all states are reset before decoding a new video. This state is separate from the DiT history managed by HieraKV and threshold-triggered KV recomputation.

For causal convolution with a three-frame kernel, unit stride and dilation, left replicate padding gives
\[
    y_t=\sum_{i=0}^{2}\mathcal{W}_i\ast\widetilde{x}_{t-2+i},
    \qquad \widetilde{x}_{\tau}=x_{\max(\tau,0)},
\]
where $\mathcal{W}_i\ast$ is the spatial convolution for temporal slice $i$, with bias omitted. Each output needs only two preceding inputs. Streaming therefore prepends the cached suffix to the current chunk and adds replicate padding only when fewer than two past inputs exist. After convolution, the cache retains the last two available real inputs, excluding padding. This supplies exactly the same temporal inputs as full-sequence convolution.

At each temporal upsampling layer, the global first frame produces one output frame and each later frame produces two. With two such layers, the decoded length is
\[
    T_{\mathrm{pix}}=1+4(T_{\mathrm{lat}}-1).
\]
An initial chunk of $L$ latent frames thus produces $4(L-1)+1$ pixel frames, and each later chunk produces $4L$. Per-layer phase indicators ensure that the special first-frame branch runs only once.

Bottleneck attention is causal between frames and bidirectional within each frame. With $P$ cached frames and $L$ current frames, let $a\in\{0,\ldots,L-1\}$ index a current query frame and $b\in\{0,\ldots,P+L-1\}$ index a concatenated key frame. Every spatial token pair between those frames uses the mask
\[
    M_{a,b}=
    \begin{cases}
        0, & b\leq P+a,\\
        -\infty, & b>P+a.
    \end{cases}
\]
All cached frames and the current or earlier frames in the active chunk are visible. With untruncated history, the cached keys and values match those used by full-sequence attention. A finite cache instead retains the latest $N_{\mathrm{VAE}}$ latent frames after evaluating the current chunk, so truncation affects the next chunk. This budget counts decoder input frames, whereas the DiT recomputation threshold counts generation chunks.

These correspondences establish streaming equivalence with full history. Convolution caches preserve past inputs, phase indicators align temporal upsampling, and attention caches preserve visible keys and values. Pointwise normalization, nonlinearities, and residual additions preserve the correspondence, so concatenating the decoded chunks recovers the full-sequence decoder output.

\subsubsection{Row-Sharded Parallel Decoding}
Temporal chunks are decoded in order. Spatial parallelism begins after the global bottleneck attention, where the remaining operators have local spatial receptive fields. Each rank computes the input convolution and middle block, then processes a contiguous row band through the upsampling blocks and output projection.

For latent height $H$ and $R$ ranks, rank $r$ owns $n_r$ rows starting at $a_r$, where
\[
    n_r=\left\lfloor\frac{H}{R}\right\rfloor
        +\mathbf{1}[r<H\bmod R],
    \qquad a_r=\sum_{q<r}n_q.
\]
We require $H\geq R$ so that every rank receives at least one row. Before each $3\times3$ spatial convolution, neighboring ranks exchange one boundary row. At a global boundary, replicate padding supplies the missing row. Local temporal and width padding then reproduce the full decoder inputs. Convolution caches retain the temporal suffix including its halos, so later chunks exchange only current boundary rows.

Pixel shuffling maps input row $h$ to rows $2h$ and $2h+1$, preserving row ownership through upsampling. After four spatial upsampling stages, rank $r$ owns pixel rows $[16a_r,16(a_r+n_r))$. An all-gather reconstructs each complete decoded chunk. Unequal row bands are padded for communication and cropped afterwards, and all ranks process the same latent chunks in the same order.

\subsubsection{Combined Design and Computational Cost}
Temporal caches provide past inputs, while halo exchange provides neighboring rows. With the same chunk partition and cache policy, row sharding preserves single-device streaming decoding up to numerical differences from precision and kernel selection. Memory savings come primarily from distributing high-resolution upsampling and its convolution caches, while bottleneck attention remains replicated. For replicated computation $F_{\mathrm{rep}}$ and sharded computation $F_{\mathrm{band}}$, the ideal speedup is
\[
    S(R)=\frac{F_{\mathrm{rep}}+F_{\mathrm{band}}}
                   {F_{\mathrm{rep}}+F_{\mathrm{band}}/R}.
\]
This assumes evenly loaded ranks and excludes communication and synchronization. Replicated bottleneck computation limits the speedup as more ranks are added.

\subsection{Inference Speed Measurement}
\label{app:inference_speed_measurement}

We benchmark MovieRoll-Distilled text-to-video generation on a single node with NVIDIA H20 GPUs, as reported in Tab.~\ref{tab:inference_speed_scaling}. Each video contains 1441 frames at $480\times864$ and 24 FPS, corresponding to approximately 60s and 90 rollout chunks. We compare FlashAttention-3~\citep{flash_attn3} and SageAttention~\citep{sage_attn,sage_attn2} at sequence-parallel degrees of 1, 2, 4, and 8, with FSDP disabled. Inference uses BF16 precision except for VAE decoding, which uses FP16. PS-VAE uses the sequence-parallel ranks, with single-device streaming decoding at degree 1.

For each configuration, we generate two videos sequentially with seed 930 and the prompt ``A red fox trots across a snowy clearing at dawn, its breath visible in the cold air.'' We report timings for the second video. Total time runs from the first output-save log entry to the second, covering denoising, decoding, output handling, and saving the decoded tensor in \texttt{.pt} format. Model loading and initial text encoding occur before this interval. We read denoising time from the final progress entry of the second rollout at one-second resolution, counting duplicate progress entries only once. Throughput is 1441 divided by the unrounded total time. GPU monitoring confirms that no keeper processes run during the measured generation intervals.

\section{Streaming Video Prompt Rewriting Skill}
\label{app:prompt_rewriting}

The system prompt below defines the rewriting interfaces, continuity rules, and validation procedure for the skill in Sec.~\ref{sec:method_rewrite_skill}. Multi-event evaluation uses the sequential interface with only past and current events.


\begin{tcolorbox}[
    breakable,
    colback=gray!3,
    colframe=gray!60,
    title={Streaming Video Prompt Rewriting Skill},
    fonttitle=\bfseries\small,
    fontupper=\small
]

\textbf{Description.} Rewrites short user inputs into structured,
single-paragraph prompts for streaming (indefinitely extended) video
generation, in single-event and interactive multi-event settings.\par
\textbf{Goal.} Expand a short user input into an objective, detailed
prompt that describes a scene which can be extended chunk by chunk
indefinitely, without reaching a dead end.\par

\textbf{Modes.}
\begin{itemize}
\setlength{\itemsep}{0pt}
\setlength{\parsep}{0pt}
\setlength{\topsep}{2pt}
    \item In \textit{Single} mode, one input sentence $\rightarrow$ one prompt.
    \item In \textit{Sequential (interactive)} mode, given the finalized rewrites
    of all previous events and one new raw event, rewrite only the new
    event. Future events are never visible.
    \item In \textit{Batch} mode, $N$ known events $\rightarrow$ $N$ prompts,
    separated by \texttt{@@@EVENT\_BREAK@@@}.
\end{itemize}

\textbf{Hard constraints.}
\begin{enumerate}
\setlength{\itemsep}{0pt}
\setlength{\parsep}{0pt}
\setlength{\topsep}{2pt}
    \item \textit{Camera.} Specify exactly one base behavior in a single canonical sentence for
    a static camera (``The camera remains static, keeping
    \textit{[subject]} centered.'') or a tracking camera (``The camera tracks
    \textit{[subject]}, continuously keeping \textit{[subject]}
    centered.''). Use tracking when the subject moves through space and
    static otherwise. Convert any other movement (pan, zoom, dolly, tilt,
    orbit, crane, roll) to the closer of the two. A static camera never
    reframes.

    \item \textit{Centering.} The main subject stays at the frame center
    throughout. Close-ups only for stationary subjects.

    \item \textit{Open-ended action.} Describe ongoing or repeating motion.
    Never use ending words (``finally,'' ``eventually,'' ``in the end,''
    ``at last,'' ``wraps up''). After the opening beats, switch to
    continuity framing (``Throughout,'' ``Meanwhile'').

    \item \textit{Natural pace.} Real-world speed unless the input specifies
    otherwise. Avoid stacked intensifiers. Summarize crowds instead of
    animating many small elements. Under a static camera the subject must
    still visibly move (breathing, swaying, gesturing).

    \item \textit{Fidelity.} Keep every entity, exact count, event order,
    requested style, and on-screen text. Never invent or drop subjects.

    \item \textit{Style.} Default to cinematic realism, also for fantastical
    subjects, unless the input names another style. End with a single short
    style tag, e.g., ``Cinematic photography realistic style.''

    \item \textit{Format.} Output only the prompt as a single paragraph in
    the present tense, without self-referential phrases
    (e.g., ``this video shows''). Use the same language as the input. If the
    language is unclear or unsupported, default to English. Keep the output
    to 150--300 words, or an equivalent length for languages where word
    counts are not directly comparable.
\end{enumerate}

\textbf{Prompt structure.} Summary sentence naming all subjects
$\rightarrow$ subject appearance $\rightarrow$ chronological action
(\textit{Initially} / \textit{Then} / \textit{Next}
$\rightarrow$ \textit{Throughout}) $\rightarrow$ environment, where every
sound has a visible source $\rightarrow$ restatement of the subject's fixed
traits $\rightarrow$ camera sentence, shot size, lighting, mood
$\rightarrow$ style tag.\par

\textbf{Multi-event consistency.}
\begin{itemize}
\setlength{\itemsep}{0pt}
\setlength{\parsep}{0pt}
\setlength{\topsep}{2pt}
    \item One camera setup (behavior, viewpoint, angle, shot size) for the
    whole sequence. New subjects enter the established frame. Switch between
    static and tracking only when an event requires it.
    \item Recurring entities keep identical descriptions. The setting stays
    the same unless the input moves it.
    \item Each prompt reintroduces entities without cross-event pronouns, using their
    full noun phrase (``It glides toward the diver.''
    $\rightarrow$ ``The giant manta ray glides toward the diver in the black
    wetsuit.'').
    \item Each event continues from the previous state and receives the full
    length budget.
\end{itemize}

\textbf{Validation.} Every output is checked automatically for length,
single paragraph, ending words, self-reference, forbidden camera motion,
exactly one camera sentence, and explicit centering under tracking.
Failed outputs are regenerated with the violations appended as feedback,
for up to 3 attempts.

\end{tcolorbox}

\end{document}